\documentclass[letterpaper]{article} 
\usepackage{aaai2027}  
\usepackage[hyphens]{url}  
\usepackage{graphicx} 
\usepackage{natbib}  
\usepackage{caption} 
\usepackage{algorithm}
\usepackage{algorithmic}

\usepackage{newfloat}
\usepackage{listings}
\DeclareCaptionStyle{ruled}{labelfont=normalfont,labelsep=colon,strut=off} 
\floatstyle{ruled}
\newfloat{listing}{tb}{lst}{}
\floatname{listing}{Listing}

\usepackage{booktabs}
\usepackage{colortbl}
\usepackage{longtable}
\usepackage{multirow}

\nocopyright

\definecolor{lightlavender}{RGB}{230,230,250}
\usepackage[most]{tcolorbox}

\title{Geo-Embed: Towards Unified Multimodal Embeddings for Urban Understanding}

\author{
    Jiapeng Li\textsuperscript{\rm 1}\equalcontrib,
    Yong Li\textsuperscript{\rm 1}\equalcontrib,
    Junjie Zhou\textsuperscript{\rm 2},
    Fan Zhang\textsuperscript{\rm 1},
    Yu Liu\textsuperscript{\rm 1}\corresponding
}
\affiliations{
    \textsuperscript{\rm 1}Peking University
    \textsuperscript{\rm 2}Beijing University of Posts and Telecommunications\\
    Beijing, China\\
}

\begin{document}

\maketitle

\begin{abstract}
Geospatial and urban applications increasingly require models to compare heterogeneous evidence across street-view imagery, remote-sensing observations, text descriptions, region proposals, and temporal change cues. However, existing multimodal embedding models and benchmarks are still largely designed and evaluated around general-purpose image-text matching, leaving unclear whether unified embedding space can support heterogeneous geospatial tasks involving spatial relationships, fine-grained semantics, and temporal changes. To address this gap, we make three key contributions. First, we introduce \textbf{GeoMEB}, a large-scale multimodal embedding benchmark that standardizes 45 urban evaluation tasks across retrieval, visual question answering, change detection, classification, and visual grounding, together with training collections comprising 1.32M examples and 286K evaluation queries. Second, we present \textbf{Geo-Embed}, a unified embedding model that adapts a shared vision-language backbone to instruction-conditioned query-target matching over heterogeneous geospatial inputs, including single images, multiple images, text, regions, and masks. On GeoMEB, Geo-Embed achieves the strongest overall performance among representative multimodal embedders, with a 15.3\% relative improvement over the strongest baseline. These results motivate future geospatial embedders that organize training and evaluation around explicit query-target relations, including semantic, cross-view, region-level, and temporal correspondence.

\end{abstract}


\section{Introduction}

Embedding models are becoming a core interface for geospatial and urban intelligence because many city-scale applications depend on comparing heterogeneous evidence about places~\cite{brown2025alphaearthc,liu2024remoteclipa,zhang2024rs5m}. Urban search, geo-localization, disaster monitoring, and socioeconomic sensing all require matching visual, textual, spatial, or temporal observations to task-relevant targets~\cite{lu2018exploring,wang2024skyscript,chen2020spatialtemporal,liuCityLensEvaluatingLarge2025}. This shared comparison problem makes embedding quality central to scalable urban analysis. Although these applications differ in form, they require urban queries and candidate targets to be represented in a shared space where task-relevant similarity can be computed efficiently.

Despite rapid progress in CLIP-style dual encoders and recent VLM-based
embedders~\cite{radfordLearningTransferableVisual2021,jiangVLM2VecTrainingVisionLanguage2025,zhangGMEImprovingUniversal2025,liQwen3VLEmbeddingQwen3VLRerankerUnified2026}, their ability to serve as a unified interface for urban intelligence remains unclear. First, existing embedding benchmarks are still dominated by generic image-text retrieval, general multimodal reasoning, or mixed-domain retrieval settings~\cite{mengVLM2VecV2AdvancingMultimodal2025,huangMMEBV3MeasuringPerformance2026,xiaomieb}, and they do not systematically organize urban problems under a unified, task-aware common embedding protocol. Second, urban similarity is relation-specific rather than purely semantic: cross-view geo-localization depends on viewpoint alignment~\cite{zhu2021vigora,jia2024g3}, region grounding depends on spatial granularity and region correspondence~\cite{ertler2020mapillary,li2024vrsbench}, and disaster monitoring depends on temporal change sensitivity~\cite{alcantarilla2018streetviewa,chen2020spatialtemporal,huang2025built}.
As a result, current models and evaluation resources provide limited evidence on whether a single representation space can support heterogeneous urban query-target relations across modalities, viewpoints, regions, and temporal states.

\begin{figure*}[t]
    \centering
    \includegraphics[width=0.98\textwidth]{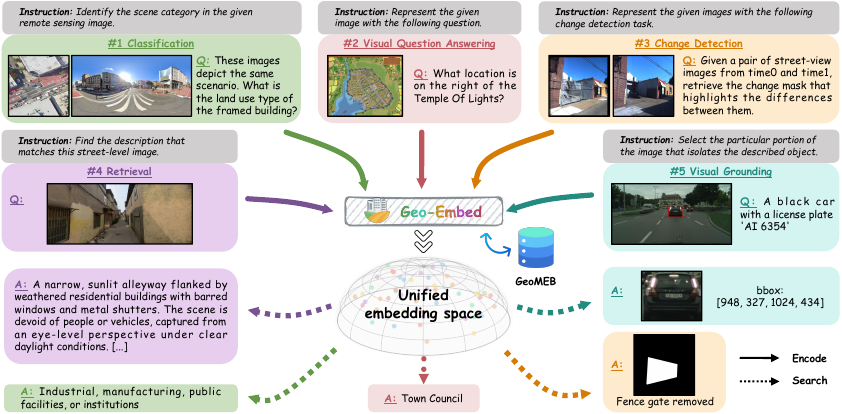}
    \caption{Overview of GeoMEB and Geo-Embed. GeoMEB unifies classification, visual question answering, change detection, retrieval, and visual grounding as instruction-conditioned embedding tasks; Geo-Embed encodes heterogeneous urban queries and target candidates into a shared representation space, enabling efficient similarity-based search across text, single- and multi-image inputs, regions, and masks.}
    \label{fig:GeoMEB_overview}
\end{figure*}

To address these gaps, we introduce \textbf{GeoMEB}, a geospatial multimodal embedding benchmark that standardizes 45 tasks under a unified instruction-conditioned ranking protocol. GeoMEB covers retrieval, visual question answering, change detection, classification, and visual grounding, and casts each task as matching an instructed query to candidates from a task-specific pool. The benchmark includes 1.32M training examples and 286K evaluation queries across image, text, interleaved image-text, multi-image, region, and mask inputs (Figure~\ref{fig:GeoMEB_overview}). We further present \textbf{Geo-Embed}, a unified embedding model for
geospatial multimodal data. Geo-Embed encodes queries and targets with side-specific instructions using a shared vision-language backbone, and trains the resulting representations with contrastive supervision over heterogeneous geospatial pairs.

Experiments show that Geo-Embed achieves the strongest performance among representative CLIP-family models, instruction-tuned VLMs, and embedding-specialized VLMs. More importantly, the results suggest that future geospatial embedders should organize training and evaluation around query-target relations, as
semantic matching, cross-view alignment, region grounding, and temporal change matching respond differently to data composition and task mixing.

Our main contributions are as follows:
\begin{itemize}
    \item We introduce GeoMEB, a 45-task geospatial multimodal embedding
    benchmark with a unified instruction-conditioned ranking protocol.
    \item We develop Geo-Embed, an instruction-conditioned VLM embedding model
    for heterogeneous geospatial query-target matching.
    \item We provide a systematic study of multimodal embedders on GeoMEB, covering
baseline comparisons, ablations, instruction sensitivity, and representation
analyses.
\end{itemize}






\section{Related Work}

\subsection{Multimodal embedding models}
Multimodal embedding models map inputs from different modalities into a shared representation space for efficient retrieval. Existing approaches use either modality-specific dual encoders or VLM-based embedders that process interleaved inputs and task instructions. CLIP-style dual encoders learn image-text alignment with a contrastive objective and permit target candidate embeddings to be precomputed~\cite{radfordLearningTransferableVisual2021,tschannen2025siglip}. However, fixed towers are less suited to interleaved inputs or instructions that redefine the desired relation. VLM2Vec exemplifies the second paradigm by appending task instructions, using the final-token hidden state as the sequence representation, and optimizing it with InfoNCE~\cite{jiangVLM2VecTrainingVisionLanguage2025,oord2019representation}. Data-centric methods scale paired supervision for universal retrieval~\cite{zhou2024megapairs}. Other systems broaden task and input coverage to videos, visual documents, audio, and universal multimodal retrieval~\cite{zhangGMEImprovingUniversal2025,mengVLM2VecV2AdvancingMultimodal2025,xu2025omniembednemotrona}. Recent native multimodal embedders further integrate retrieval and ranking within stronger VLM backbones~\cite{liQwen3VLEmbeddingQwen3VLRerankerUnified2026,shanbhogueGeminiEmbedding22026}. Reasoning-oriented approaches generate context or optimize retrieval behavior before producing embeddings~\cite{cui2025think,lan2025umer1,jiangEmbedRLReinforcementLearning2026}.

\subsection{Geospatial and urban representation learning}
Urban applications such as cross-view localization, temporal change analysis, and large-scale mapping require representations that remain informative across sensors, viewpoints, spatial scales, and time. Contrastive remote-sensing models align overhead imagery with language supervision at scale~\cite{liu2024remoteclipa,zhang2024rs5m,wang2024skyscript}. VLM-based approaches extend this alignment to grounded instructions and interleaved geospatial inputs~\cite{kuckreja2024geochat}. AlphaEarth Foundations instead learns a global geospatial embedding field for large-scale mapping~\cite{brown2025alphaearthc}. These advances provide strong representations for remote-sensing retrieval, grounded understanding, and global mapping. Urban settings, however, require a common representation to relate overhead and street-level observations, geographic scales, temporal states, regions, and urban semantics. Existing geospatial methods typically cover only a subset of these relations within separate task formulations.

\subsection{Multimodal embedding benchmarks}
Multimodal embedding benchmarks evaluate whether shared representations generalize across tasks, input modalities, and reasoning requirements. MMEB provides a broad suite of multimodal retrieval tasks~\cite{jiangVLM2VecTrainingVisionLanguage2025}. Its extensions add videos and visual documents~\cite{mengVLM2VecV2AdvancingMultimodal2025}, while MMEB-V3 expands evaluation to audio and agent-centric scenarios~\cite{huangMMEBV3MeasuringPerformance2026}. MIEB broadens image embedding evaluation across task categories and languages~\cite{xiaomieb}. MR$^2$-Bench instead emphasizes reasoning-intensive query-target relations~\cite{zhou2025mr2bench}. City-scale analysis depends on linking observations of the same place across platforms, spatial scales, and time, as well as connecting visual evidence with urban conditions. However, existing benchmarks emphasize general multimodal capabilities and do not jointly assess city-specific relations involving cross-view correspondence, geographic scale, temporal change, region grounding, and social economic indicators. GeoMEB fills this gap by standardizing these relations under a single instruction-conditioned ranking protocol.

\section{GeoMEB: A multimodal embedding benchmark for massive urban tasks}

\subsection{Benchmark overview}

\begin{table*}[!t]
\centering
\scriptsize
\setlength{\tabcolsep}{1.2pt}
\renewcommand{\arraystretch}{0.82}
\newcommand{\metataskcell}[2]{\parbox[c]{0.56in}{\raggedright\hspace{0pt}\textbf{\textit{#1}}\\[-1pt]\textbf{\textit{(#2 tasks)}}}}
\begin{tabular}{>{\columncolor{white}}m{0.60in}m{1.80in}m{0.54in}m{1.10in}m{1.18in}rrr}
\toprule
\textbf{Meta-task} & \textbf{Dataset (\#tasks)} & \textbf{Task type} & \textbf{Setting} & \textbf{Query $\rightarrow$ Target} & \textbf{\#Train} & \textbf{\#Queries} & \textbf{\#Candidates} \\
\midrule
\cellcolor{white} & RSICD~\cite{lu2018exploring} (2) & ICM & Remote sensing (RS) & image $\leftrightarrow$ text & 43,670 & 1,093$\sim$4,972 & 1,093$\sim$4,972 \\
\rowcolor{gray!12}\cellcolor{white} & UAV-GeoLoc~\cite{wu2025uavgeoloca} (2) & CG & Cross-view & image $\rightarrow$ image & 440,277 & 33,346$\sim$44,832 & 11,684$\sim$24,922 \\
\cellcolor{white} & VIGOR~\cite{zhu2021vigora} (1) & CG & Cross-view & image+text $\rightarrow$ image & 3,045 & 2,693 & 2,568 \\
\rowcolor{gray!12}\cellcolor{white} & SV Localization~\cite{li2026unified} (2) & ICM & Street view (SV) & image $\leftrightarrow$ text & 160,037 & 39,887 & 39,887 \\
\cellcolor{white} & Im2GPS3k~\cite{vo2017revisiting} (4) & CG; ICM & Street view & image $\leftrightarrow$ text & -- & 1,141$\sim$2,997 & 1,141$\sim$2,997 \\
\rowcolor{gray!12}\cellcolor{white}\multirow[c]{-6}{=}{\metataskcell{Retrieval}{13}} & VRSBench~\cite{li2024vrsbench} (2) & ICM & Remote sensing & image $\leftrightarrow$ text & 157,674 & 9,350 & 9,350 \\
\midrule
\cellcolor{white} & VRSBench~\cite{li2024vrsbench} (1) & MVU & Remote sensing & image+text $\rightarrow$ text & 157,674 & 37,409 & 2,373 \\
\rowcolor{gray!12}\cellcolor{white} & MME-RealWorld~\cite{zhang2025mmerealworld} (6) & MVU & Street view; Cross-view; Document; Remote sensing & image+text $\rightarrow$ text & 23,620 & 668$\sim$1,293 & 5 \\
\cellcolor{white}\multirow[c]{-3}{=}{\metataskcell{VQA}{13}} & UrBench~\cite{zhou2025urbench} (6) & USR; UER & Remote sensing; Street view; Cross-view & image+text $\rightarrow$ text & 9,274 & 50$\sim$225 & 2$\sim$900 \\
\midrule
\cellcolor{white} & Cityscapes~\cite{cordts2016cityscapes} (1) & SOG & Street view & image+text $\rightarrow$ image & 14,868 & 2,500 & 2,500 \\
\rowcolor{gray!12}\cellcolor{white} & Mapillary~\cite{ertler2020mapillary} (1) & SOG & Street view & image+text $\rightarrow$ image & 61,503 & 8,924 & 8,924 \\
\cellcolor{white}\multirow[c]{-3}{=}{\metataskcell{Grounding}{3}} & VRSBench~\cite{li2024vrsbench} (1) & ROG & Remote sensing & image+text $\rightarrow$ image & 157,674 & 16,159 & 16,159 \\
\midrule
\cellcolor{white} & LEVIR-CD~\cite{chen2020spatialtemporal} (1) & CMM & Remote sensing & image $\rightarrow$ image & 509 & 128 & 128 \\
\rowcolor{gray!12}\cellcolor{white} & SYSU-CD~\cite{shi2022deeply} (1) & CMM & Remote sensing & image+text $\rightarrow$ image & 16,000 & 4,000 & 4,000 \\
\cellcolor{white} & VL-CMU-CD~\cite{alcantarilla2018streetviewa} (1) & CMM & Street view & image+text $\rightarrow$ image & 3,732 & 429 & 429 \\
\rowcolor{gray!12}\cellcolor{white}\multirow[c]{-4}{=}{\metataskcell{Change\\Detection}{4}} & OSF Recovery~\cite{huang2025built} (1) & DDA & Street view & image $\rightarrow$ text & 190 & 137 & 4 \\
\midrule
\cellcolor{white} & AID~\cite{xia2017aid} (1) & LCC & Remote sensing & image $\rightarrow$ text & 8,000 & 2,000 & 30 \\
\rowcolor{gray!12}\cellcolor{white} & PlacePulse~\cite{salesses2012place} (6) & SBC & Street view & image $\rightarrow$ text & 40,640 & 654$\sim$2,947 & 14$\sim$18 \\
\cellcolor{white} & CityLens~\cite{liuCityLensEvaluatingLarge2025} (4) & SBC & Cross-view & image $\rightarrow$ text & 98,524 & 1,000 & 20 \\
\rowcolor{gray!12}\cellcolor{white}\multirow[c]{-4}{=}{\metataskcell{Classification}{12}} & UrBench~\cite{zhou2025urbench} (1) & LCC & Cross-view & image+text $\rightarrow$ text & 9,274 & 120 & 12 \\
\bottomrule
\end{tabular}
\caption{GeoMEB dataset and task statistics grouped by meta-task and dataset family. Arrows denote query-to-target modality direction, and $\leftrightarrow$ denotes bidirectional retrieval. Ranges give the minimum and maximum across task instances.}
\label{tab:GeoMEB_overview}
\end{table*}

\begin{figure}[t]
    \centering
    \includegraphics[width=\columnwidth]{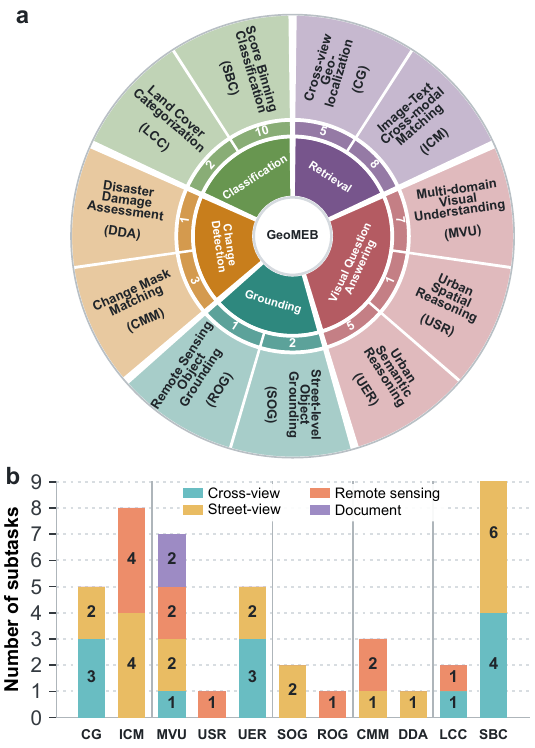}
    \caption{GeoMEB task taxonomy and input composition. (a) The 45 evaluation tasks grouped into five meta-tasks and 11 task types. (b) Their distribution across four modalities.}
    \label{fig:GeoMEB_task_overview}
\end{figure}

We introduce the \textbf{Geospatial Multimodal Embedding Benchmark (GeoMEB)},
which standardizes 45 evaluation tasks across five meta-tasks: retrieval,
visual question answering, change detection, classification, and visual
grounding. These tasks span 11 fine-grained task types and cover cross-view,
street-view, remote-sensing, and document inputs
(Figure~\ref{fig:GeoMEB_task_overview}). GeoMEB contains 286K
evaluation queries, while the associated training set provides
1.32 million examples. Table~\ref{tab:GeoMEB_overview} lists dataset
families, task settings, query-target modalities, and the
training, query, and candidate pools sizes.

\subsection{Benchmark dimensions}
\textbf{Task coverage.}
Tasks include matching street-view and satellite images, retrieving image captions, answering questions about scenes, grounding referred objects, detecting changes from image pairs, and selecting labels for urban perception and socioeconomic indicators. These cases are grouped by their query-target relation, spanning image-text matching, cross-view localization, spatial reasoning, temporal comparison, region grounding, and score-label choosing. Definitions of 11 task types are provided in Appendix~\ref{sec:detailed_GeoMEB}.

\noindent \textbf{Urban-view and temporal coverage.}
GeoMEB combines street-level imagery, unmanned aerial vehicle imagery, satellite imagery, and urban documents such as maps. Cross-view tasks connect ground-level scenes with imagery captured at different viewpoints and spatial scales, while change tasks compare observations of the same or related locations over time. GeoMEB therefore evaluates variation in viewpoint, geographic extent, sensor source, and observation time rather than treating urban imagery as a single visual domain.

\noindent \textbf{Modality and target diversity.}
Queries may contain text, single or multiple images, or interleaved image-text content. Targets may consist of captions, answer choices, class labels, geolocations, images, cropped regions, or change masks (see Appendix~\ref{sec:task_examples} for task-level examples). 

\subsection{Meta-task and evaluation design}
GeoMEB reformulates every task as ranking over a task-specific candidate set. For task $r$, a model encodes an instruction-conditioned query $q$ and each candidate $c\in\mathcal{C}_r$ into normalized embeddings $\mathbf{z}_q$ and $\mathbf{z}_c$, then ranks candidates by cosine similarity,
\begin{equation}
    \mathrm{sim}(q,c)=\mathbf{z}_q^{\top}\mathbf{z}_c,\qquad c\in\mathcal{C}_r .
\end{equation}


\paragraph{Task standardization.}
\textit{\textbf{Retrieval}} tasks rank candidate images or texts for a query image, text, or image-text input. \textit{\textbf{Visual Question Answering}} is converted into ranking over the answer vocabulary or multiple-choice options. \textit{\textbf{Classification}} uses textual class names or discretized score descriptions as candidates. \textit{\textbf{Grounding}} ranks candidate regions against an image-text query, while \textit{\textbf{Change Detection}} ranks masks or damage-status labels against multi-temporal observations.

\paragraph{Instruction and split standardization.}
Separate query and candidate instructions specify the relation being evaluated and the representation role of each input. We use the same task templates during training and evaluation. When official splits are available, we preserve them; otherwise, we construct random train/evaluation splits while ensuring that no target instances are shared across splits.
Task definitions, dataset transformations, and instruction templates are detailed in Appendices~\ref{sec:detailed_GeoMEB},
\ref{sec:dataset_details}, and~\ref{sec:task_examples}, respectively.

\section{Geo-Embed: Unified embedding model for massive urban tasks}
\label{sec:method}

\begin{figure}[H]
    \centering
    \includegraphics[width=\columnwidth]{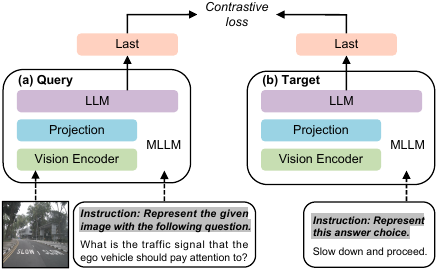}
    \caption{Architecture of Geo-Embed. The query and its target candidate are paired with side-specific instructions and independently encoded by the shared MLLM. Final-token hidden states from the last layer form the two representations, which are aligned through contrastive learning.}
    \label{fig:Geo-Embed_method}
\end{figure}

Using a shared vision-language backbone and side-specific instructions, Geo-Embed maps heterogeneous geospatial queries and candidates into a shared embedding space for task-aware matching. Figure~\ref{fig:Geo-Embed_method} illustrates its two-sided encoding architecture and contrastive training objective.

\subsection{Model architecture} 


We first define how the query and candidate inputs are represented. Let $r\in\mathcal{T}$ denote an urban task with natural-language instructions $p_r^{q}$ and $p_r^{c}$ for the query and candidate sides, respectively. We decompose a query as $q_i=(q_i^{\mathrm{vis}},q_i^{\mathrm{txt}})$ and a candidate as $c_j=(c_j^{\mathrm{vis}},c_j^{\mathrm{txt}})$, where either modality component may also be empty. Images are converted into visual tokens by the visual tokenizer of the backbone, while textual descriptions and structured urban attributes are serialized as text tokens. We construct the two input sequences as
\begin{equation}
\begin{array}{rcl}
    \widetilde{q}_i & = & [\mathrm{Vis}(q_i^{\mathrm{vis}});\,
    \mathtt{Instruct:}\ p_r^{q};\,\mathtt{Query:}\ q_i^{\mathrm{txt}}],\\
    \widetilde{c}_j & = & [\mathrm{Vis}(c_j^{\mathrm{vis}});\,
    \mathtt{Instruct:}\ p_r^{c};\,\mathtt{Candidate:}\ c_j^{\mathrm{txt}}].
\end{array}
\end{equation}
The same serialization covers text-only, image-only, image-text, and multi-image inputs by omitting empty components and preserving the order of the remaining tokens. Both sides are therefore instruction-conditioned: $p_r^{q}$ expresses the requested relation, whereas $p_r^{c}$ anchors the representation role of an image, region, caption, answer, or label. This separation allows the same raw content to receive different embeddings when it appears in different roles or tasks.

Geo-Embed uses a shared vision-language backbone $F_\theta$ to encode both sides. We take the last-layer hidden representation of the final token as the sequence representation:
\begin{equation}
    \mathbf{h}_{q_i}=F_\theta(\widetilde{q}_i)^{(L)}_{\mathrm{last}},
    \qquad
    \mathbf{h}_{c_j}=F_\theta(\widetilde{c}_j)^{(L)}_{\mathrm{last}}.
\end{equation}
We then apply $\ell_2$ normalization to obtain fixed-dimensional embeddings,
\begin{equation}
    \mathbf{z}_{q_i}=\frac{\mathbf{h}_{q_i}}{\|\mathbf{h}_{q_i}\|_2},
    \qquad
    \mathbf{z}_{c_j}=\frac{\mathbf{h}_{c_j}}{\|\mathbf{h}_{c_j}\|_2}.
\end{equation}

\begingroup
\setlength{\tabcolsep}{1.9pt}
\renewcommand{\arraystretch}{0.92}
\newcommand{\grouphead}[1]{{\scriptsize\textbf{#1}}}
\newcommand{\reshead}[1]{{\scriptsize\textbf{#1}}}
\newcommand{\tasknum}[1]{\multicolumn{1}{c}{#1}}
\begin{table*}[t]
\centering
\tiny
\resizebox{\textwidth}{!}{%
\begin{tabular}{lc!{\vrule width 0.6pt}rrr!{\vrule width 0.6pt}rrr!{\vrule width 0.6pt}rrrr!{\vrule width 0.6pt}rrr!{\vrule width 0.6pt}rrr!{\vrule width 0.6pt}>{\centering\arraybackslash}p{0.32in}}
\toprule
\multicolumn{2}{c!{\vrule width 0.6pt}}{} & \multicolumn{3}{c|}{\grouphead{Classification}} & \multicolumn{3}{c|}{\grouphead{Retrieval}} & \multicolumn{4}{c|}{\grouphead{VQA}} & \multicolumn{3}{c|}{\grouphead{Grounding}} & \multicolumn{3}{c|}{\grouphead{Change Detection}} & \multicolumn{1}{c}{\raisebox{-0.30\normalbaselineskip}{\makebox[0.38in][c]{\reshead{\scalebox{0.78}{Overall Avg}}}}} \\
\cmidrule(lr){3-5} \cmidrule(lr){6-8} \cmidrule(lr){9-12} \cmidrule(lr){13-15} \cmidrule(lr){16-18}
\textbf{Model} & \textbf{Size} & \reshead{LCC} & \reshead{SBC} & \reshead{Overall} & \reshead{ICM} & \reshead{CG} & \reshead{Overall} & \reshead{MVU} & \reshead{USR} & \reshead{UER} & \reshead{Overall} & \reshead{SOG} & \reshead{ROG} & \reshead{Overall} & \reshead{CMM} & \reshead{DDA} & \reshead{Overall} & \multicolumn{1}{c}{} \\
\textbf{\#Tasks} & \multicolumn{1}{c}{} & \tasknum{2} & \tasknum{10} & \tasknum{12} & \tasknum{8} & \tasknum{5} & \tasknum{13} & \tasknum{7} & \tasknum{1} & \tasknum{5} & \tasknum{13} & \tasknum{2} & \tasknum{1} & \tasknum{3} & \tasknum{3} & \tasknum{1} & \tasknum{4} & \tasknum{45} \\
\midrule
\rowcolor{gray!12}\multicolumn{19}{c}{\scriptsize\textbf{\textit{CLIP-family models}}} \\
CLIP~\cite{radfordLearningTransferableVisual2021} & 0.43B & 47.73 & 4.35 & 11.58 & 29.31 & 4.12 & 19.62 & 7.37 & 20.45 & 11.86 & 10.10 & 1.25 & 4.16 & 2.22 & 0.78 & -- & 0.58 & 8.82 \\
OpenCLIP~\cite{fang2023data} & 0.99B & 41.91 & 4.88 & 11.06 & 40.06 & 7.99 & 27.73 & 11.69 & 27.27 & 11.20 & 12.70 & 1.66 & 5.41 & 2.91 & 0.26 & -- & 0.20 & 10.92 \\
RemoteCLIP~\cite{liu2024remoteclipa} & 0.43B & 57.54 & 1.72 & 11.02 & 26.00 & 0.24 & 16.10 & 9.39 & 7.95 & 10.62 & 9.75 & 0.74 & 0.95 & 0.81 & 0.52 & -- & 0.39 & 7.61 \\
GeoRSCLIP~\cite{zhang2024rs5m} & 0.99B & 43.73 & 7.01 & 13.13 & 37.58 & 8.05 & 26.22 & 9.47 & 18.18 & 11.66 & 10.98 & 1.51 & 5.36 & 2.79 & 0.52 & -- & 0.39 & 10.70 \\
SkyScript-CLIP~\cite{wang2024skyscript} & 0.43B & 47.04 & 4.79 & 11.83 & 30.72 & 6.77 & 21.51 & 7.35 & 19.32 & 12.15 & 10.11 & 1.16 & 5.58 & 2.63 & 1.04 & -- & 0.78 & 9.37 \\
SigLIP2~\cite{tschannen2025siglip} & 0.88B & 36.30 & 6.09 & 11.13 & 31.94 & 3.62 & 21.05 & 6.25 & 20.45 & 16.13 & 11.14 & 2.44 & 5.08 & 3.32 & 0.26 & -- & 0.20 & 9.37 \\
\midrule
\rowcolor{gray!12}\multicolumn{19}{c}{\scriptsize\textbf{\textit{Instruction-tuned VLMs}}} \\
GeoChat~\cite{kuckreja2024geochat} & 7B & 15.54 & \textbf{8.91} & 10.02 & 0.95 & 0.37 & 0.73 & 11.90 & 27.27 & 11.75 & 13.03 & 0.03 & 0.53 & 0.19 & 0.26 & -- & 0.20 & 4.83 \\
Qwen3-VL-Instruct~\cite{bai2025qwen3vl} & 2B & 10.96 & 3.74 & 4.94 & 0.30 & 0.03 & 0.20 & 20.78 & 1.14 & 12.96 & 16.26 & 0.04 & 0.01 & 0.03 & 0.08 & \underline{41.61} & 10.46 & 6.38 \\
Qwen3-VL-Instruct~\cite{bai2025qwen3vl} & 8B & 6.38 & 4.96 & 5.19 & 0.59 & 0.18 & 0.43 & 21.54 & 15.91 & 13.30 & 17.94 & 0.05 & 0.59 & 0.23 & 0.35 & 33.58 & 8.65 & 6.49 \\
\midrule
\rowcolor{gray!12}\multicolumn{19}{c}{\scriptsize\textbf{\textit{Embedding-specialized VLMs}}} \\
UME-R1~\cite{lan2025umer1} & 7B & 50.98 & 5.71 & 13.25 & 31.11 & 4.98 & 21.06 & 48.26 & 22.73 & 26.58 & 37.96 & 4.55 & 4.17 & 4.43 & 0.19 & 37.96 & 9.63 & 17.27 \\
VLM2Vec-V2.0~\cite{mengVLM2VecV2AdvancingMultimodal2025} & 3B & 50.50 & 6.10 & 13.50 & 55.79 & 4.24 & 35.96 & 25.12 & 2.27 & 15.48 & 19.65 & 3.25 & 5.85 & 4.12 & 0.52 & 29.93 & 7.87 & 16.22 \\
OmniEmbed-Nemotron~\cite{xu2025omniembednemotrona} & 3B & 33.98 & 6.43 & 11.02 & 36.29 & 2.87 & 23.44 & 14.15 & 11.36 & 11.75 & 13.01 & 1.96 & 2.81 & 2.24 & 0.65 & 34.31 & 9.07 & 11.76 \\
Seed1.6-Embedding~\cite{bytedanceseed2025seed16embedding} & N/A & 56.48 & \underline{7.25} & \underline{15.46} & 56.70 & 6.82 & 37.52 & 42.27 & \underline{43.18} & 24.37 & 35.45 & 6.38 & \underline{7.68} & \underline{6.81} & \underline{1.81} & \textbf{42.34} & \textbf{11.94} & 21.44 \\
Qwen3-VL-Embedding~\cite{liQwen3VLEmbeddingQwen3VLRerankerUnified2026} & 8B & 54.12 & 3.99 & 12.35 & \underline{63.31} & \underline{8.76} & \underline{42.33} & \textbf{56.20} & \textbf{44.32} & \underline{30.07} & \textbf{45.23} & \underline{7.02} & 6.27 & 6.77 & 1.11 & 11.68 & 3.75 & \underline{22.09} \\
Qwen3-VL-Embedding~\cite{liQwen3VLEmbeddingQwen3VLRerankerUnified2026} & 2B & \underline{60.60} & 5.96 & 15.07 & 60.25 & 6.86 & 39.72 & \underline{50.79} & 23.86 & 26.20 & 39.26 & 4.99 & 3.56 & 4.52 & 0.63 & 35.77 & 9.42 & 21.60 \\
\rowcolor{lightlavender}Geo-Embed (Ours) & 2B & \textbf{66.75} & \underline{7.25} & \textbf{17.17} & \textbf{63.95} & \textbf{13.80} & \textbf{44.66} & 49.96 & 39.77 & \textbf{35.36} & \underline{43.56} & \textbf{9.13} & \textbf{13.00} & \textbf{10.42} & \textbf{2.80} & 37.96 & \underline{11.59} & \textbf{25.48} \\
\bottomrule
\end{tabular}}
\caption{Results on GeoMEB. Scores are task-count-weighted averages over fine-grained task types and meta-task groups. Best values are bolded and second-best values are underlined. ``--'' denotes unavailable results counted as 0 in overall averages.}
\label{tab:main_results}
\end{table*}
\endgroup

\subsection{Contrastive training} 

For each training instance from task $r$, we pair an instruction-conditioned query $(p_r^{q},q_i)$ with one relevant instruction-conditioned candidate $(p_r^{c},c_i^+)$. The remaining candidates in the same batch provide negative supervision and are likewise encoded with their assigned candidate instructions. Let $\mathcal{N}_i$ denote the in-batch negative set for $q_i$, consisting of candidates paired with other queries in the batch. Thus, each query is optimized against one positive and multiple negatives. 
We use cosine similarity with temperature scaling,
\begin{equation}
    s(q_i,c)=\frac{\mathrm{sim}(q_i,c)}{\tau},
\end{equation}
where $\tau$ is a temperature hyperparameter.
And then we train Geo-Embed with the InfoNCE objective~\cite{oord2019representation}
\begin{equation}
    \mathcal{L}_i = -\log
    \frac{\exp\left(s(q_i,c_i^+)\right)}
    {\exp\left(s(q_i,c_i^+)\right)
    +\displaystyle\sum_{c_j^-\in\mathcal{N}_i}
    \exp\left(s(q_i,c_j^-)\right)}.
\end{equation}
For batch $\mathcal{B}$, the final objective averages the per-query
losses:
\begin{equation}
    \mathcal{L}_{\mathrm{con}}=
    \frac{1}{|\mathcal{B}|}
    \sum_{i\in\mathcal{B}}\mathcal{L}_i.
\end{equation}
This objective increases the similarity of each matched query-candidate
pair relative to its in-batch negatives without encoding additional
candidates.

\section{Experiments and Results}

\subsection{Experiment setting}

Geo-Embed is initialized from Qwen3-VL-Embedding model and trained with the above instruction-conditioned contrastive objective using LoRA rank 16 and a batch size of 1024. All tasks are serialized into a unified query-target ranking format. Detailed hyperparameters, hardware, and implementation details are provided in Appendix~\ref{sec:training_eval_details}.

\noindent \textbf{Baselines.}
We compare with three representative baseline model groups: CLIP-family models, instruction-tuned VLMs, and embedding-specialized VLMs.

\noindent \textbf{Metrics.}
All tasks are evaluated as ranking over task-specific candidate pools. Image-text cross-modal retrieval tasks (ICM) use the mean value of Recall@1, @5, and @10 metrics; other tasks use the Recall@1 metric. We report weighted averages for fine-grained task types and meta-task groups in Table~\ref{tab:main_results}, with detailed results in Appendix~\ref{sec:detailed_results}.

\subsection{Main results}

\begin{figure}[!htbp]
    \centering
    \includegraphics[width=\columnwidth]{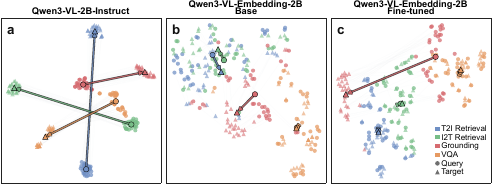}
    \caption{Feature space visualization: (a) Qwen3-VL-Instruct-2B, (b) Qwen3-VL-Embedding-2B, and (c) Qwen3-VL-Embedding-2B after VRSBench fine-tuning. Circles and triangles denote query and target embeddings, respectively.}
    \label{fig:tsne_feature_space}
\end{figure}

\noindent\textbf{Geo-Embed achieves the strongest overall performance on GeoMEB.}
Geo-Embed achieves the highest overall average score of 25.48, outperforming Qwen3-VL-Embedding-8B (22.09), Qwen3-VL-Embedding-2B (21.60), and Seed1.6-Embedding (21.44). Geo-Embed ranks first on classification, retrieval, and grounding, and remains second-best on both VQA and change detection. 
These gains indicate that urban contrastive fine-tuning improves
task-specific ranking performance across heterogeneous query-target relations. Figure~\ref{fig:tsne_feature_space} further illustrates this geometric shift: Qwen3-VL-Instruct-2B forms highly compact clusters because it is not designed for retrieval, making nearest-neighbor retrieval difficult; Qwen3-VL-Embedding-2B produces a more dispersed retrieval-oriented space; and fine-tuning brings most query-target pairs closer, although some tasks remain imperfectly aligned after training.

\noindent\textbf{Generic multimodal ability does not guarantee strong urban retrieval.}
CLIP-family models retain reasonable performance on some classic image-text and classification tasks, but their scores drop sharply on VQA, grounding, and change detection, where inputs are often interleaved, instruction-dependent, or region- and mask-oriented. Instruction-tuned VLMs show a different limitation: Qwen3-VL-Instruct and GeoChat can follow multimodal prompts, but their overall embedding scores remain low, indicating that instruction-following ability does not necessarily produce a suitable embedding space for retrieval. Even strong generic embedders are uneven across task types: Qwen3-VL-Embedding-8B achieves the best on VQA, and Seed1.6-Embedding is strongest on change detection. These results show that generic multimodal alignment alone is insufficient for GeoMEB, where models must support cross-view matching, urban semantics, region grounding, temporal comparison, and answer selection within one embedding interface.

\noindent\textbf{Fine-grained results reveal urban tasks remain unsolved.}
Across models, ICM and LCC are the most mature tasks, corresponding to image-caption matching and land-cover or scene-label recognition. Several embedding-specialized models performed well on these tasks, and Geo-Embed further improves ICM to 63.95 and LCC to 66.75, suggesting that current embedders can already handle semantic correspondence when query and target share relatively direct visual or textual cues. VQA also benefits from embedding-specialized backbones, with Qwen3-VL-Embedding-8B and Geo-Embed reaching 45.23 and 43.56 overall. In contrast, CG, SBC, SOG/ROG, and CMM remain much harder, covering cross-view geo-localization, urban perception indicator estimation, region grounding, and mask-level change matching. These gaps suggest that current multimodal embedders are useful for coarse semantic matching, but still struggle to align geospatial viewpoints, infer abstract city attributes, localize region-level targets, and compare fine-grained temporal changes.These task types therefore define concrete stress tests for future geospatial embedding models.

\section{Ablation analysis}

\noindent \textbf{Effect of interleaved sub-batch sampling.}
We study how interleaved sub-batching affects contrastive fine-tuning. During model training, Sub0 denotes fully random batching without sub-batch grouping, while larger sub-batch sizes divide each batch into homogeneous groups sampled by dataset source or input modality. Sub0 maximizes source diversity but provide weaker local competition, whereas larger same-group sub-batches create harder within-group comparisons but reduce the number of groups represented in each batch.

Table~\ref{tab:subbatch_sampling} shows that the effect depends on the grouping rule. Under dataset grouping, Sub0 obtains the best overall score of 18.07, while the best grouped setting Sub128 reaches 17.98. Although Sub128 improves retrieval and VQA tasks, and Sub32 slightly improves change detection tasks, these gains do not turn into a better overall average. In contrast, modality grouping shows a clearer positive pattern: Sub256 shows the best RS, SV, and document scores, while cross-view performance is highest under Sub0 and changes only slightly across grouped settings. These results suggest that modality-aware grouping can improve within-modality discrimination, but interleaved sub-batching does not consistently outperform random batching across the full GeoMEB task mix.

\begin{table}[!t]
    \centering
\small
\setlength{\tabcolsep}{2.7pt}
\renewcommand{\arraystretch}{0.90}
\caption{Effect of interleaved sub-batch size under two grouping rules. Upper block reports task-group average scores, and lower block reports modality-group average scores.}
\label{tab:subbatch_sampling}
\begin{tabular}{lccccc}
\toprule
Group & Sub0 & Sub32 & Sub64 & Sub128 & Sub256 \\
\midrule
\multicolumn{6}{l}{\textit{\textbf{Dataset source grouping}}} \\
Class. & \textbf{19.12} & 18.77 & 18.02 & 19.11 & 18.73 \\
Retr. & 31.42 & 31.53 & 31.24 & \textbf{31.68} & 31.40 \\
VQA & 23.33 & 22.95 & 23.38 & \textbf{24.02} & 22.69 \\
Ground. & \textbf{7.17} & 6.03 & 5.33 & 6.03 & 6.37 \\
ChgDet & 9.34 & \textbf{9.42} & 9.21 & 9.07 & 9.12 \\
\cmidrule(lr){1-6}
\rowcolor{gray!10}\textbf{Overall} & \textbf{18.07} & 17.74 & 17.44 & 17.98 & 17.66 \\
\midrule
\multicolumn{6}{l}{\textit{\textbf{Modality grouping}}} \\
RS & 21.32 & 21.73 & 21.52 & 21.30 & \textbf{22.17} \\
SV & 27.59 & 27.33 & 28.18 & 27.84 & \textbf{28.49} \\
Cross-view & \textbf{29.78} & 29.67 & 29.15 & 29.26 & 29.43 \\
Document & 41.05 & 40.65 & 40.90 & 40.50 & \textbf{41.10} \\
\bottomrule
\end{tabular}
\end{table}

\begin{figure}[!t]
    \centering
    \includegraphics[width=\columnwidth]{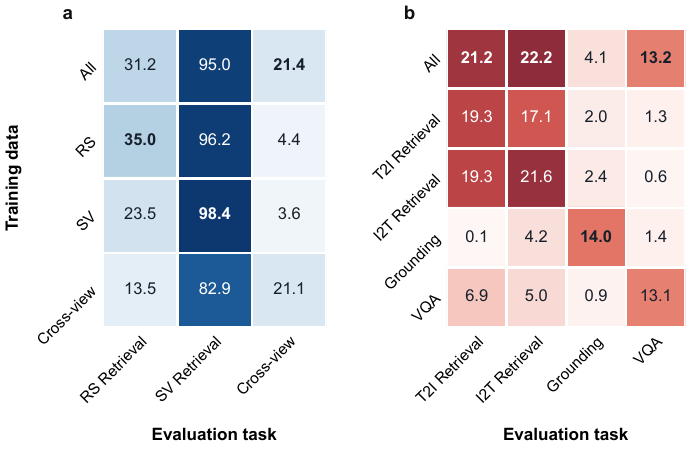}
    \caption{Performance comparison of models trained with different data subsets. (a) compares RS-only, SV-only, cross-view-only, and all-source training for retrieval evaluations; (b) compares retrieval-only, grounding-only, VQA-only, and all-source training across VRSBench task families.}
    \label{fig:ablation_heatmaps}
\end{figure}

\noindent \textbf{Task-modality interactions reveal compatibility and interference.}
We next examine how training-data modality affects the cross-task generalization. The training groups use comparable subsets from RS captioning (VRSBench, \(\sim\)40K), SV localization (CityLens, \(\sim\)40K), cross-view matching (VIGOR, \(\sim\)42K), and their all-source combination (\(\sim\)122K). Figure~\ref{fig:ablation_heatmaps}a shows clear modality-level interactions. RS-only training gives the best RS retrieval score of 35.0, and SV-only training gives the best SV retrieval score of 98.4, both exceeding the all-source setting on their closest evaluation tasks. In contrast, cross-view evaluation requires cross-view or all-source supervision: RS-only and SV-only training give very low cross-view scores, while the all-source setting reaches 21.4. This suggests that single-modality supervision can sharpen within-modality retrieval, but cross-view matching depends on explicit cross-view alignment signals.

On VRSBench, Figure~\ref{fig:ablation_heatmaps}b shows a similar interaction across task sources. All-source training gives the best T2I and I2T retrieval scores, reaching 21.2 and 22.2, and remains competitive on VQA with a score of 13.2. However, grounding behaves differently: grounding-only training reaches 14.0, whereas all-source training drops substantially. This indicates that multi-task supervision is not automatically additive: compatible tasks can reinforce shared semantic alignment, whereas structurally different tasks may interfere with specialized query-target relations. This observation is consistent with the feature space visualization in Figure~\ref{fig:tsne_feature_space}: fine-tuning improves several query-target neighborhoods, but not all task relations become more compact, as shown by the enlarged VQA query-target separation.

\noindent \textbf{Instruction-induced shifts are not uniformly target-oriented.}
We further test whether task instructions move query embeddings closer to their matched targets in the original embedding space. We annotate 50 aligned VIGOR triplets, each containing a street-view image, a remote-sensing image, and a neutral cross-view description.
For a directed source-target modality pair \(m_s \rightarrow m_t\), let
\(\mathbf{z}^{\,\mathrm{ins}}_{q_i}\) and
\(\mathbf{z}^{\,\mathrm{noins}}_{q_i}\) denote the normalized query embeddings of example \(i\) encoded with the task-specific instruction and without an instruction, respectively (see Appendix~\ref{sec:task_examples} for details). Let
\(\mathbf{z}^{\,\mathrm{}}_{t_i}\) denote the matched target embedding encoded with the modality-default target instruction. We define the direction-level instruction-induced alignment shift as
\begin{equation}
\Delta_{\mathrm{cos}}(m_s \!\rightarrow\! m_t)
=
\frac{1}{N}
\sum_{i=1}^{N}
\left[
(\mathbf{z}^{\,\mathrm{ins}}_{q_i})^\top \mathbf{z}^{\,\mathrm{}}_{t_i}
-
(\mathbf{z}^{\,\mathrm{noins}}_{q_i})^\top \mathbf{z}^{\,\mathrm{}}_{t_i}
\right].
\end{equation}
Positive values indicate that the task instruction moves the query closer to its matched target, whereas negative values indicate that the instruction moves it farther away.

\begin{figure}[!h]
    \centering
    \includegraphics[width=\columnwidth]{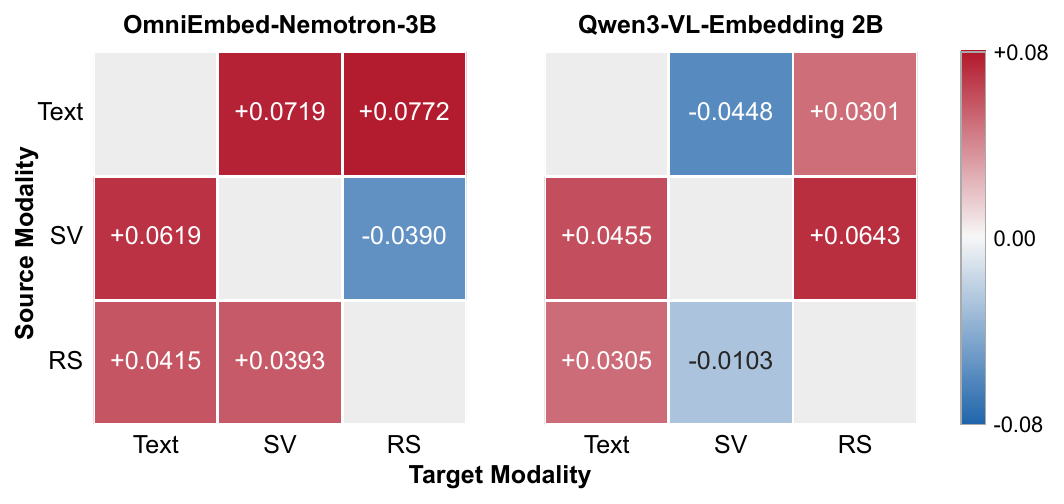}
    \caption{Instruction-induced alignment shifts, reporting the mean change in matched query-target cosine similarity when queries are encoded with or without task-specific instruction.}
    \label{fig:instruction_shift_heatmap}
\end{figure}


Figure~\ref{fig:instruction_shift_heatmap} shows that instruction-induced shifts do not translate into uniformly target-oriented movement. OmniEmbed-Nemotron-3B exhibits broadly positive shifts, with five of the six directed cross-modal settings moving closer to the matched target and the largest improvements appearing in text-to-visual directions, such as Text$\rightarrow$RS (+0.0772) and Text$\rightarrow$SV (+0.0719). However, this behavior is still asymmetric: SV$\rightarrow$RS decreases alignment (-0.0390), even though the reverse direction RS$\rightarrow$SV improves (+0.0393). Qwen3-VL-Embedding-2B shows the same direction-dependent pattern in a different form. Spatially grounded instructions improve several directions, especially SV$\rightarrow$RS (+0.0643) and SV$\rightarrow$Text (+0.0455), but still reduce alignment for Text$\rightarrow$SV (-0.0448) and RS$\rightarrow$SV (-0.0103). These results suggest that instruction-induced shifts interact unevenly with the pretrained embedding geometry rather than establishing a consistent bidirectional alignment between modality pairs. Future geospatial embedders should therefore not only respond to task instructions, but also align the induced movement with the intended spatial relation between query and target.

\noindent \textbf{Hyperparameter sensitivity.}
We identify two key parameters that affect Geo-Embed performance: LoRA rank and batch size. Experiments use a random 10\% GeoMEB subset. Table~\ref{tab:hyperparameter_sensitivity} shows that LoRA rank has a non-monotonic effect: rank 16 gives the best overall average, rank 4 remains competitive and performs slightly better on classification and retrieval, while rank 32 degrades the average score. Increasing the batch size from 64 to 1024 further improves the overall score from 18.30 to 18.80, with the clearest gains on grounding and change detection. We therefore use LoRA rank 16 and batch size 1024 as the default setting.

\begin{table}[!htbp]
    \centering
\small
\setlength{\tabcolsep}{2.0pt}
\renewcommand{\arraystretch}{0.92}
\caption{Performance comparison of models trained with different LoRA ranks and batch sizes. Scores are reported by meta-task and overall averages.}
\label{tab:hyperparameter_sensitivity}
\begin{tabular}{lcccccc}
\toprule
Setting & Class. & Retr. & VQA & Ground. & ChgDet & Overall \\
\midrule
\multicolumn{7}{l}{\textit{\textbf{LoRA rank comparison} (batch size = 64)}} \\
LoRA rank = 4  & \textbf{15.28} & \textbf{39.58} & 20.78 & 6.30 & 9.40 & 18.27 \\
LoRA rank = 8  & 13.91 & 39.32 & 20.76 & 6.40 & \textbf{9.59} & 18.00 \\
LoRA rank = 16 & 14.64 & 39.35 & \textbf{20.99} & \textbf{7.17} & 9.34 & \textbf{18.30} \\
LoRA rank = 32 & 13.93 & 39.33 & 20.71 & 5.67 & 9.43 & 17.81 \\
\midrule
\multicolumn{7}{l}{\textit{\textbf{Batch size comparison} (LoRA rank = 16)}} \\
Batch size = 64   & 14.64 & 39.35 & 20.99 & 7.17 & 9.34 & 18.30 \\
Batch size = 256  & 14.85 & 39.78 & 21.37 & 6.90 & 9.50 & 18.48 \\
Batch size = 512  & 13.19 & \textbf{40.07} & \textbf{21.77} & 8.50 & 8.72 & 18.45 \\
Batch size = 1024 & 14.92 & 39.94 & 20.26 & \textbf{8.87} & \textbf{10.02} & \textbf{18.80} \\
\bottomrule
\end{tabular}
\end{table}

\section{Conclusion}

This work reframes geospatial multimodal embedding as relation-centered
query-target matching across heterogeneous urban evidence. We introduced
GeoMEB, a 45-task benchmark that standardizes semantic matching, cross-view
alignment, region-level grounding, and temporal change correspondence under a
unified instruction-conditioned ranking protocol, and Geo-Embed, a unified
embedding model trained with task-aware contrastive supervision. Experiments
show that Geo-Embed achieves the strongest overall performance, while
fine-grained analyses reveal that geospatial embedding quality depends strongly
on the target relation. These results suggest that future geospatial embedders
should make query-target relations explicit in training, sampling,
instruction design, and evaluation. GeoMEB provides a foundation for building
geospatial embedding models that are comparable across tasks and reusable across
city-scale applications.
\bibliography{aaai2027}

\clearpage
\appendix

\section*{Appendix}
\addcontentsline{toc}{section}{Appendix}

\setcounter{table}{0}
\setcounter{figure}{0}

\section{Task definitions}
\label{sec:detailed_GeoMEB}
GeoMEB contains 45 evaluation subtasks spanning retrieval, visual question answering, change detection, classification, and visual grounding. The 11 task types are cross-view geo-localization (CG), image-text cross-modal matching (ICM), score binning classification (SBC), multi-domain visual understanding (MVU), urban spatial reasoning (USR), urban semantic reasoning (UER), street-level object grounding (SOG), remote-sensing object grounding (ROG), change mask matching (CMM), disaster damage assessment (DDA), and land-cover categorization (LCC). Complete GeoMEB inventory is summarized in Appendix Table~\ref{tab:GeoMEB_overview}.

GeoMEB converts each task into instruction-conditioned ranking over a task-specific target pool. The taxonomy contains five meta-tasks and eleven fine-grained task types:

\begin{description}
    \item[\textbf{Retrieval.}]
    Retrieval tasks ask the model to identify the matching item for a query from a large candidate pool. The query and target may use the same modality or different modalities, such as image-to-text (i2t) retrieval, text-to-image (t2i) retrieval, UAV-to-satellite matching, or street-view-to-satellite geo-localization. These tasks evaluate whether the embedding space preserves semantic and geospatial correspondence across viewpoint, scale, and representation changes.
    \begin{itemize}
        \item \textbf{Cross-view geo-localization (CG):} matches observations from different geographic views or representations, such as street-level images, satellite images, and location descriptions. It emphasizes viewpoint invariance and spatial correspondence.
        \item \textbf{Image-text cross-modal matching (ICM):} aligns images with textual descriptions in either direction. It emphasizes semantic compatibility between visual content and language.
    \end{itemize}

    \item[\textbf{Visual question answering.}]
    VQA tasks start from an image-conditioned question and are reformulated as answer retrieval. Each query contains visual evidence and a natural-language question, while the target pool contains candidate answers from the original answer vocabulary or multiple-choice set. 
    This formulation tests whether an embedding model can represent not only visual content, but also the spatial, semantic, and contextual relation requested by the question.
    \begin{itemize}
        \item \textbf{Multi-domain visual understanding (MVU):} covers general visual understanding across remote sensing, street view, documents, and mixed urban scenes.
        \item \textbf{Urban spatial reasoning (USR):} focuses on spatial quantities and relations, such as counting, localization, and spatial comparison.
        \item \textbf{Urban semantic reasoning (UER):} covers scene-level semantic judgments, including object attributes, road understanding, role-based reasoning, traffic-sign interpretation, and visual-prompt reasoning.
    \end{itemize}

    \item[\textbf{Grounding.}]
    Grounding tasks require localizing the visual region described by an image-text query. Instead of generating bounding boxes or masks, we cast grounding as ranking candidate crops or regions and selecting the one that matches the referring expression. This setting evaluates fine-grained region-level alignment, where the correct target may depend on object attributes, spatial relations, and scene context.
    \begin{itemize}
        \item \textbf{Street-level object grounding (SOG):} grounds textual references in street-view imagery, where objects appear at human scale and under perspective distortion.
        \item \textbf{Remote-sensing object grounding (ROG):} grounds textual references in overhead imagery, where targets are often smaller, denser, and more scale-dependent.
    \end{itemize}

    \item[\textbf{Change detection.}]
    Change detection tasks compare observations of the same or related place across time. The query contain bi-temporal or multi-temporal imagery, sometimes with task text, and the candidate can be a change mask or a textual damage or recovery label. These tasks test whether embeddings capture temporal evidence and physical change rather than only static scene similarity.
    \begin{itemize}
        \item \textbf{Change mask matching (CMM):} matches image pairs to change masks or visual change targets, emphasizing differences rather than static scene content.
        \item \textbf{Disaster damage assessment (DDA):} maps multi-temporal observations to damage or recovery-status labels, testing semantic interpretation of physical change.
    \end{itemize}

    \item[\textbf{Classification.}]
    Classification tasks convert recognition or scoring problems into ranking over class candidates. For scene recognition, candidates are class names; for urban indicators or perception scores, candidates are discretized score labels. This design tests whether the embedding space can associate visual urban evidence with semantic categories and ordinal descriptions.
    \begin{itemize}
        \item \textbf{Land-cover categorization (LCC):} maps remote-sensing or cross-view visual evidence to land-cover or scene categories.
        \item \textbf{Score binning classification (SBC):} converts continuous or ordinal urban indicators into discrete textual targets, including perceived urban qualities and socioeconomic or built-environment indicators.
    \end{itemize}
\end{description}

\section{Dataset descriptions}
\label{sec:dataset_details}

This section provides dataset descriptions for the GeoMEB inventory. Below we describe the original dataset sources and how each is used for specfic tasks.

\paragraph{AID.}
AID~\cite{xia2017aid} is a dataset for aerial scene classification, designed to evaluate recognition over high-resolution remote-sensing scenes with diverse land-use and land-cover categories. GeoMEB uses AID for scene classification by converting the original 30-way recognition problem into image-to-text retrieval, where each image is matched against a candidate pool of scene-category labels.

\paragraph{RSICD.}
RSICD~\cite{lu2018exploring} is a remote-sensing image captioning dataset containing aerial images with sentence-level descriptions. GeoMEB uses these image-caption pairs for bidirectional image-text cross-modal matching: image-to-text retrieval treats the remote-sensing image as the query and captions as candidates, while text-to-image retrieval uses the caption as the query and retrieves the matching image.

\begin{table*}[!t]
\centering
\scriptsize
\setlength{\tabcolsep}{1.2pt}
\renewcommand{\arraystretch}{0.82}

\label{tab:GeoMEB_details}
\begin{tabular}{p{0.78in}p{1.00in}p{0.60in}p{0.62in}p{1.08in}rrr}
\toprule
\textbf{Dataset} & \textbf{Subtask} & \textbf{Task type} & \textbf{Setting} & \textbf{Query $\rightarrow$ Target} & \textbf{\#Train} & \textbf{\#Queries} & \textbf{\#Candidates} \\
\midrule
\multicolumn{8}{c}{\textbf{Retrieval (13 tasks)}} \\
\midrule
RSICD & i2t & ICM & RS & image $\rightarrow$ text & 43,670 & 1,093 & 4,972 \\
\rowcolor{gray!12} RSICD & t2i & ICM & RS & text $\rightarrow$ image & 43,670 & 4,972 & 1,093 \\
UAV-GeoLoc & country & CG & Cross-view & image $\rightarrow$ image & 172,237 & 44,832 & 24,922 \\
\rowcolor{gray!12} UAV-GeoLoc & terrain & CG & Cross-view & image $\rightarrow$ image & 268,040 & 33,346 & 11,684 \\
VIGOR & -- & CG & Cross-view & image+text $\rightarrow$ image & 3,045 & 2,693 & 2,568 \\
\rowcolor{gray!12} SV Localization & i2t & ICM & SV & image $\rightarrow$ text & 160,037 & 39,887 & 39,887 \\
SV Localization & t2i & ICM & SV & text $\rightarrow$ image & 160,037 & 39,887 & 39,887 \\
\rowcolor{gray!12} Im2GPS3k & GPS i2t & CG & SV & image $\rightarrow$ text & -- & 2,997 & 1,141 \\
Im2GPS3k & GPS t2i & CG & SV & text $\rightarrow$ image & -- & 1,141 & 2,997 \\
\rowcolor{gray!12} Im2GPS3k & Caption i2t & ICM & SV & image $\rightarrow$ text & -- & 1,805 & 1,805 \\
Im2GPS3k & Caption t2i & ICM & SV & text $\rightarrow$ image & -- & 1,805 & 1,805 \\
\rowcolor{gray!12} VRSBench & caption i2t & ICM & RS & image $\rightarrow$ text & 157,674 & 9,350 & 9,350 \\
VRSBench & caption t2i & ICM & RS & text $\rightarrow$ image & 157,674 & 9,350 & 9,350 \\
\midrule
\multicolumn{8}{c}{\textbf{Visual Question Answering (13 tasks)}} \\
\midrule
VRSBench & vqa & MVU & RS & image+text $\rightarrow$ text & 157,674 & 37,409 & 2,373 \\
\rowcolor{gray!12} MME-RealWorld & AutonomousDriving & MVU & SV & image+text $\rightarrow$ text & 5,211 & 1,293 & 5 \\
MME-RealWorld & MME-HD-CN & MVU & Cross-view & image+text $\rightarrow$ text & 2,635 & 684 & 5 \\
\rowcolor{gray!12} MME-RealWorld & diagram\_and\_table & MVU & Document & image+text $\rightarrow$ text & 4,766 & 1,167 & 5 \\
MME-RealWorld & monitoring\_images & MVU & SV & image+text $\rightarrow$ text & 2,824 & 668 & 5 \\
\rowcolor{gray!12} MME-RealWorld & ocr\_cc & MVU & Document & image+text $\rightarrow$ text & 5,004 & 1,236 & 5 \\
MME-RealWorld & remote\_sensing & MVU & RS & image+text $\rightarrow$ text & 3,180 & 858 & 5 \\
\rowcolor{gray!12} UrBench & counting & USR & RS & image+text $\rightarrow$ text & 351 & 88 & 7 \\
UrBench & object-attribute-recognition oe & UER & Cross-view & image+text $\rightarrow$ text & 340 & 85 & 15 \\
\rowcolor{gray!12} UrBench & road-understanding oe & UER & SV & image+text $\rightarrow$ text & 201 & 51 & 2 \\
UrBench & role-based-reasoning & UER & Cross-view & image+text $\rightarrow$ text & 898 & 225 & 900 \\
\rowcolor{gray!12} UrBench & traffic-sign-reasoning & UER & SV & image+text $\rightarrow$ text & 400 & 100 & 400 \\
UrBench & visual-prompt-reasoning & UER & Cross-view & image+text $\rightarrow$ text & 199 & 50 & 200 \\
\midrule
\multicolumn{8}{c}{\textbf{Grounding (3 tasks)}} \\
\midrule
Cityscapes & -- & SOG & SV & image+text $\rightarrow$ image & 14,868 & 2,500 & 2,500 \\
\rowcolor{gray!12} Mapillary & -- & SOG & SV & image+text $\rightarrow$ image & 61,503 & 8,924 & 8,924 \\
VRSBench & referring & ROG & RS & image+text $\rightarrow$ image & 157,674 & 16,159 & 16,159 \\
\midrule
\multicolumn{8}{c}{\textbf{Change Detection (4 tasks)}} \\
\midrule
LEVIR-CD & -- & CMM & RS & image $\rightarrow$ image & 509 & 128 & 128 \\
\rowcolor{gray!12} SYSU-CD & -- & CMM & RS & image+text $\rightarrow$ image & 16,000 & 4,000 & 4,000 \\
VL-CMU-CD & -- & CMM & SV & image+text $\rightarrow$ image & 3,732 & 429 & 429 \\
\rowcolor{gray!12} OSF Recovery & -- & DDA & SV & image $\rightarrow$ text & 190 & 137 & 4 \\
\midrule
\multicolumn{8}{c}{\textbf{Classification (12 tasks)}} \\
\midrule
AID & scene classification & LCC & RS & image $\rightarrow$ text & 8,000 & 2,000 & 30 \\
\rowcolor{gray!12} PlacePulse & safety & SBC & SV & image $\rightarrow$ text & 10,995 & 2,754 & 14 \\
PlacePulse & lively & SBC & SV & image $\rightarrow$ text & 11,765 & 2,947 & 16 \\
\rowcolor{gray!12} PlacePulse & beautiful & SBC & SV & image $\rightarrow$ text & 6,052 & 1,520 & 16 \\
PlacePulse & wealthy & SBC & SV & image $\rightarrow$ text & 5,462 & 1,372 & 18 \\
\rowcolor{gray!12} PlacePulse & depressing & SBC & SV & image $\rightarrow$ text & 3,780 & 953 & 18 \\
PlacePulse & boring & SBC & SV & image $\rightarrow$ text & 2,586 & 654 & 18 \\
\rowcolor{gray!12} CityLens & gdp & SBC & Cross-view & image $\rightarrow$ text & 6,556 & 1,000 & 20 \\
CityLens & pop & SBC & Cross-view & image $\rightarrow$ text & 56,560 & 1,000 & 20 \\
\rowcolor{gray!12} CityLens & height & SBC & Cross-view & image $\rightarrow$ text & 7,056 & 1,000 & 20 \\
CityLens & health & SBC & Cross-view & image $\rightarrow$ text & 28,352 & 1,000 & 20 \\
\rowcolor{gray!12} UrBench & scene-recognition & LCC & Cross-view & image+text $\rightarrow$ text & 480 & 120 & 12 \\
\midrule
\multicolumn{8}{c}{\textbf{Training-only Collection (1 item)}} \\
\midrule
MP16Pro & -- & -- & Training-only & -- & 200,000 & -- & -- \\
\bottomrule
\end{tabular}
\caption{Detailed GeoMEB task-level statistics.}
\end{table*}

\paragraph{UAV-GeoLoc.}
UAV-GeoLoc~\cite{wu2025uavgeoloca} targets UAV geo-localization, where aerial views captured from UAV platforms are matched to geo-tagged satellite references under viewpoint, scale, and geometric discrepancies. GeoMEB uses its country-level and terrain-level variants as two cross-view retrieval subtasks; in both cases, the query side contains the UAV view and the target side contains satellite candidates, testing whether an embedding model can bridge aerial-to-satellite appearance differences.

\paragraph{VIGOR.}
VIGOR~\cite{zhu2021vigora} is a cross-view geo-localization dataset designed to move beyond the idealized one-to-one matching assumption between street-view and satellite images. GeoMEB evaluates street-view-to-satellite retrieval on VIGOR, where each query combines street-level visual evidence with text context and the model retrieves the matching satellite image from a candidate pool, making the task a direct test of cross-view urban alignment.

\paragraph{CityLens.}
CityLens~\cite{liuCityLensEvaluatingLarge2025} benchmarks vision-language models for urban socioeconomic sensing by linking satellite and street-view observations with region-level urban indicators. GeoMEB uses four indicator-prediction subtasks---GDP, population density, average building height, and healthcare accessibility---where a query combines satellite imagery and multiple street-view images, and the target is a discretized normalized score label.

\paragraph{SV Localization.}
SV Localization is derived from the street-view data used in Unified Urban Tuning~\cite{li2026unified}. It contains street-level imagery from both Google Street View (GSV) and Baidu Street View (BSV), covering 241 cities while each city provides roughly 850 images.


\paragraph{Im2GPS3k.}
Im2GPS3k~\cite{vo2017revisiting} is a classic image geolocalization dataset with geographic metadata and brief latitude-longitude descriptions. GeoMEB uses it for zero-shot street-view geolocation via bidirectional GPS-region retrieval. We further annotate each image with a detailed scene description to construct bidirectional caption retrieval.

\paragraph{VRSBench.}
VRSBench~\cite{li2024vrsbench} is a versatile remote-sensing vision-language dataset that combines detailed captions, object references, and question-answer pairs for image understanding. GeoMEB uses its caption retrieval, VQA, and referring-expression grounding components, covering image-text retrieval, image-question-to-answer matching, and image-expression-to-region retrieval within the same remote-sensing source.

\paragraph{UrBench.}
UrBench~\cite{zhou2025urbench} is a dataset for evaluating large multimodal models in multi-view urban scenarios, with tasks designed around urban perception, spatial reasoning, and semantic understanding. GeoMEB uses its counting, object-attribute recognition, road understanding, role-based reasoning, traffic-sign reasoning, visual-prompt reasoning, and scene-recognition subtasks, covering VQA-style reasoning and classification under standardized ranking.

\paragraph{MME-RealWorld.}
MME-RealWorld~\cite{zhang2025mmerealworld} evaluates multimodal models on high-resolution real-world scenarios that are difficult even for humans, including visually dense and domain-diverse image-question settings. GeoMEB uses its autonomous driving, high-resolution Chinese imagery, diagram-and-table, monitoring-image, OCR-centered, and remote-sensing subsets as VQA tasks. Each task is converted into image-plus-question to answer-choice retrieval, with a small candidate pool of answer options.

\paragraph{Cityscapes.}
Cityscapes~\cite{cordts2016cityscapes} is a large-scale urban street-scene dataset for semantic and instance-level understanding, collected from stereo video sequences across diverse city environments. GeoMEB uses Cityscapes for street-level object grounding: the query consists of a street-view image and a referring expression, and the candidate pool contains cropped regions, requiring the model to retrieve the region that matches the described object.

\paragraph{Mapillary.}
The Mapillary Traffic Sign Dataset~\cite{ertler2020mapillary} is a large-scale street-level traffic-sign dataset for detection and classification across diverse geographic, weather, and lighting conditions. GeoMEB uses Mapillary as a street-level grounding source focused on traffic-sign and street-scene regions. As with Cityscapes, GeoMEB evaluates image-plus-text queries against cropped image-region candidates.

\paragraph{LEVIR-CD.}
LEVIR-CD~\cite{chen2020spatialtemporal} is a remote-sensing change-detection dataset built from bi-temporal image pairs and designed to evaluate whether models can identify meaningful land-surface changes despite illumination and registration variation. GeoMEB reformulates it as change-mask matching: the query contains two images of the same location at different times, and the target is the corresponding change mask.

\paragraph{SYSU-CD.}
SYSU-CD~\cite{shi2022deeply} is another aerial-image change-detection dataset, introduced to study robust change representations under noise, scale variation, and pseudo-change interference.

\paragraph{VL-CMU-CD.}
VL-CMU-CD~\cite{alcantarilla2018streetviewa} extends change detection to street-view imagery captured over time, focusing on structural changes observed from vehicle-mounted camera sequences. GeoMEB uses VL-CMU-CD as a street-level change-detection task: the query contains a pair of street-view images from different times, and the candidate is a change mask that localizes visual changes.

\paragraph{OSF Recovery.}
OSF Recovery~\cite{huang2025built} studies post-disaster recovery by using street-level imagery to assess how built environments recover after extreme weather events. GeoMEB converts this task into retrieval over recovery-status labels, so the model must match multi-temporal visual evidence to the correct disaster-recovery category.

\paragraph{PlacePulse.}
PlacePulse~\cite{salesses2012place} captures collective human perception of urban street-view environments through crowd judgments of visual qualities such as safety and liveliness. GeoMEB uses six perception dimensions---safety, liveliness, beauty, wealth, depression, and boringness---and converts each continuous perception score into textual score labels at 0.5-point intervals for image-to-text ranking.

\paragraph{MP16Pro.}
MP16Pro~\cite{jia2024g3} is released with G3 for worldwide geolocalization by augmenting MP16 with textual geographical descriptions associated with image locations. We uses MP16Pro only as auxiliary training set. It provides street-view geolocation-style supervision, helping the model learn location-aware urban representations.

\section{Task formulation and examples}
\label{sec:task_examples}

\begin{figure*}[!t]
    \centering
    \includegraphics[width=0.85\textwidth]{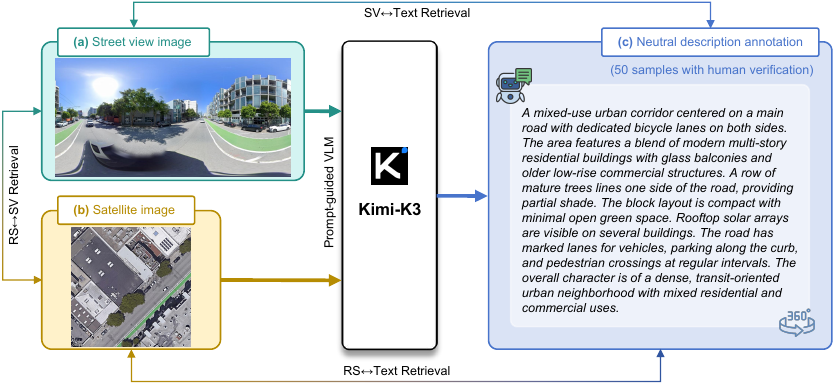}
    \caption{Annotated aligned VIGOR triplets used for instruction-shift analysis.}
    \label{fig:vigor_triplet_example}
\end{figure*}

GeoMEB uses an instruction-conditioned query-target framework for all task examples. Query-side instructions define the evaluated relation, while candidate-side instructions define the representation role of each target candidate. Appendix Table~\ref{tab:supp_task_examples_retrieval}$\sim$\ref{tab:supp_task_examples_classification} provide standardized formulations grouped by meta-task. Figure~\ref{fig:vigor_triplet_example} shows a VIGOR triplet used for instruction-induced embedding shift analysis.

\newcommand{\taskexamplecell}[1]{\textit{#1}}


\begin{table*}[!t]
\centering
\tiny
\setlength{\tabcolsep}{2.5pt}
\renewcommand{\arraystretch}{0.78}
\begin{tabular}{m{0.07\textwidth}m{0.08\textwidth}m{0.24\textwidth}m{0.15\textwidth}>{\columncolor{gray!12}}m{0.24\textwidth}>{\columncolor{gray!12}}m{0.15\textwidth}}
\toprule
\textbf{Dataset} & \textbf{Subtask} & \textbf{Query text} & \textbf{Query image} & \textbf{Target text} & \textbf{Target image} \\
\midrule
RSICD & i2t & \taskexamplecell{Represent this remote sensing image for retrieval.} & {\includegraphics[width=0.50\linewidth]{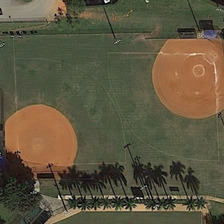}} & \taskexamplecell{Find a remote sensing image matching this description.}\par \textbf{There are two big baseball fields on this earth.} & - \\
\midrule
RSICD & t2i & \taskexamplecell{Find a remote sensing image matching this description.} \par \textbf{Several stones are on the beach.} & - & \taskexamplecell{Represent this remote sensing image.} & {\includegraphics[width=0.50\linewidth]{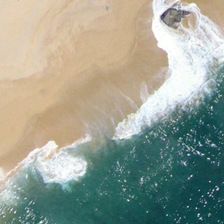}} \\
\midrule
UAV-GeoLoc & country & \taskexamplecell{Find the satellite image matching this UAV aerial view.} & {\includegraphics[width=0.50\linewidth]{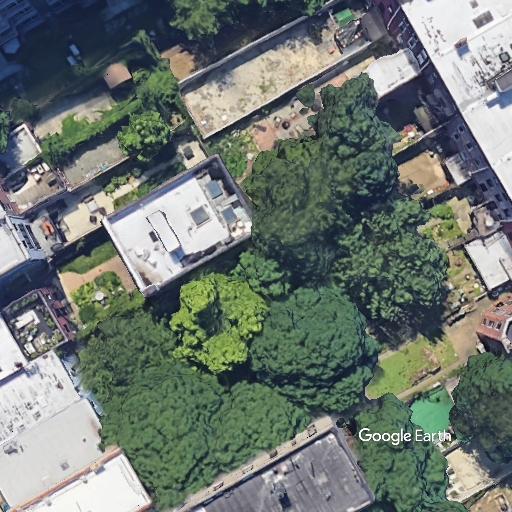}} & \taskexamplecell{Represent the given satellite image.} & {\includegraphics[width=0.50\linewidth]{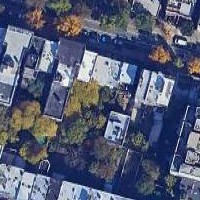}} \\
\midrule
UAV-GeoLoc & terrain & \taskexamplecell{Find the satellite image matching this UAV aerial view.} & {\includegraphics[width=0.50\linewidth]{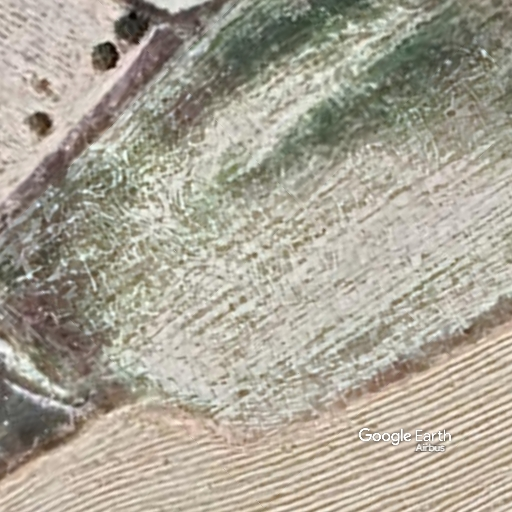}}  & \taskexamplecell{Represent the given satellite image.} & {\includegraphics[width=0.50\linewidth]{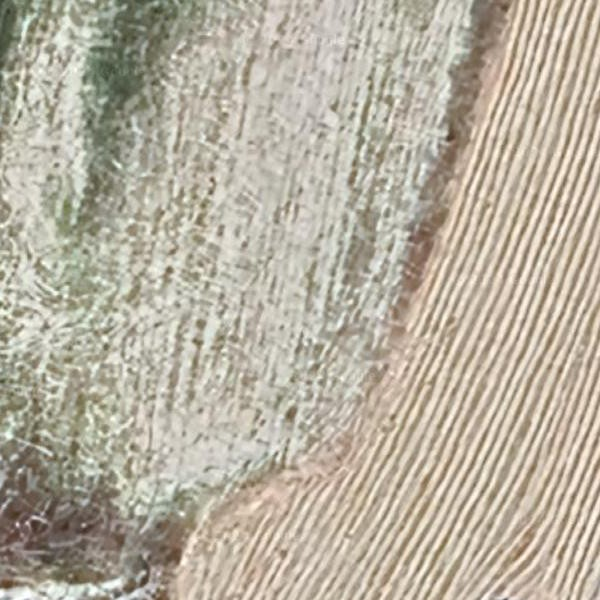}} \\
\midrule
VIGOR & cross-view retrieval & \taskexamplecell{Given the street view image and its description, identify the matching satellite image.} & {\includegraphics[width=0.95\linewidth]{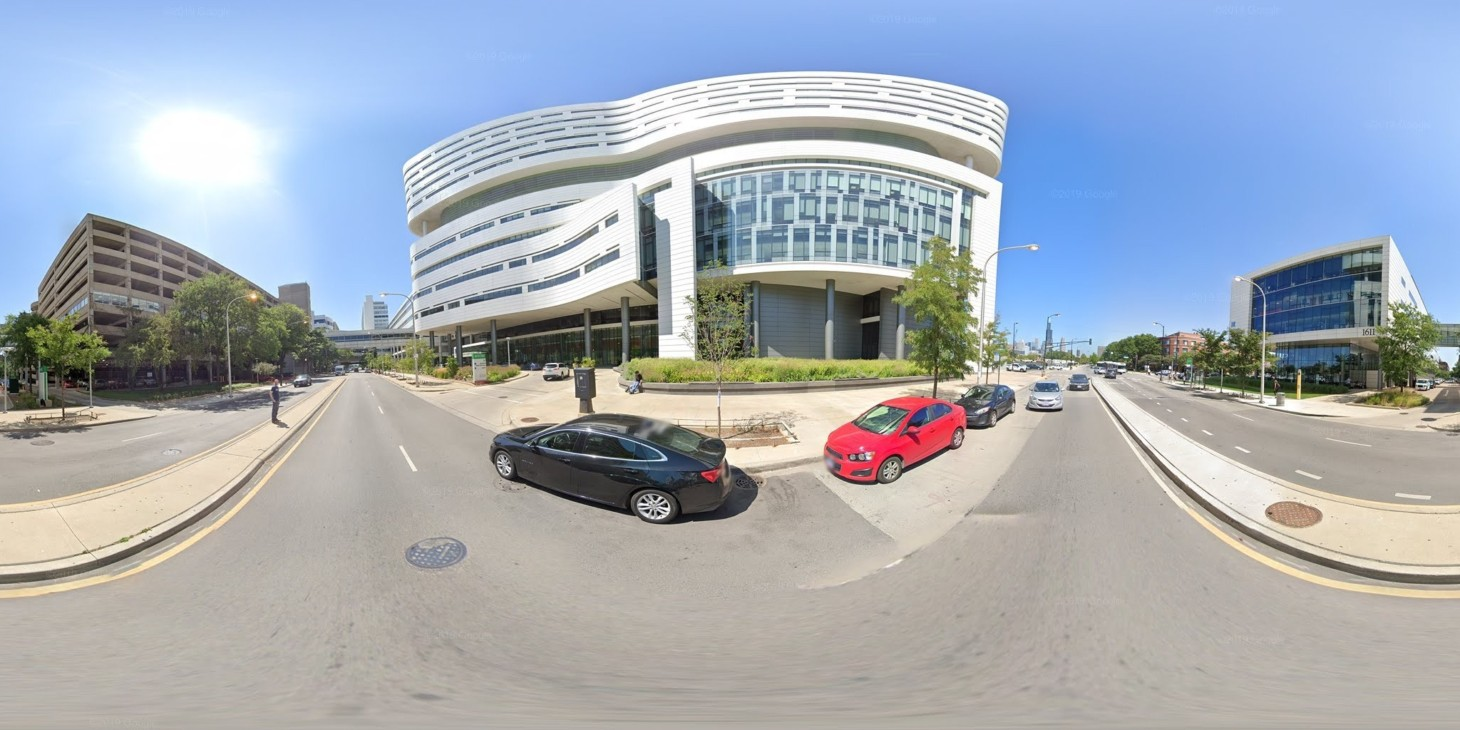}} & \taskexamplecell{Represent the given satellite image for cross-view retrieval.} & {\includegraphics[width=0.50\linewidth]{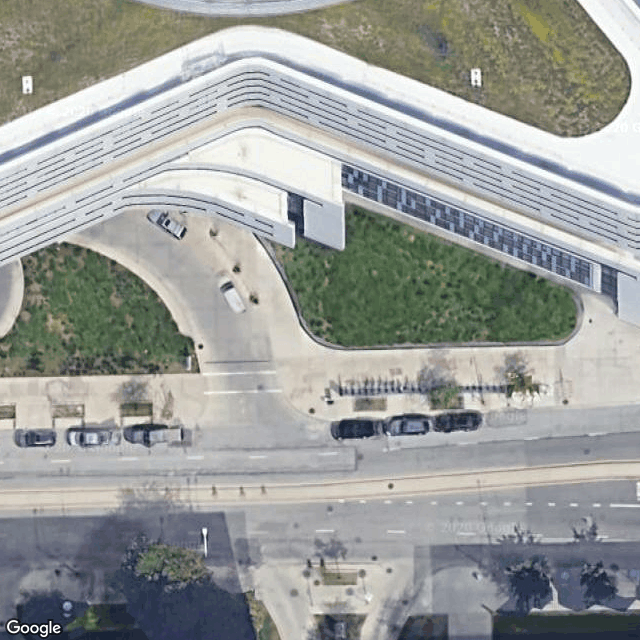}} \\
\midrule
SV Localization & i2t & \taskexamplecell{Find the description that matches this street-level image.} & (See full caption below due to length limit.) & \taskexamplecell{Represent the given street-level scene description.}\par \textbf{(See full caption below due to length limit.)} & - \\
\midrule
SV Localization & t2i & \taskexamplecell{Find the street-level image that matches this description.}\par \textbf{(See full caption below due to length limit.)} & - & \taskexamplecell{Represent the given street-level image.} & (See full caption below due to length limit.) \\
\midrule
Im2GPS3k & GPS i2t & \taskexamplecell{Find the geographic region that best matches the location shown in this street-level photo.} & {\includegraphics[width=0.50\linewidth]{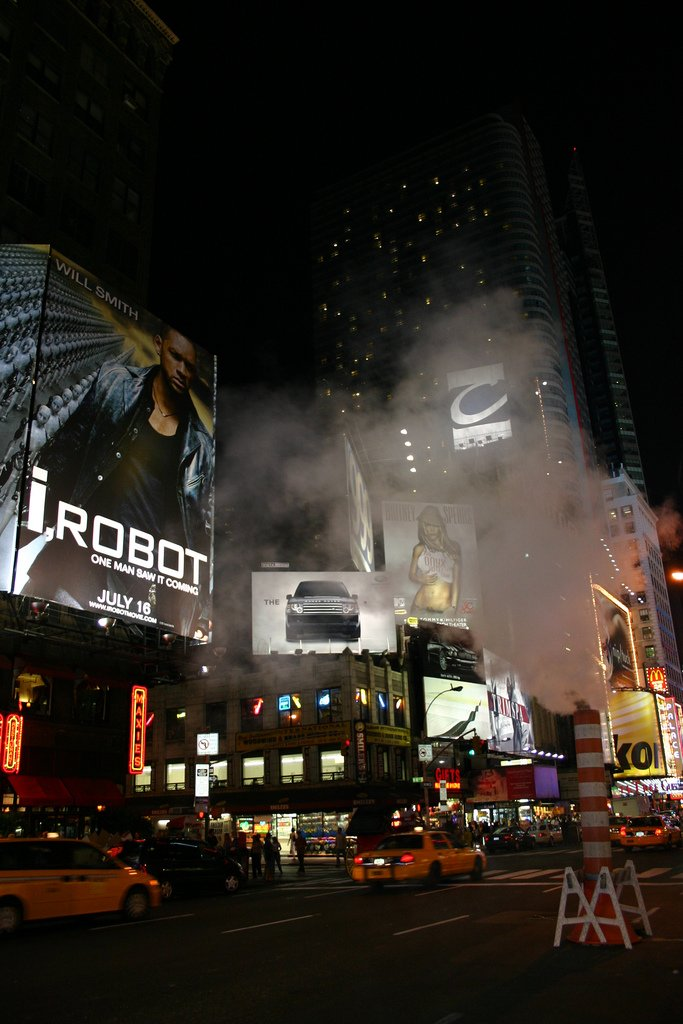}} & \taskexamplecell{Represent the given geographic region.}\par \textbf{A geographic region between latitude 40.7°N to 40.8°N and longitude 74.1°W to 74.0°W.} & - \\
\midrule
Im2GPS3k & GPS t2i & \taskexamplecell{Find the street-level images taken within this geographic region.}\par \textbf{A geographic region between latitude 19.6°N to 19.7°N and longitude 156.0°W to 155.9°W.} & - & \taskexamplecell{Represent the given street-level image.} & {\includegraphics[width=0.65\linewidth]{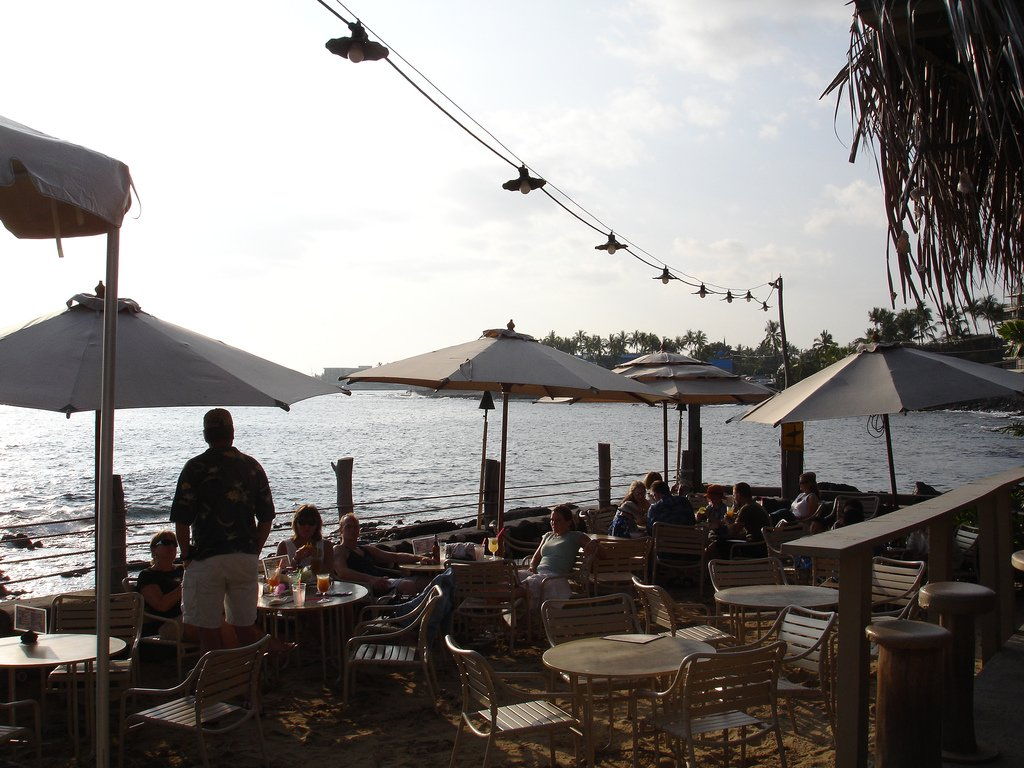}} \\
\midrule
Im2GPS3k & Caption i2t & \taskexamplecell{Find the description that matches this street-level image.} & (See full caption below due to length limit.) & \taskexamplecell{Represent the given street-level scene description.}\par \textbf{(See full caption below due to length limit.)} & - \\
\midrule
Im2GPS3k & Caption t2i & \taskexamplecell{Find the street-level image that matches this description.}\par \textbf{(See full caption below due to length limit.)} & - & \taskexamplecell{Represent the given street-level image.} & (See full caption below due to length limit.) \\
\midrule
VRSBench & caption i2t & \taskexamplecell{Find a caption matching this remote sensing image.} & {\includegraphics[width=0.50\linewidth]{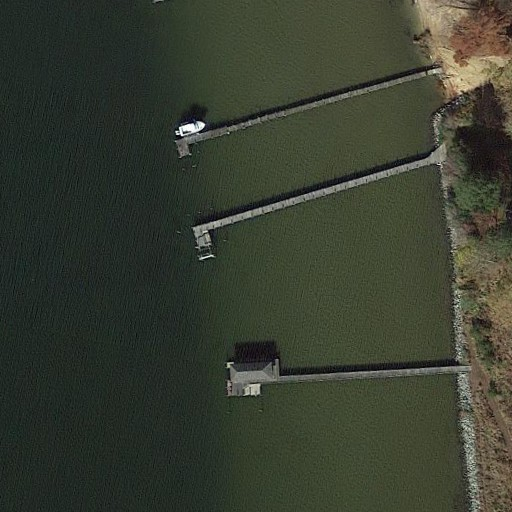}} & \taskexamplecell{Represent this remote sensing image caption.}\par \textbf{The high-resolution image displays a body of water with jetties extending into it. There is one small ship located in the top-middle area of the image, aligned with one of the jetties.} & - \\
\midrule
VRSBench & caption t2i & \taskexamplecell{Find a remote sensing image matching this caption.}\par \textbf{The high-resolution image displays a pair of tennis courts surrounded by trees. The courts are adjacent to each other with their longer sides in parallel. The surrounding area includes a well-maintained grassy field to the south of the courts, and to the east, there is a small building.} & - & \taskexamplecell{Represent this remote sensing image.} & {\includegraphics[width=0.50\linewidth]{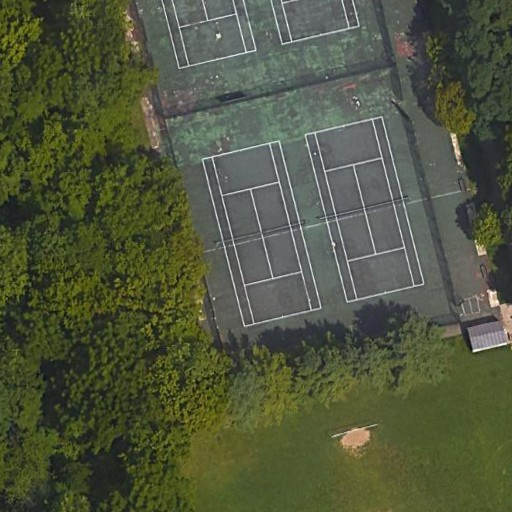}} \\
\bottomrule
\end{tabular}
\caption{Retrieval subtask examples. Each row shows one standardized retrieval formulation, with target-side fields shaded. In Query-text and Target-text columns, \textit{italic text} denotes the encoding instruction and \textbf{bold text} denotes the query or target content; the same convention applies to the following tables.}
\label{tab:supp_task_examples_retrieval}
\end{table*}


\begin{table*}[!t]
\centering
\tiny
\setlength{\tabcolsep}{2.5pt}
\renewcommand{\arraystretch}{0.78}
\begin{tabular}{m{0.07\textwidth}m{0.08\textwidth}m{0.3\textwidth}m{0.28\textwidth}>{\columncolor{gray!12}}m{0.10\textwidth}>{\columncolor{gray!12}}m{0.04\textwidth}}
\toprule
\textbf{Dataset} & \textbf{Subtask} & \textbf{Query text} & \textbf{Query image} & \textbf{Target text} & \textbf{Target image} \\
\midrule
VRSBench & vqa & \taskexamplecell{Represent the given remote sensing image with the following question.}\par \textbf{What color are the planes predominantly?} & {\includegraphics[width=0.25\linewidth]{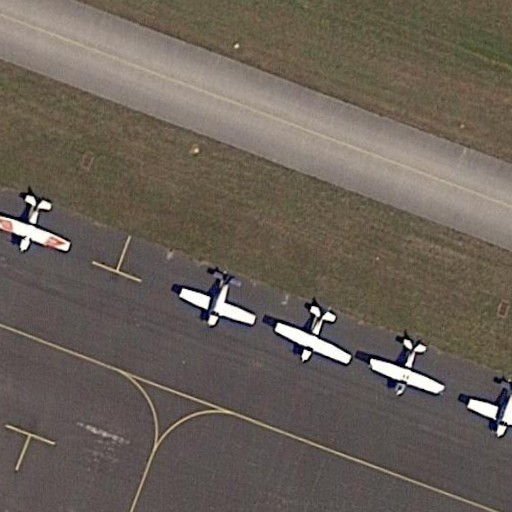}} & \taskexamplecell{Represent this answer.} \par \textbf{White.} & - \\
\midrule
MME-RealWorld & Autonomous Driving & \taskexamplecell{Represent the given image with the following question.}\par \textbf{This image shows the front view of the ego car. What is the future state of the brown suv in the middle?} & {\includegraphics[width=0.45\linewidth]{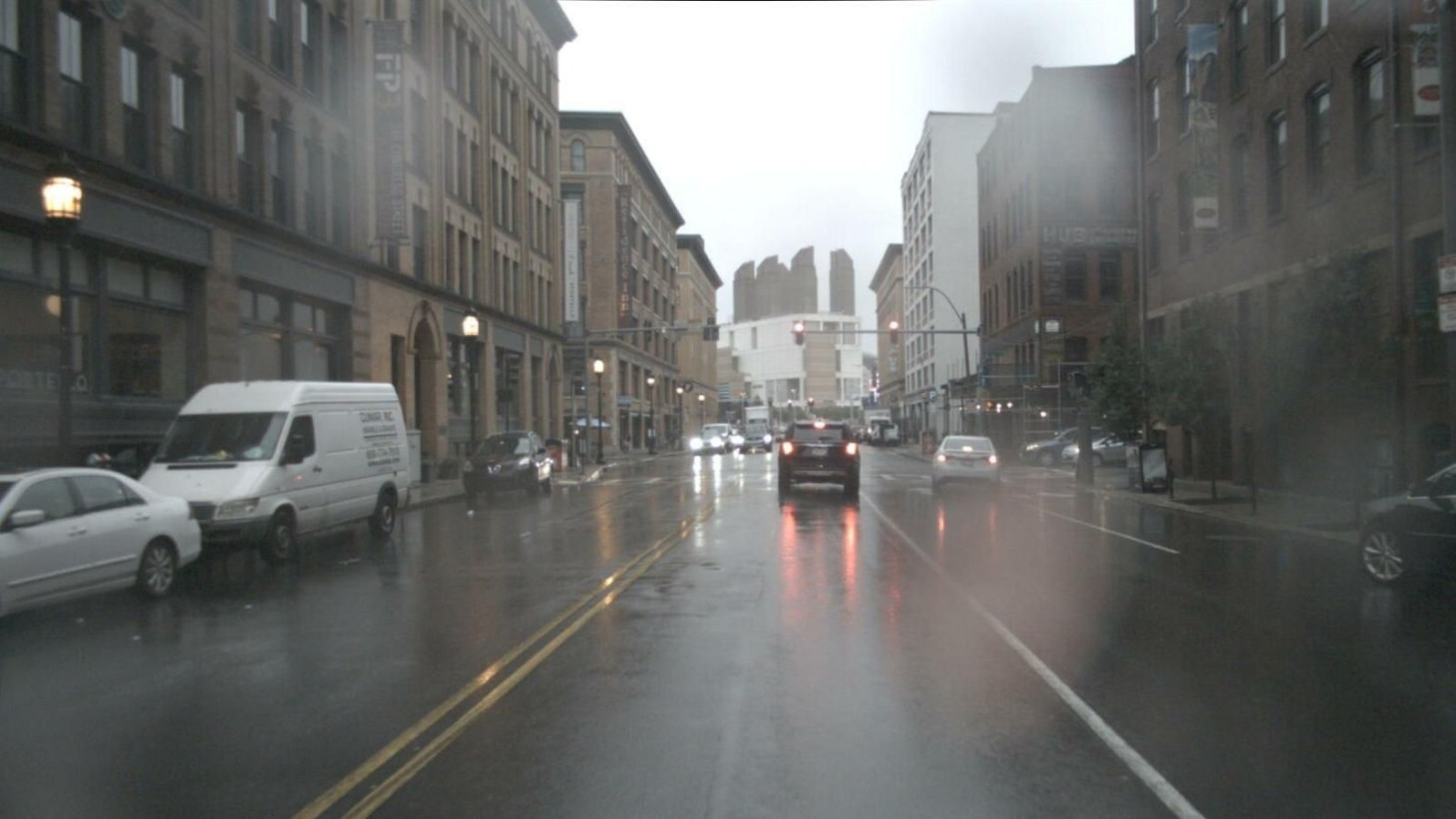}} & \taskexamplecell{Represent this answer choice.}\par \textbf{Brake gently to a stop.} & - \\
\midrule
MME-RealWorld & MME-HD-CN & \taskexamplecell{Represent the given image with the following question.}\par \textbf{What is the larger font content on the external wall above the entrance of the building in the picture? (in Chinese)} & {\includegraphics[width=0.35\linewidth]{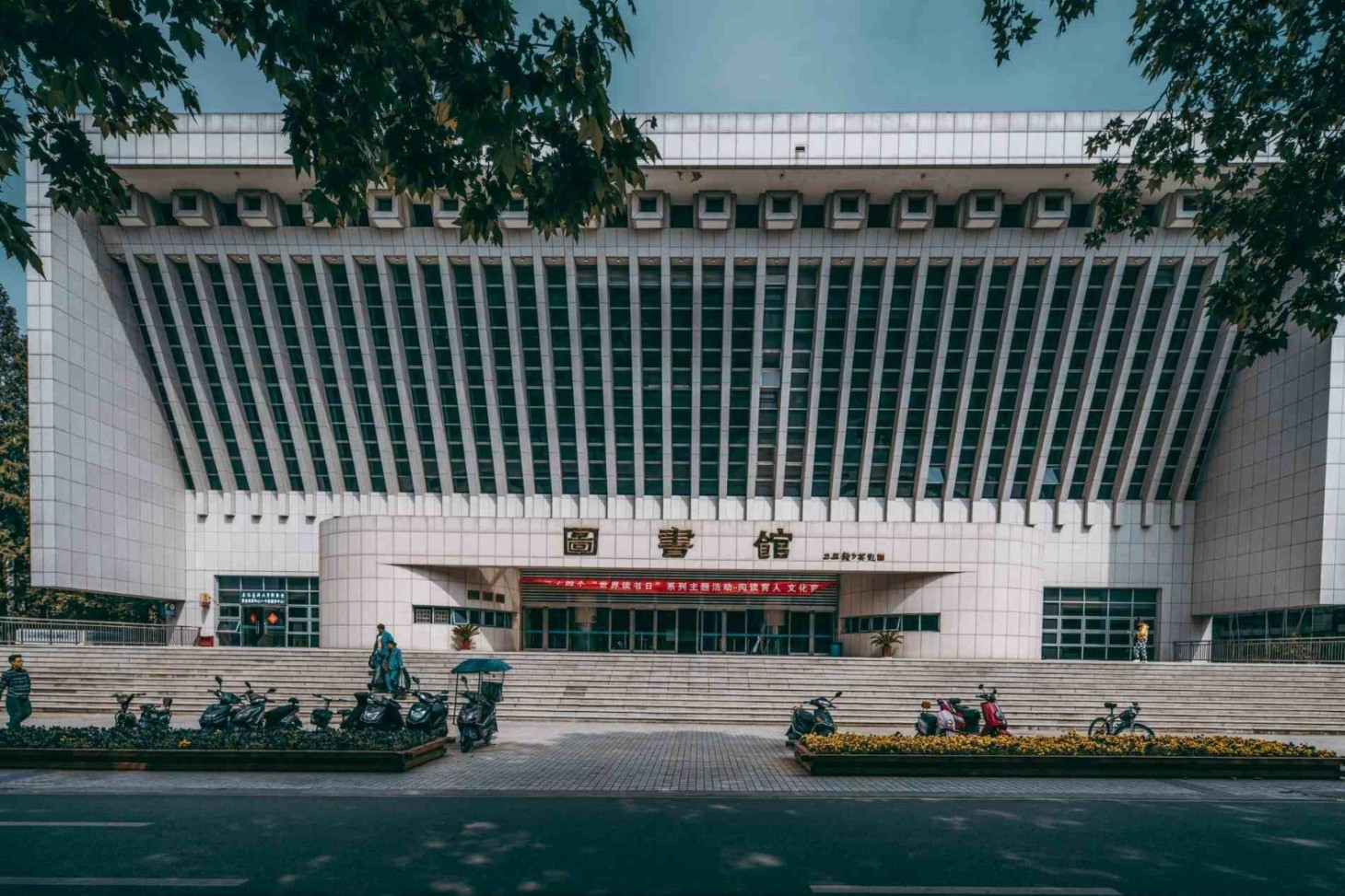}} & \taskexamplecell{Represent this answer choice.}\par \textbf{Library (in Chinese).} & - \\
\midrule
MME-RealWorld & diagram-and-table & \taskexamplecell{Represent the given image with the following question.}\par \textbf{What is the revenue (USD) of Multi-Passenger Taxis in 2033 in the Revenue Breakdown table?} & {\includegraphics[width=0.55\linewidth]{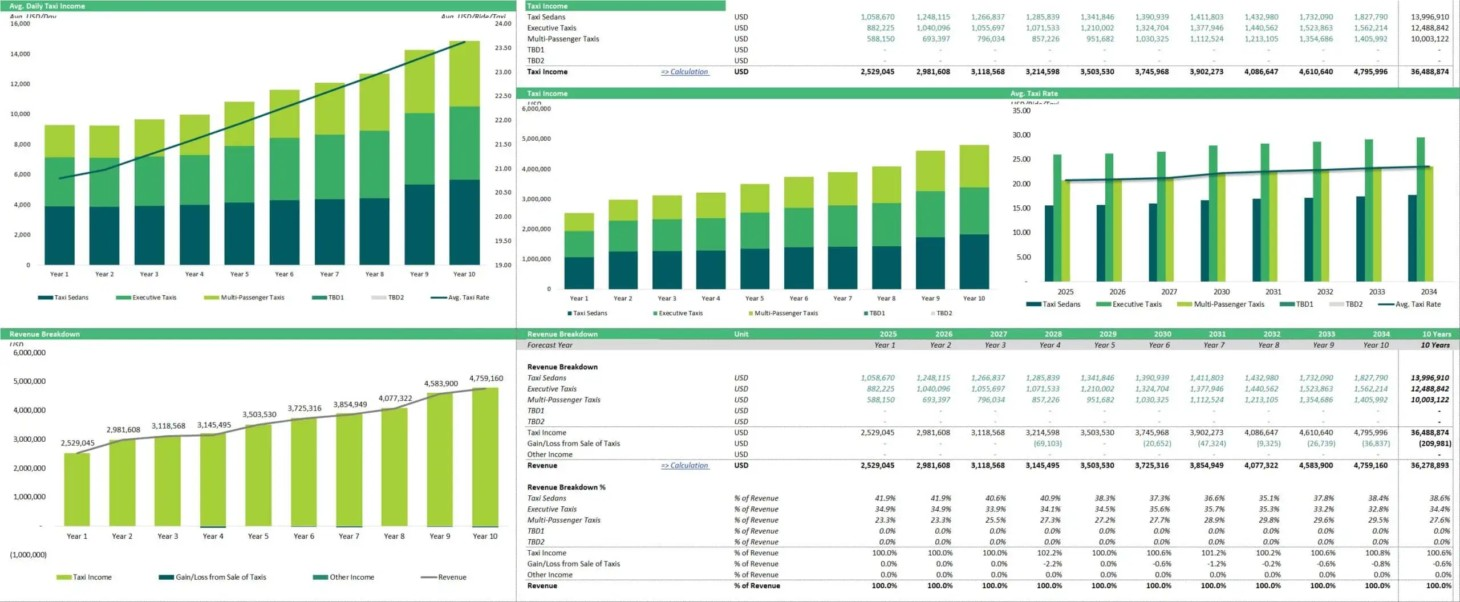}} & \taskexamplecell{Represent this answer choice.}\par \textbf{1,354,686} & - \\
\midrule
MME-RealWorld & monitoring-images & \taskexamplecell{Represent the given image with the following question.}\par \textbf{What material is the roof of the small white house on the roof of the building in the middle of the picture?} & {\includegraphics[width=0.28\linewidth]{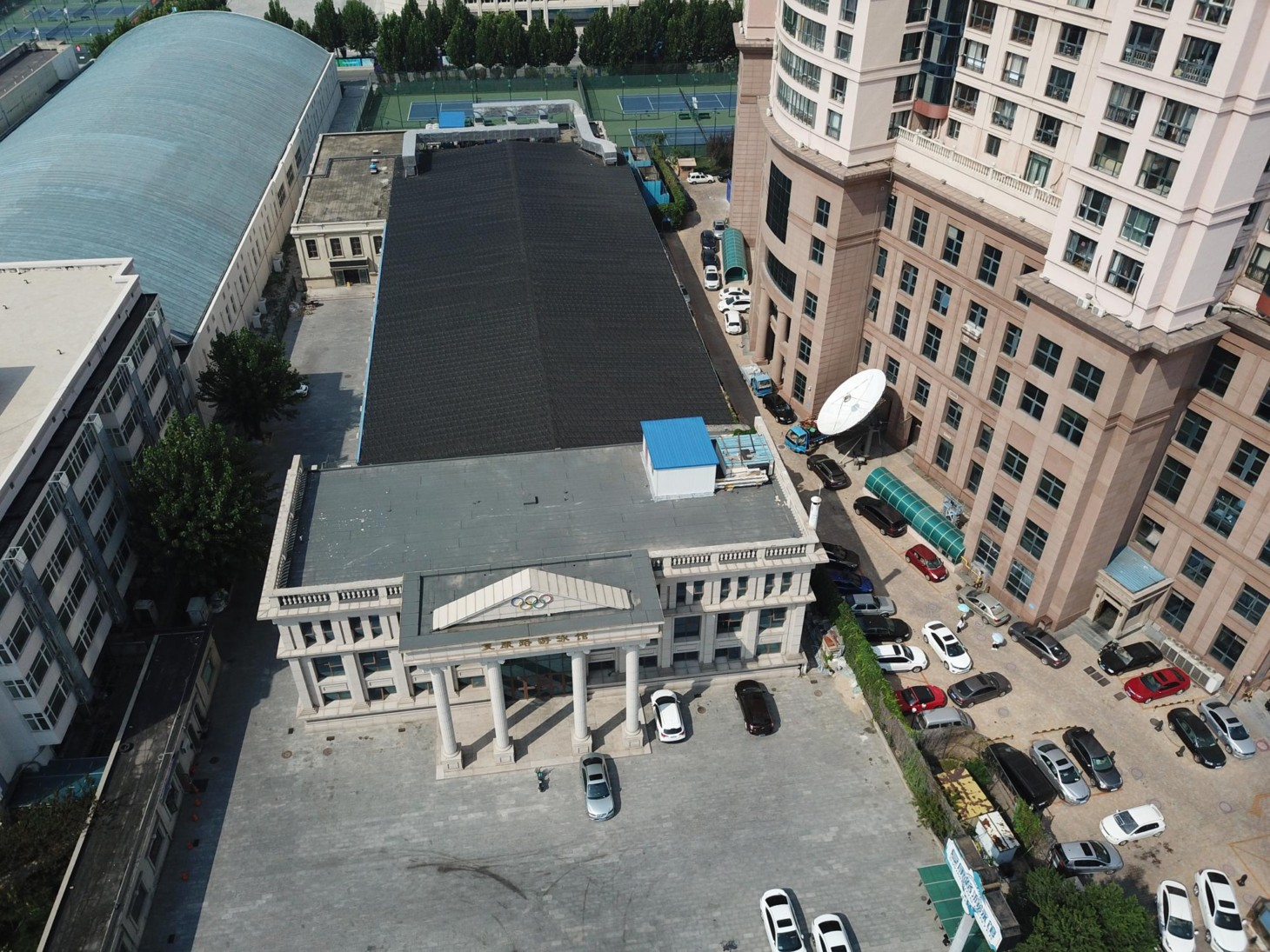}} & \taskexamplecell{Represent this answer choice.}\par \textbf{iron} & - \\
\midrule
MME-RealWorld & ocr-cc & \taskexamplecell{Represent the given image with the following question.}\par \textbf{What is the amusement project above the TarzaN SwiNGS on the right side of the cable car in the picture?} & {\includegraphics[width=0.25\linewidth]{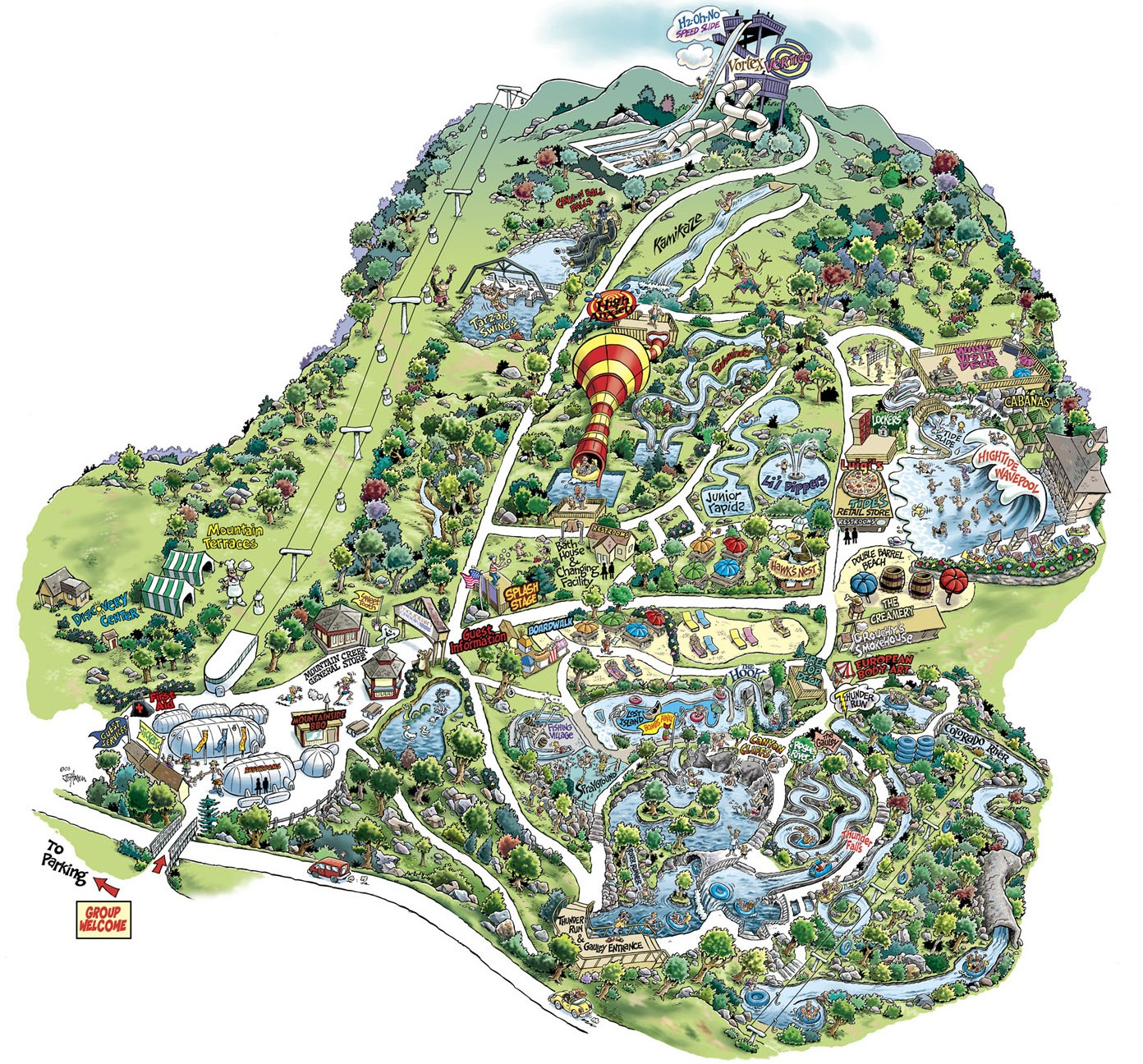}} & \taskexamplecell{Represent this answer choice.}\par \textbf{CANNON BALL FALLS} & - \\
\midrule
MME-RealWorld & remote-sensing & \taskexamplecell{Represent the given image with the following question.}\par \textbf{Where are the football fields located in this picture?} & {\includegraphics[width=0.25\linewidth]{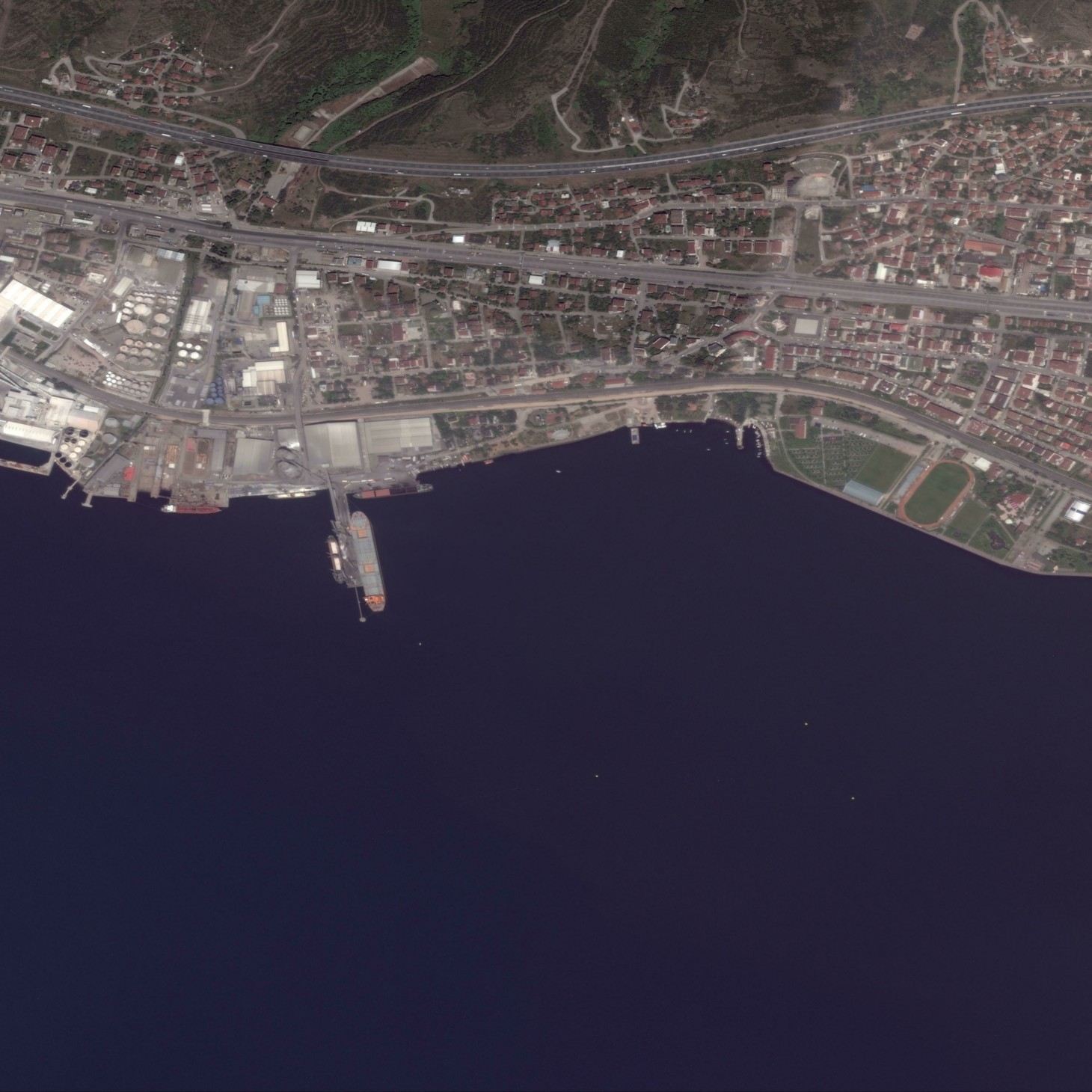}} & \taskexamplecell{Represent this answer choice.}\par \textbf{Beside the water in the upper right area of this picture} & - \\
\midrule
UrBench & counting & \taskexamplecell{Represent the given image with the following question.}\par \textbf{How many printed white or yellow crosswalks are shown in this image?} & {\includegraphics[width=0.25\linewidth]{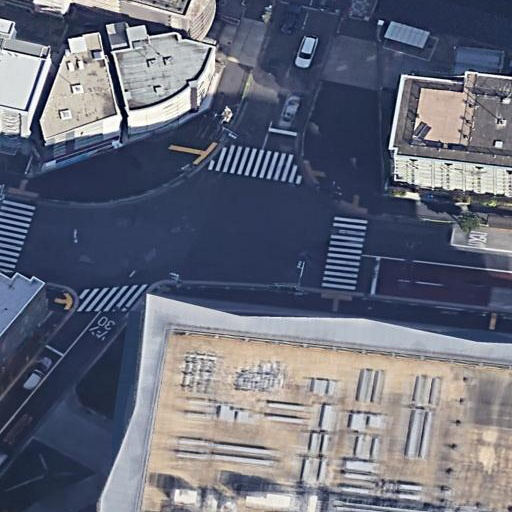}} & \taskexamplecell{Represent the user's input.}\par \textbf{4} & - \\
\midrule
UrBench & object-attribute-recognition oe & \taskexamplecell{Represent the given image with the following question.}\par \textbf{The satellite image depicting a specific area of a city. In this image, I have highlighted the rooftop of a building with a red box. The panoramic street view image is taken from the center of the area shown in the satellite image. This panoramic image covers a full 360-degree view. Your task is to identify the highlighted building in the panoramic street view image and determine the number of floors of this building based on its front or side view. To do this, follow these steps: Locate the building highlighted with the red box in the satellite image within the panoramic street view image. Analyze the building's front or side view in the street view image to determine its number of floors. Provide the number of floors of the building.} & {\includegraphics[width=0.25\linewidth]{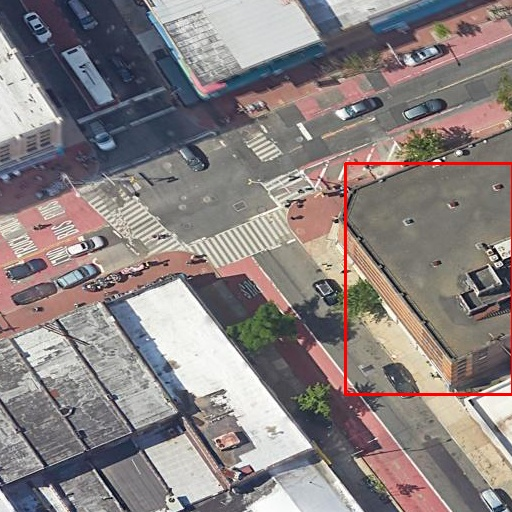}}  {\includegraphics[width=0.50\linewidth]{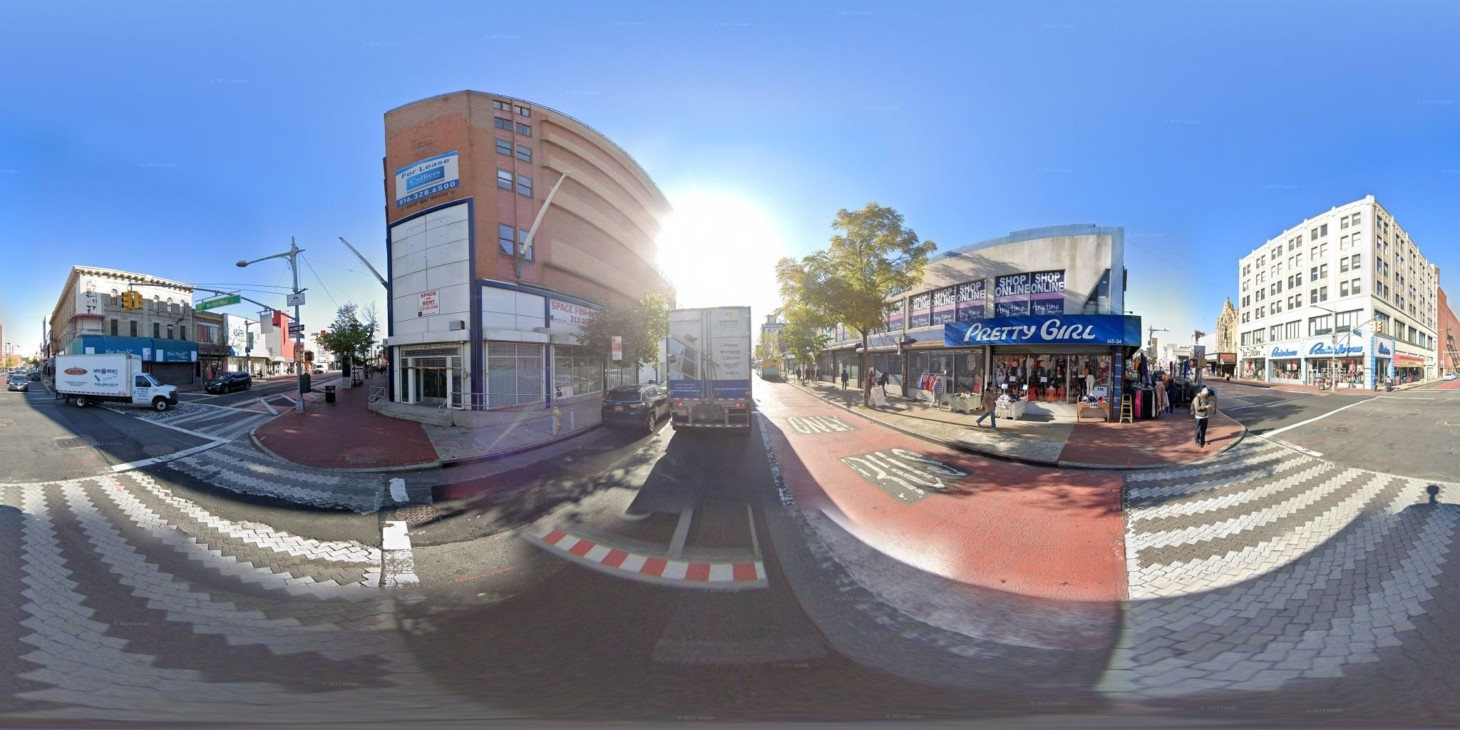}} & \taskexamplecell{Represent the user's input.} \par \textbf{12} & - \\
\midrule
UrBench & road-understanding oe & \taskexamplecell{Represent the given image with the following question.}\par \textbf{Are the following four images captured successively on the same road? If they are captured successively on the same road, then all four images will contain many of the same objects, such as buildings, but from slightly different viewpoints. Otherwise, no identical buildings would appear in any of the four images. Answer yes or no based on the hint above.} & {\includegraphics[width=0.45\linewidth]{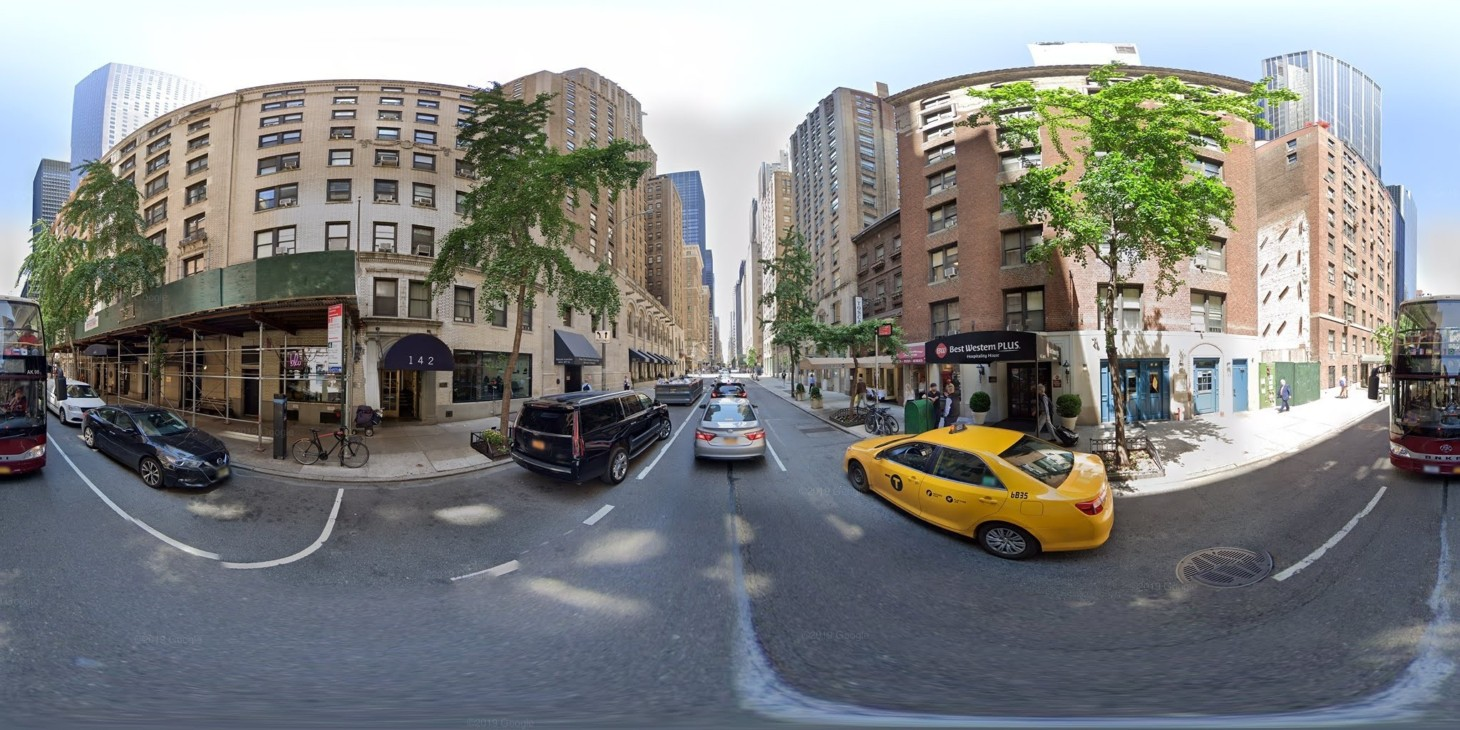}}  {\includegraphics[width=0.45\linewidth]{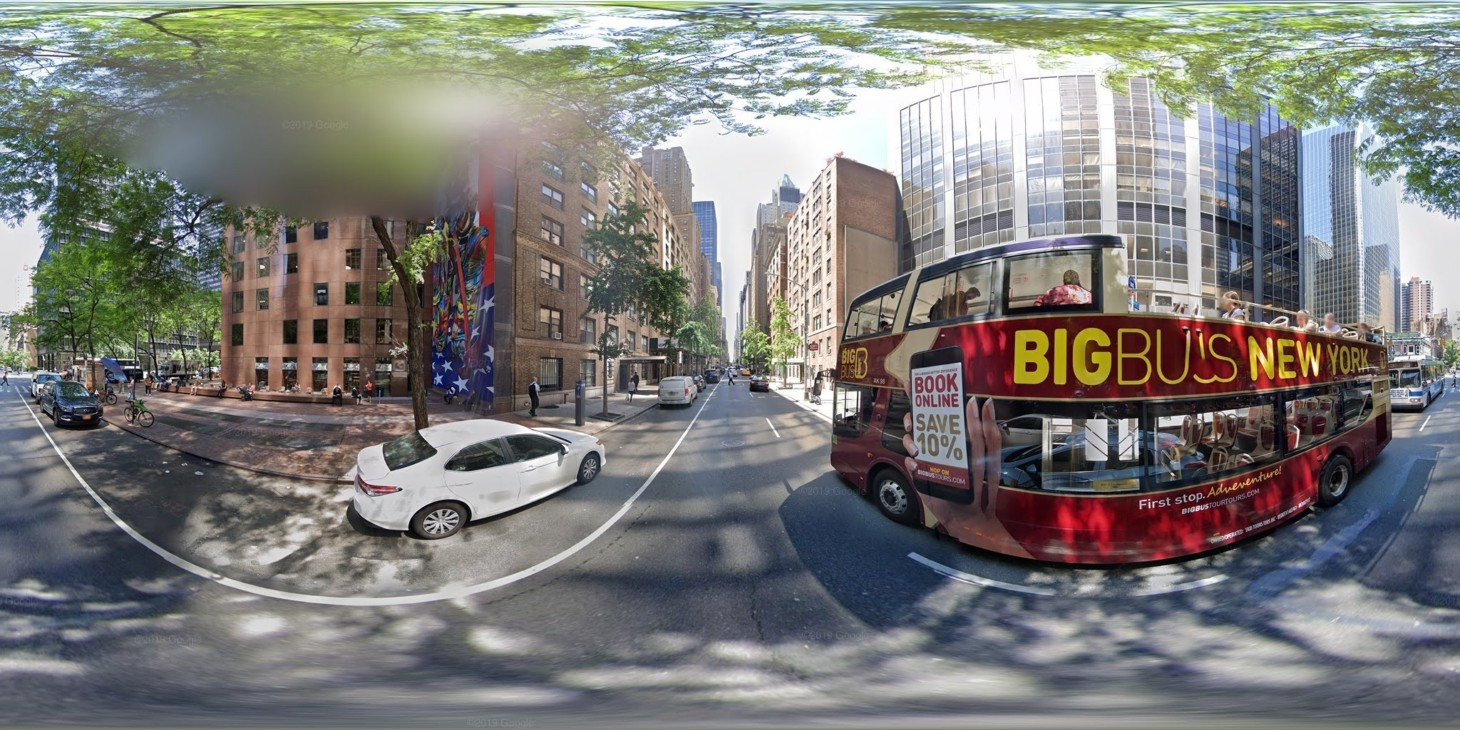}} \par {\includegraphics[width=0.45\linewidth]{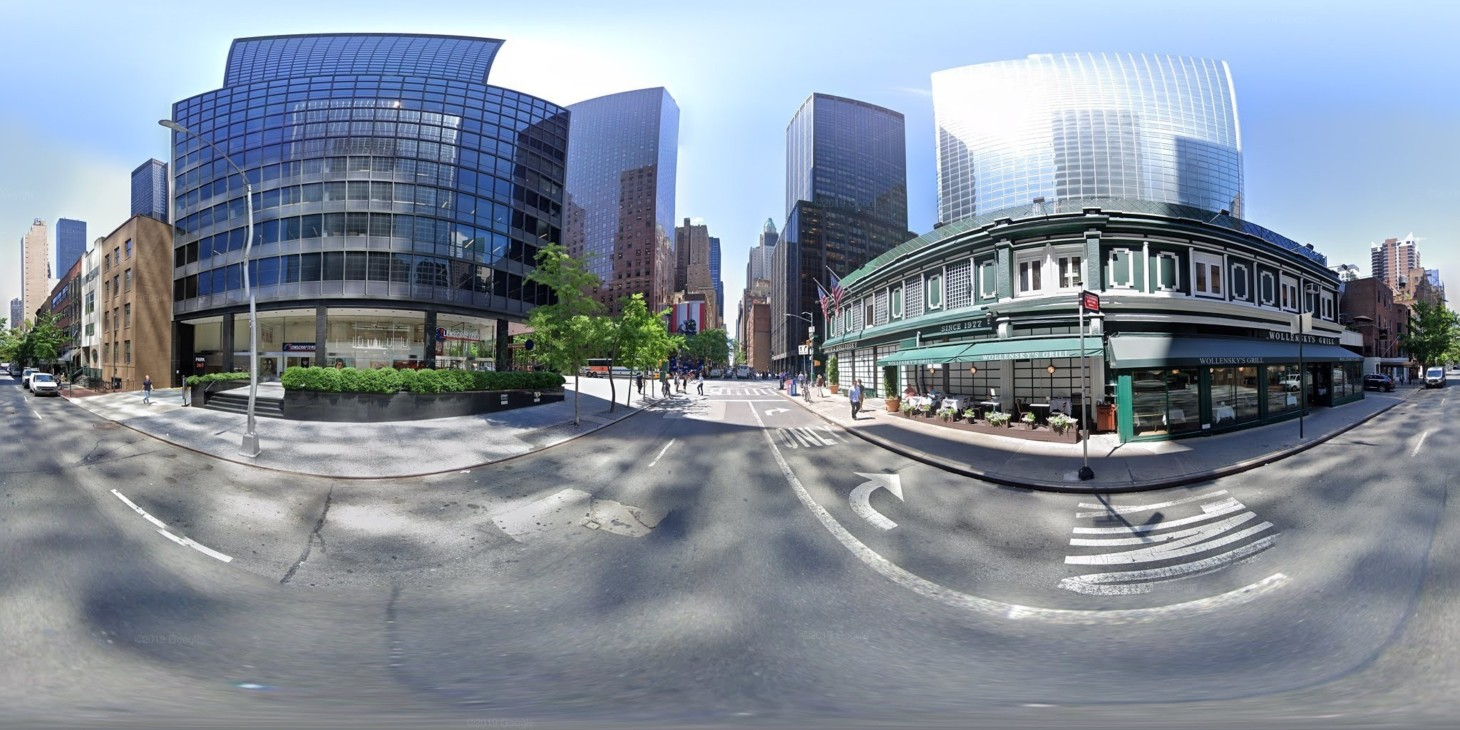}}  {\includegraphics[width=0.45\linewidth]{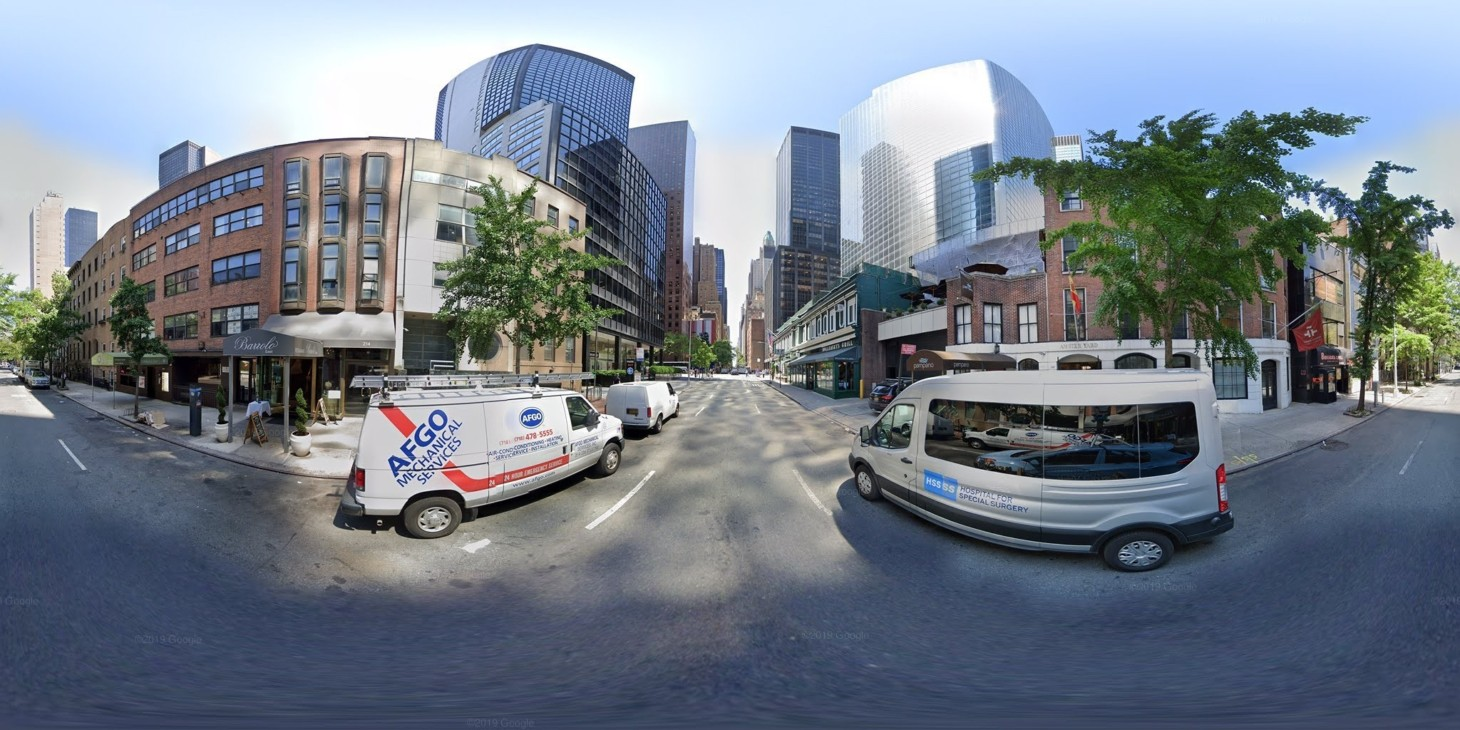}} & \taskexamplecell{Represent the user's input.} \par \textbf{Yes} & - \\
\midrule
UrBench & role-based-reasoning & \taskexamplecell{Represent the given image with the following question.}\par \textbf{If I am a pedestrian walking on this street, are there any stores that I can stop in and grab a quick bite to eat?} & {\includegraphics[width=0.45\linewidth]{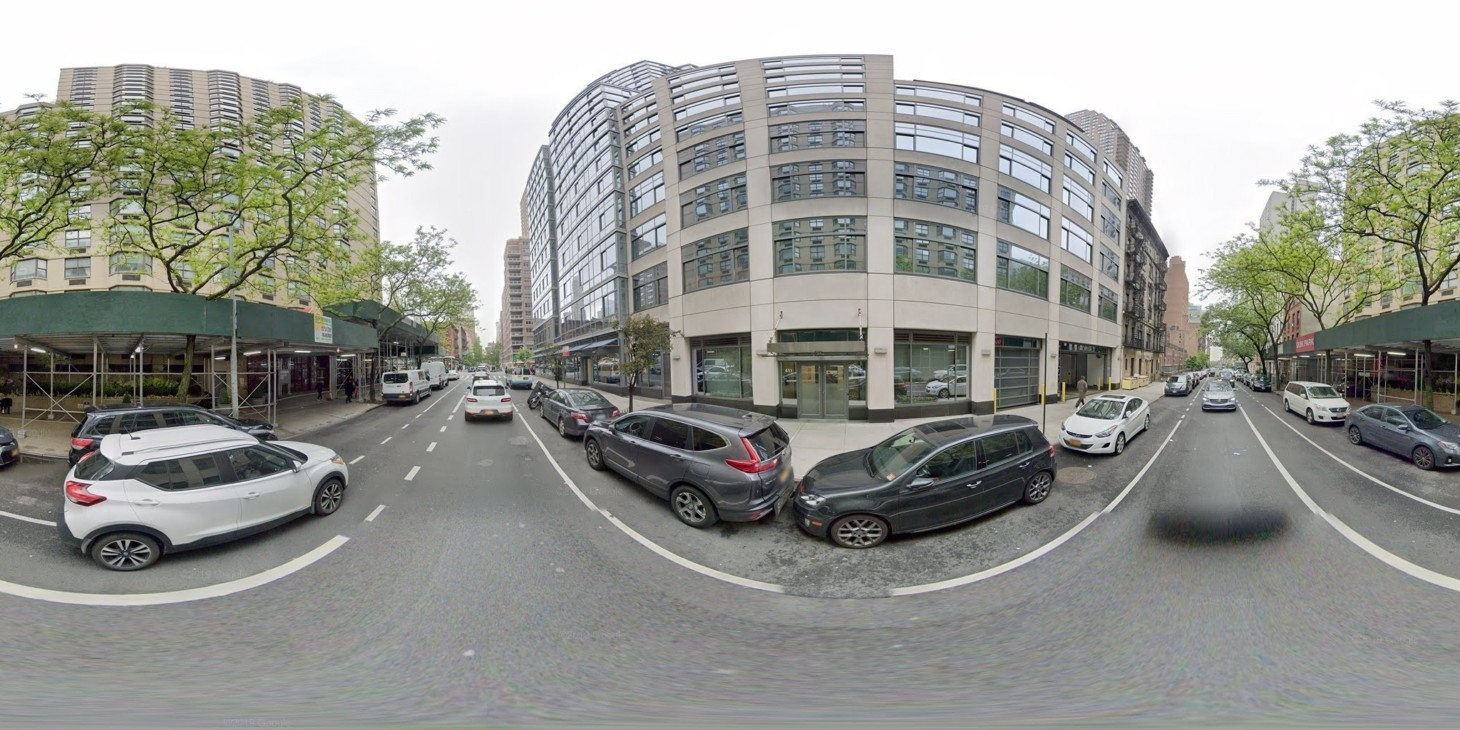}} & \taskexamplecell{Represent the user's input.}\par \textbf{No, there are no restaurants or cafes on this street.} & - \\
\midrule
UrBench & traffic-sign-reasoning & \taskexamplecell{Represent the given image with the following question.}\par \textbf{According to the traffic sign in box 0, what is the maximum speed limit on this road?} & {\includegraphics[width=0.28\linewidth]{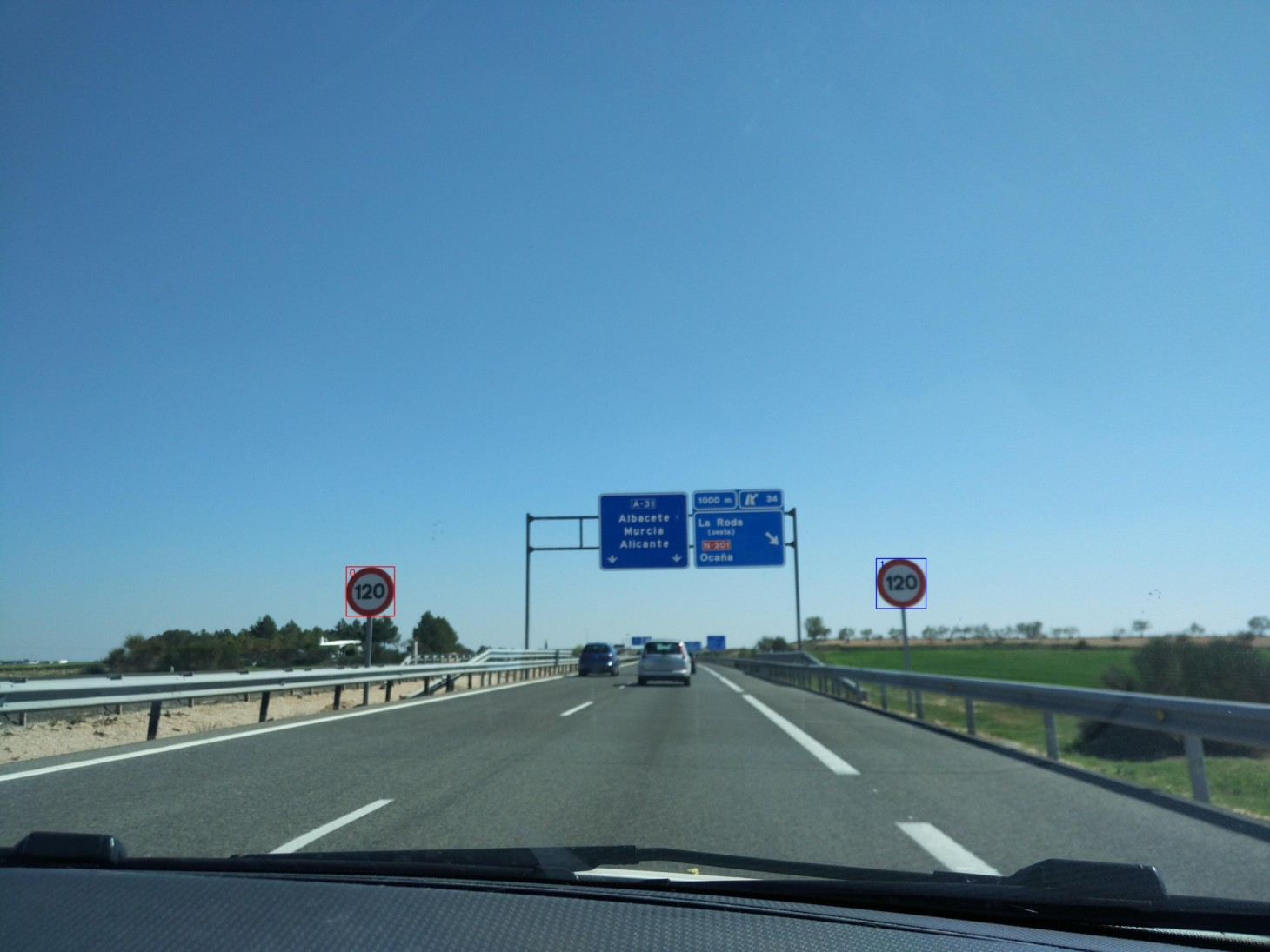}} & \taskexamplecell{Represent the user's input.}\par \textbf{120 km/h} & - \\
\midrule
UrBench & visual-prompt-reasoning & \taskexamplecell{Represent the given image with the following question.}\par \textbf{What is the person in the bounding box likely doing?} & {\includegraphics[width=0.45\linewidth]{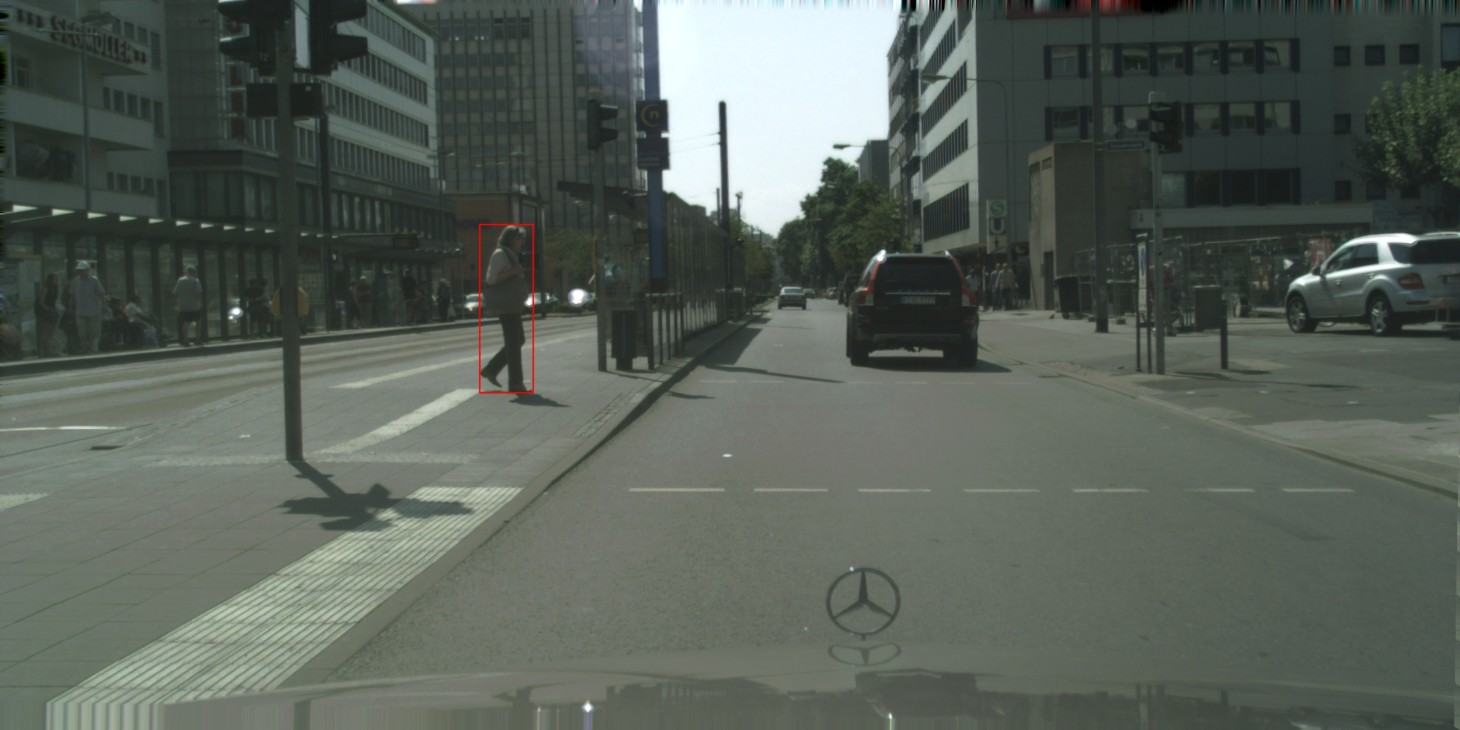}} & \taskexamplecell{Represent the user's input.}\par \textbf{Crossing the street} & - \\
\bottomrule
\end{tabular}
\caption{Visual question answering subtask examples.}
\label{tab:supp_task_examples_vqa}
\end{table*}


\begin{table*}[!t]
\centering
\tiny
\setlength{\tabcolsep}{2.5pt}
\renewcommand{\arraystretch}{0.78}
\begin{tabular}{m{0.07\textwidth}m{0.08\textwidth}m{0.15\textwidth}m{0.24\textwidth}>{\columncolor{gray!12}}m{0.17\textwidth}>{\columncolor{gray!12}}m{0.15\textwidth}}
\toprule
\textbf{Dataset} & \textbf{Subtask} & \textbf{Query text} & \textbf{Query image} & \textbf{Target text} & \textbf{Target image} \\
\midrule
Cityscapes & street-level object grounding & \taskexamplecell{Represent the given street scene image with the described object.}\par \textbf{Where exactly is a bicycle located? Return the bounding box.} & {\includegraphics[width=0.6\linewidth]{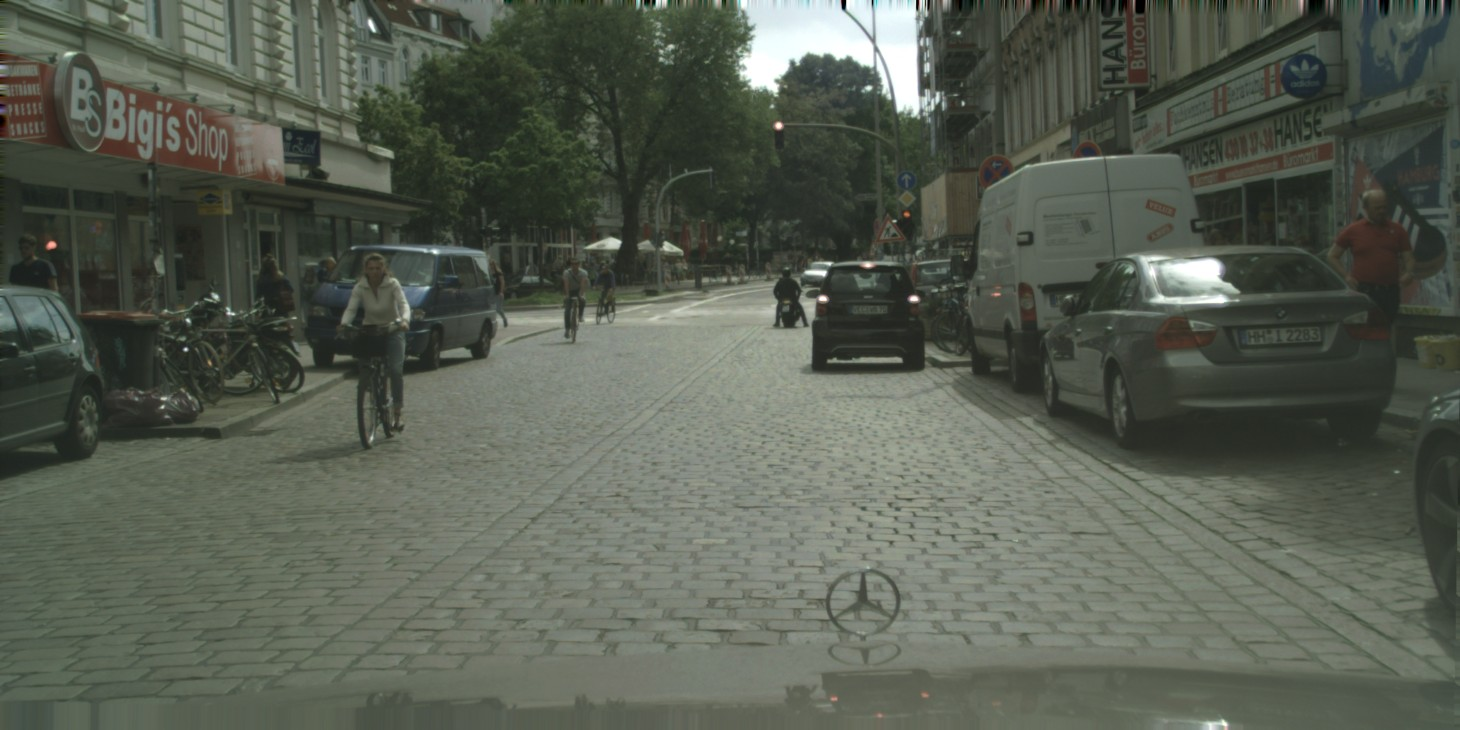}} & \taskexamplecell{Represent the given cropped region of the street scene.} & {\includegraphics[width=0.6\linewidth]{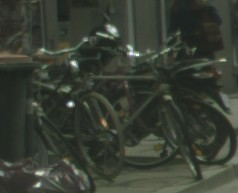}} \\
\midrule
Mapillary & street-level object grounding & \taskexamplecell{Represent the given street scene image with the described object.}\par \textbf{Identify the position of the warning winding road first right g1 sign in this image.} & {\includegraphics[width=0.4\linewidth]{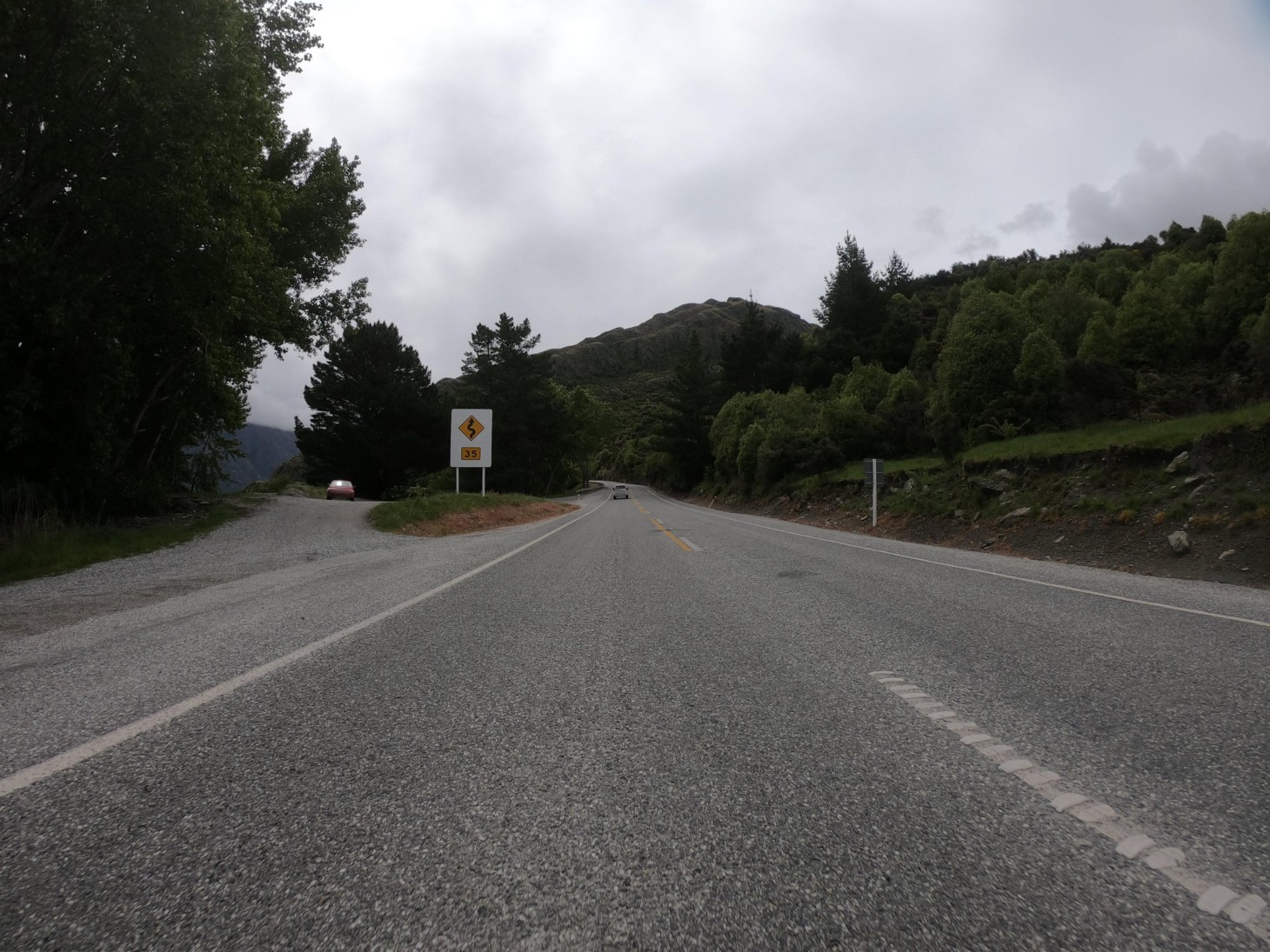}} & \taskexamplecell{Represent the given cropped region of the street scene.} & {\includegraphics[width=0.55\linewidth]{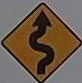}} \\
\midrule
VRSBench & referring & \taskexamplecell{Represent the given remote sensing image with the described object.}\par \textbf{The small white vehicle is parked on the right edge of the paved area.} & {\includegraphics[width=0.38\linewidth]{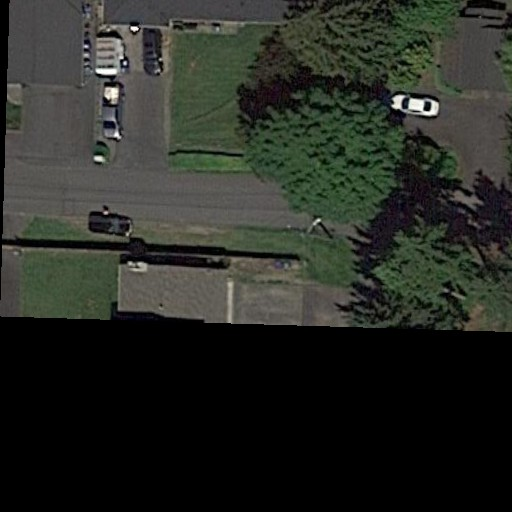}} & \taskexamplecell{Represent the given cropped region of the remote sensing image.} & {\includegraphics[width=0.55\linewidth]{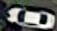}} \\
\bottomrule
\end{tabular}
\caption{Grounding subtask examples.}
\label{tab:supp_task_examples_grounding}
\end{table*}


\begin{table*}[!t]
\centering
\tiny
\setlength{\tabcolsep}{2.5pt}
\renewcommand{\arraystretch}{0.78}
\begin{tabular}{m{0.07\textwidth}m{0.06\textwidth}m{0.23\textwidth}m{0.21\textwidth}>{\columncolor{gray!12}}m{0.14\textwidth}>{\columncolor{gray!12}}m{0.15\textwidth}}
\toprule
\textbf{Dataset} & \textbf{Subtask} & \textbf{Query text} & \textbf{Query image} & \textbf{Target text} & \textbf{Target image} \\
\midrule
LEVIR-CD & change mask matching & \taskexamplecell{Represent the given remote sensing images with the following change detection task.}\par \textbf{Given two remote sensing images of the same location at different times, retrieve the change mask that highlights the differences between them.} & {\includegraphics[width=0.45\linewidth]{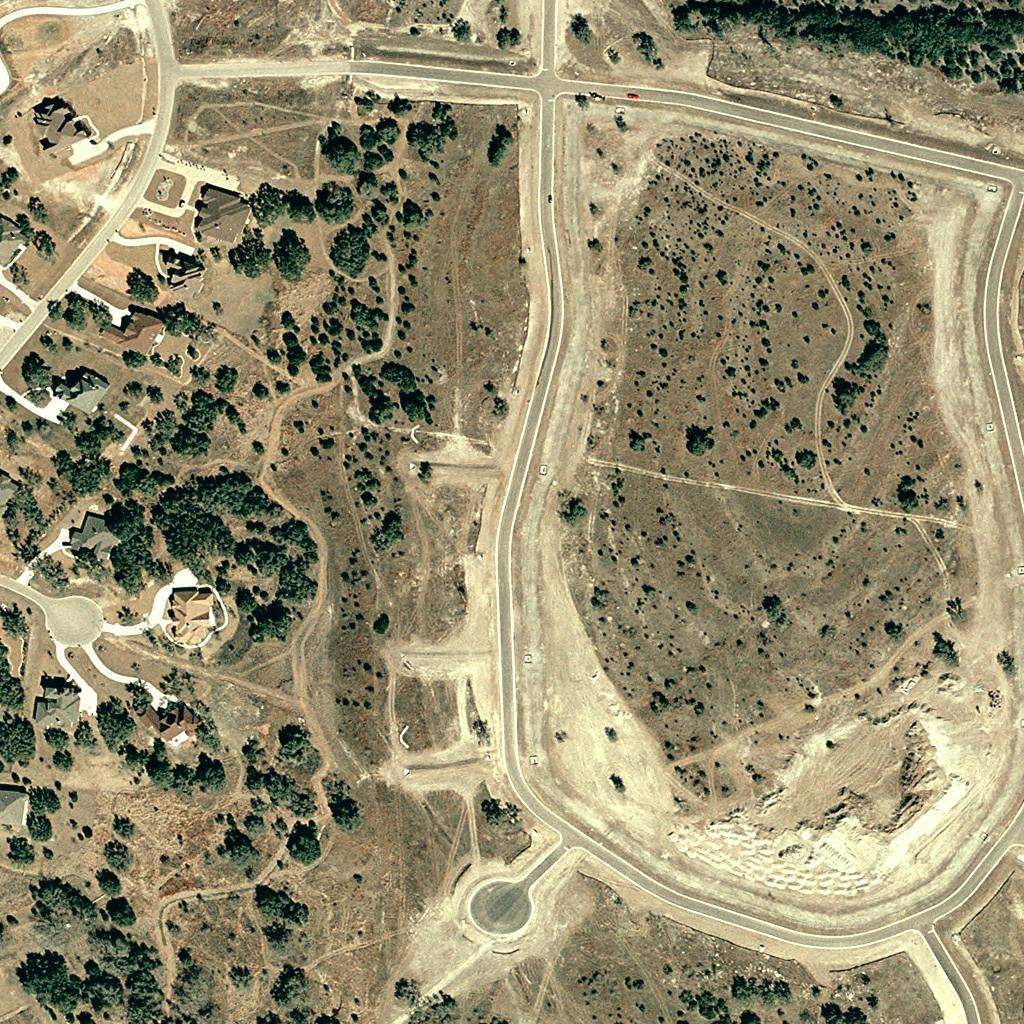}} {\includegraphics[width=0.45\linewidth]{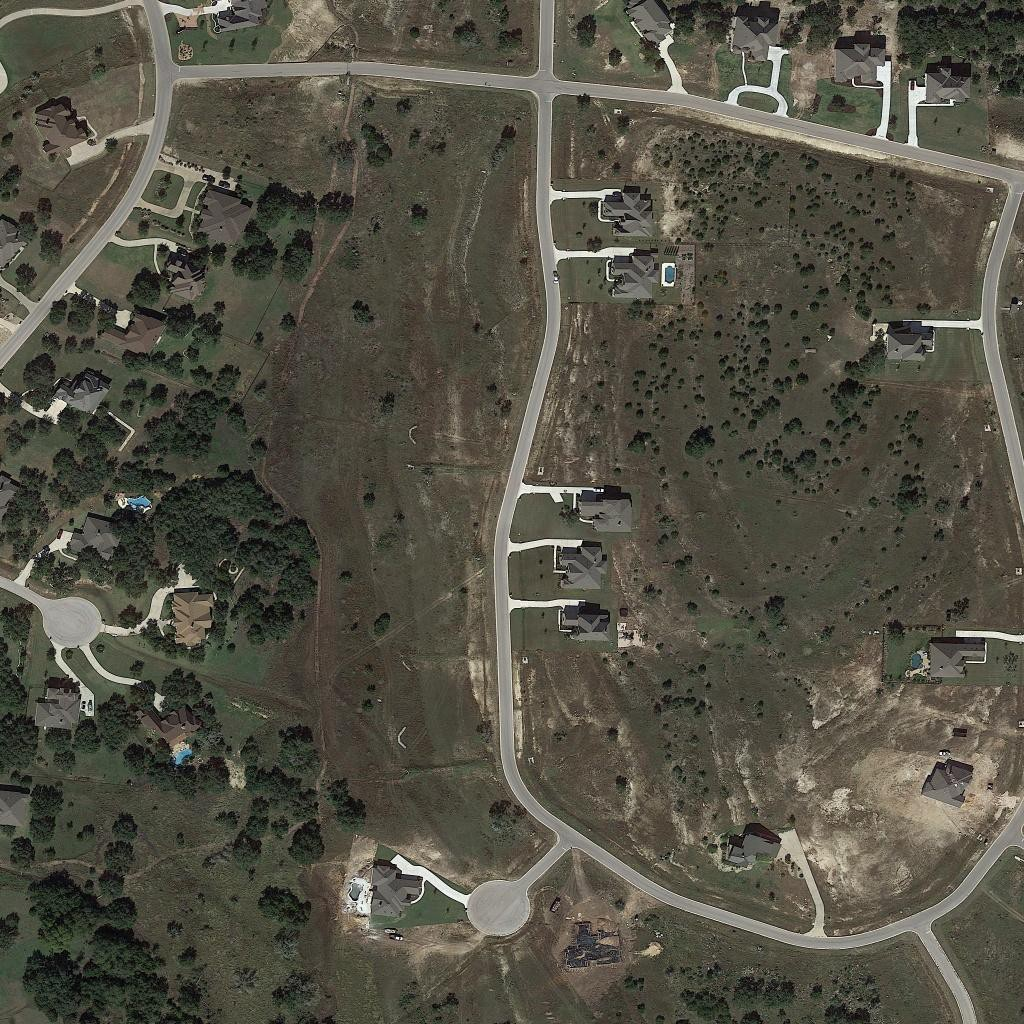}} & \taskexamplecell{Represent this change mask for change detection retrieval.} & {\includegraphics[width=0.6\linewidth]{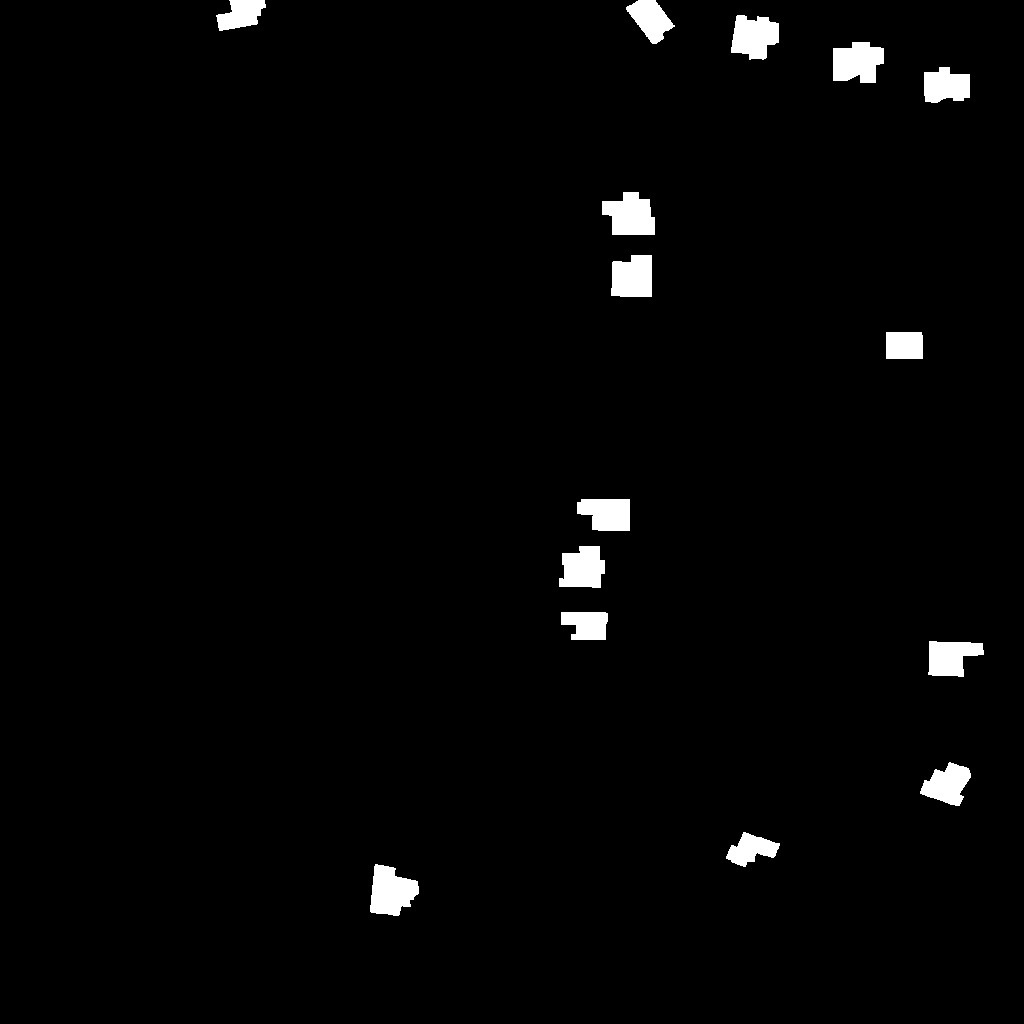}} \\
\midrule
SYSU-CD & change mask matching & \taskexamplecell{Represent the given remote sensing images with the following change detection task.}\par \textbf{Given a pair of remote sensing images from time1 and time2, retrieve the change mask that highlights the differences between them.} & {\includegraphics[width=0.45\linewidth]{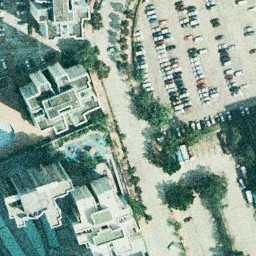}} {\includegraphics[width=0.45\linewidth]{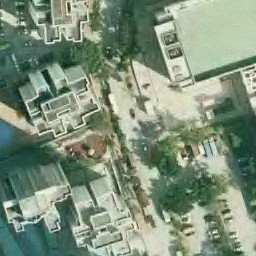}} & \taskexamplecell{Represent this change mask for change detection retrieval.} & {\includegraphics[width=0.6\linewidth]{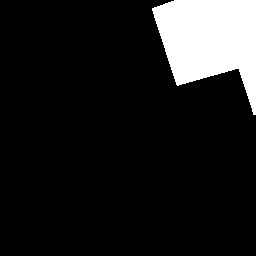}} \\
\midrule
VL-CMU-CD & change mask matching & \taskexamplecell{Represent the given remote sensing images with the following change detection task.}\par \textbf{Given a pair of remote sensing images from time1 and time2, retrieve the change mask that highlights the differences between them.} & {\includegraphics[width=0.45\linewidth]{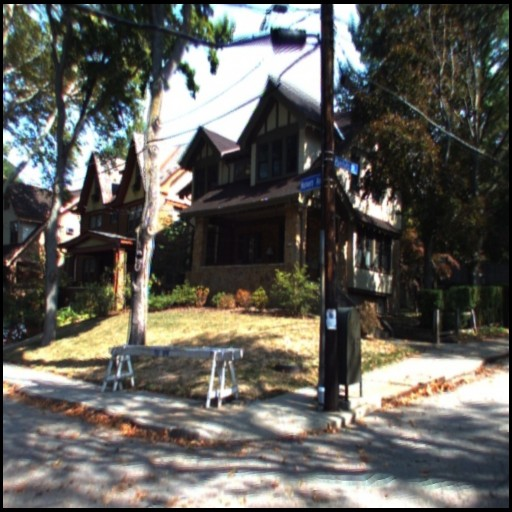}} {\includegraphics[width=0.45\linewidth]{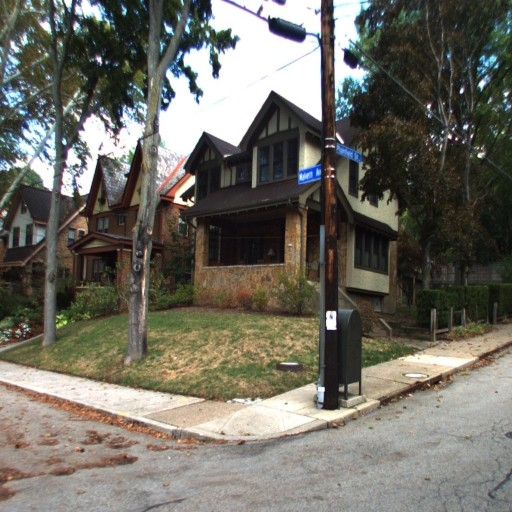}} & \taskexamplecell{Represent this change mask for change detection retrieval.} & {\includegraphics[width=0.6\linewidth]{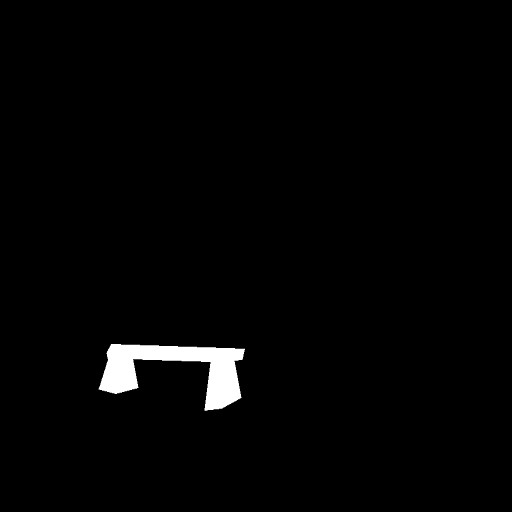}} \\
\midrule
OSF Recovery & disaster damage assessment & \taskexamplecell{Assess the post-disaster recovery status of this location by comparing multi-temporal street-level images.} & {\includegraphics[width=0.45\linewidth]{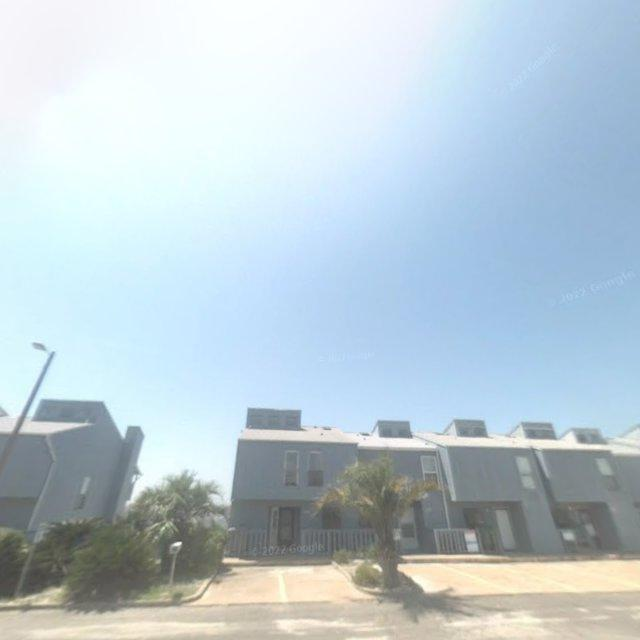}} {\includegraphics[width=0.45\linewidth]{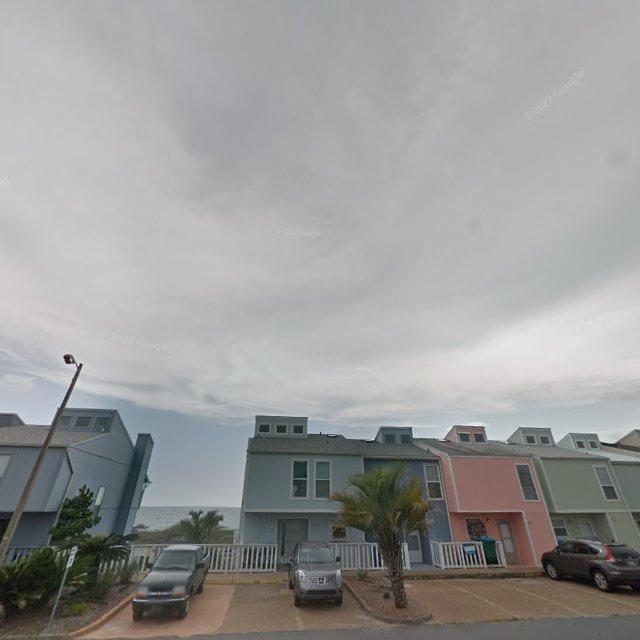}}\par {\includegraphics[width=0.45\linewidth]{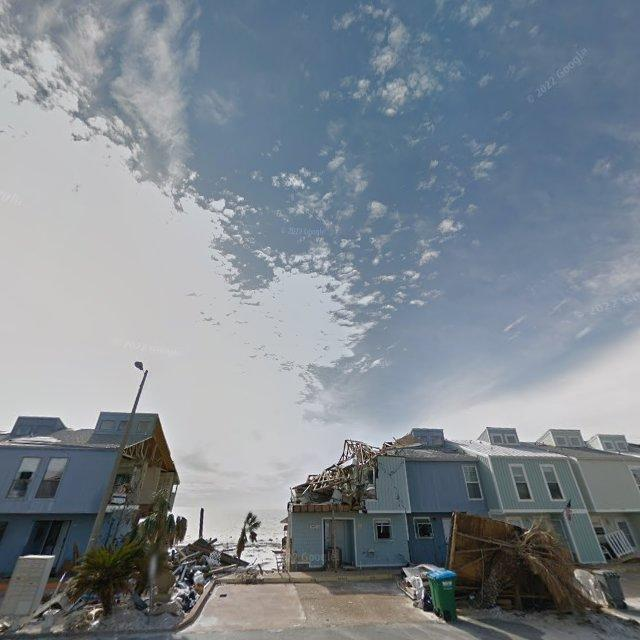}} {\includegraphics[width=0.45\linewidth]{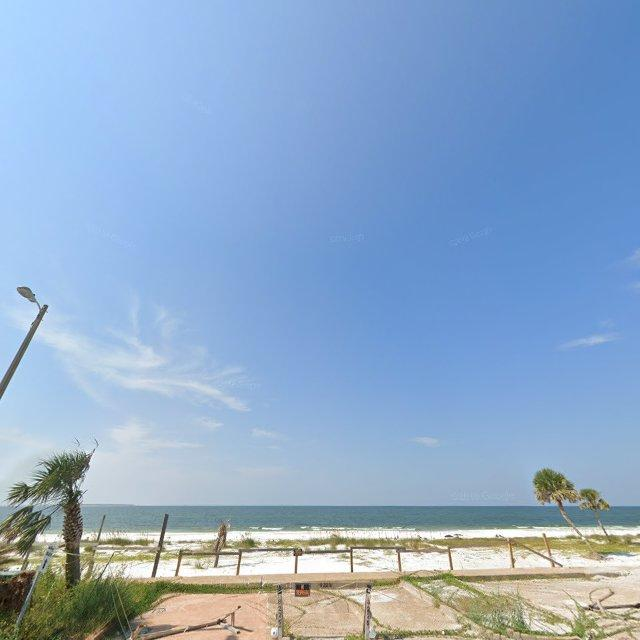}}\par {\includegraphics[width=0.45\linewidth]{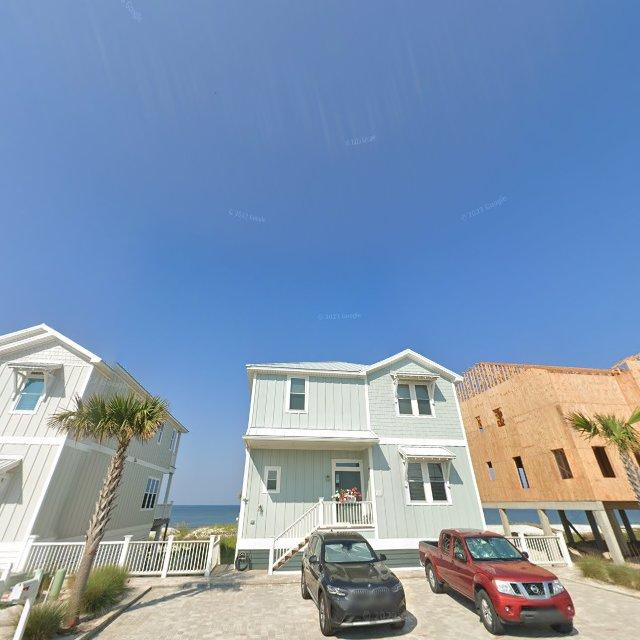}} & \taskexamplecell{Represent the given post-disaster recovery assessment.}\par \textbf{Improved Reconstruction: The quality of the reconstructed buildings is better than before the disaster.} & - \\
\bottomrule
\end{tabular}
\caption{Change-detection subtask examples.}
\label{tab:supp_task_examples_change}
\end{table*}


\begin{table*}[!t]
\centering
\tiny
\setlength{\tabcolsep}{2.5pt}
\renewcommand{\arraystretch}{0.78}
\begin{tabular}{m{0.06\textwidth}m{0.06\textwidth}m{0.18\textwidth}m{0.35\textwidth}>{\columncolor{gray!12}}m{0.17\textwidth}>{\columncolor{gray!12}}m{0.04\textwidth}}
\toprule
\textbf{Dataset} & \textbf{Subtask} & \textbf{Query text} & \textbf{Query image} & \textbf{Target text} & \textbf{Target image} \\
\midrule
AID & scene classification & \taskexamplecell{Identify the scene category in the given remote sensing image.} & {\includegraphics[width=0.24\linewidth]{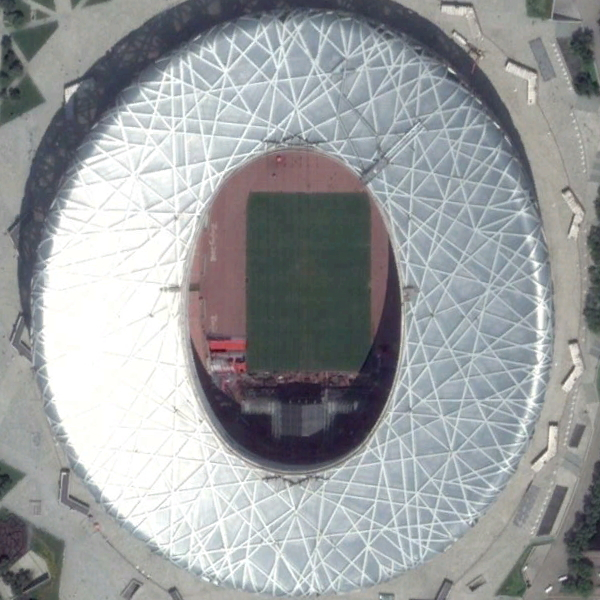}} & \taskexamplecell{Represent the remote sensing scene category.}\par \textbf{stadium} & - \\
\midrule
PlacePulse & safety/\par liveliness/\par beauty/\par wealth/\par depression/\par boringness & \taskexamplecell{Assess the perceived safety/liveliness/beauty/wealth/depression/boringness of this street view image.} & {\includegraphics[width=0.32\linewidth]{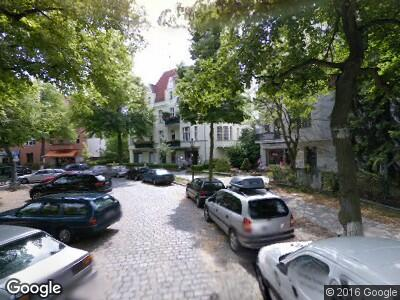}} & \taskexamplecell{Represent the safety/liveliness/beauty/wealth/depression/boringness perception level of a street view image.}\par \textbf{A street view image with wealth score in [6.5, 7.0)} & - \\
\midrule
CityLens & gdp/\par population density/\par average building height/\par healthcare accessibility & \taskexamplecell{Given a satellite image and multiple street-view images of an urban area, select the normalized GDP/population density/average building height/healthcare accessibility score (0.0 to 10.0) that best describes this area.} & {\includegraphics[width=0.24\linewidth]{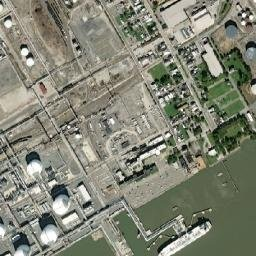}} {\includegraphics[width=0.32\linewidth]{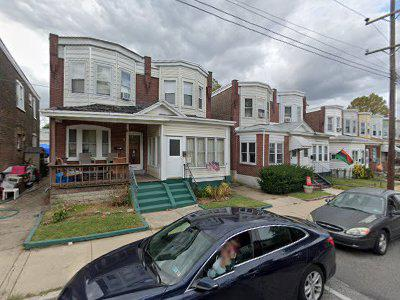}} {\includegraphics[width=0.32\linewidth]{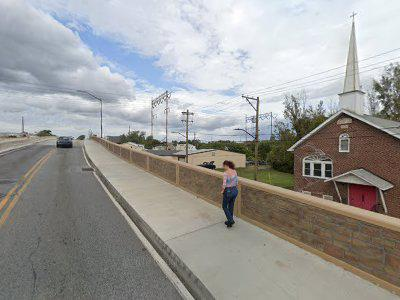}} \par {\includegraphics[width=0.32\linewidth]{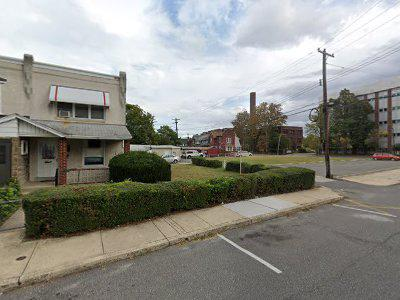}} {\includegraphics[width=0.32\linewidth]{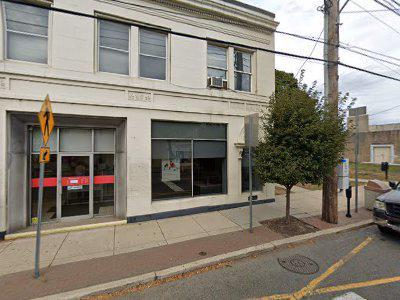}} {\includegraphics[width=0.32\linewidth]{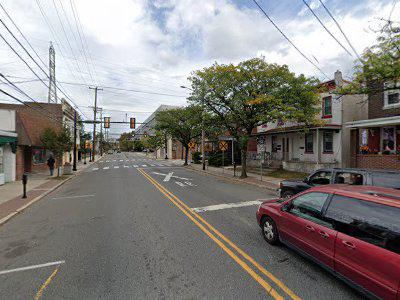}} \par {\includegraphics[width=0.32\linewidth]{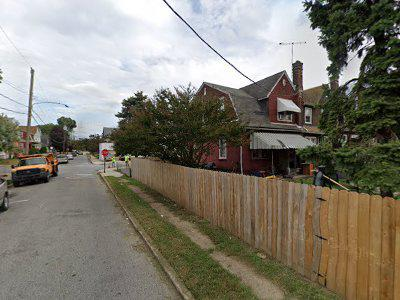}} {\includegraphics[width=0.32\linewidth]{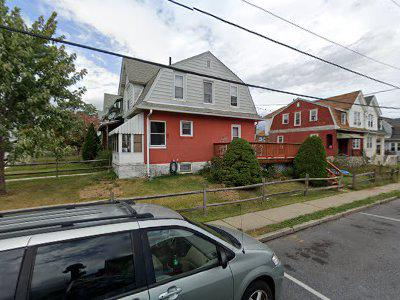}} {\includegraphics[width=0.32\linewidth]{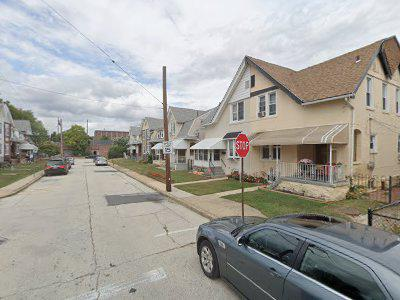}} \par {\includegraphics[width=0.32\linewidth]{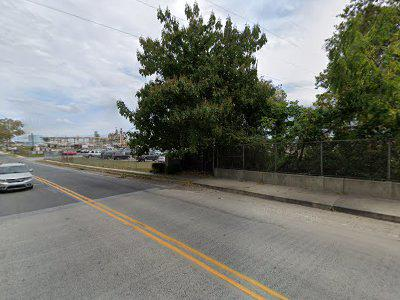}} {\includegraphics[width=0.32\linewidth]{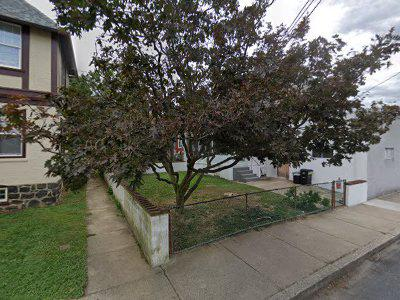}} & \taskexamplecell{Represent the given GDP (PPP 2005 international dollars) /population density/average building height/healthcare accessibility score.}\par \textbf{An urban area with health index score in [9.5, 10.0).} & - \\
\midrule
UrBench & scene-recognition & \taskexamplecell{Represent the given image with the following question.}\par \textbf{Which class of object appears in these images? Classes: bus station, fountain, graveyard, memorial or monument, motorway, parking, pitch, rail, residential, stadium, swimming pool, tower. Note that the residential class has the lowest priority and is selected when no other class matches the given image.} & {\includegraphics[width=0.25\linewidth]{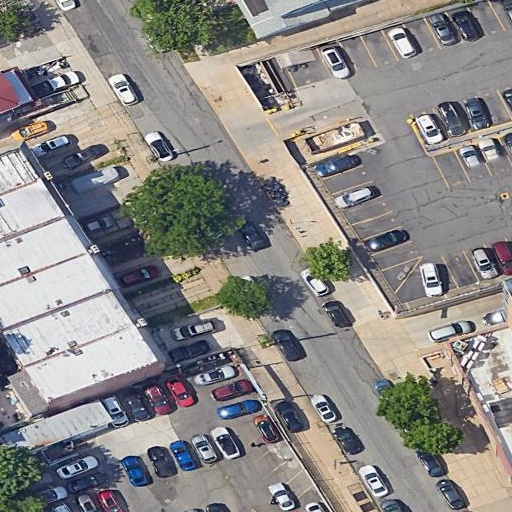}} {\includegraphics[width=0.5\linewidth]{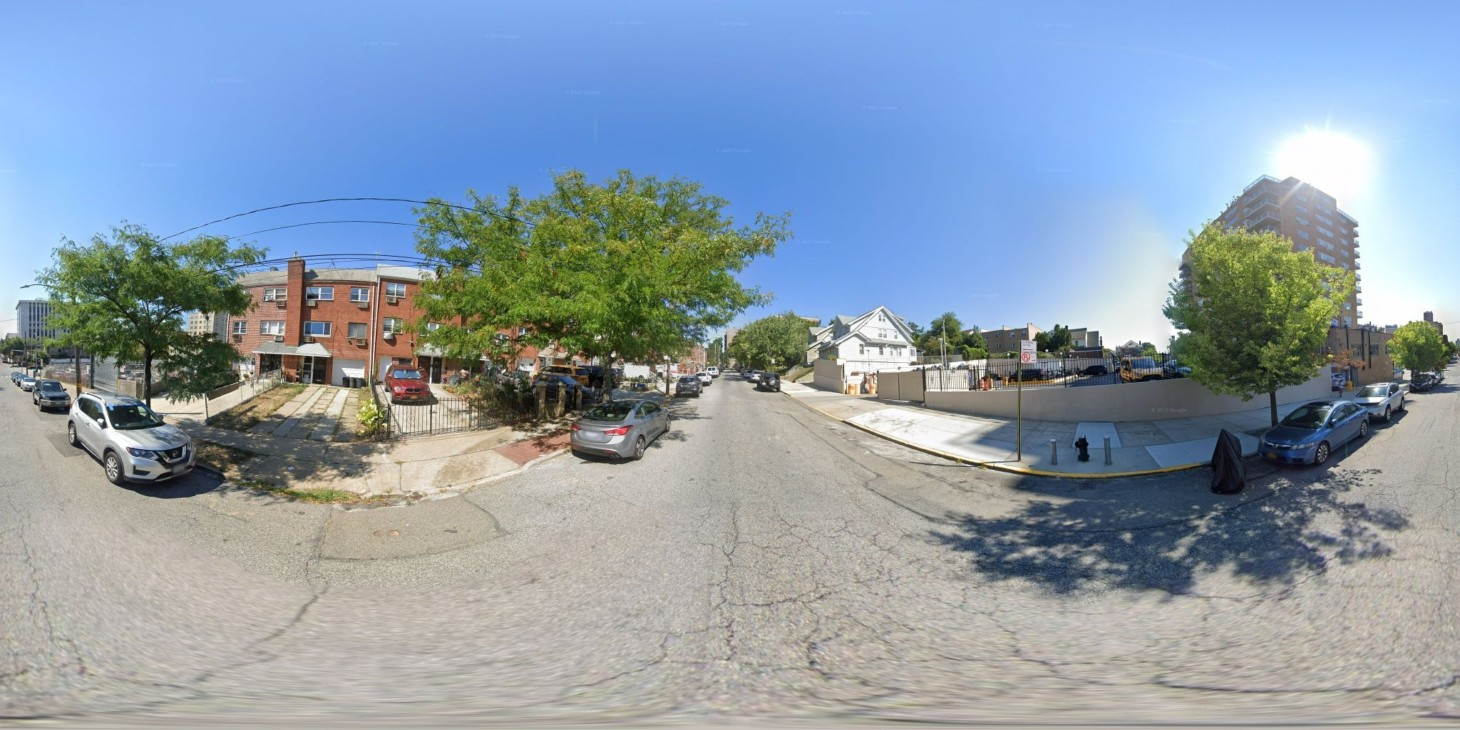}} & \taskexamplecell{Represent the user's input.}\par \textbf{parking} & - \\
\bottomrule
\end{tabular}
\caption{Classification subtask examples.}
\label{tab:supp_task_examples_classification}
\end{table*}

\begin{figure*}[p]
\centering
\begin{tcolorbox}[
    enhanced,
    colback=blue!4,
    colframe=black,
    coltitle=white,
    colbacktitle=black!75,
    title={Qualitative example of SV Localization task (i2t/t2i)},
    fonttitle=\bfseries,
    fontupper=\small,
    boxrule=0.8pt,
    arc=3mm,
    left=8pt,
    right=8pt,
    top=8pt,
    bottom=8pt
]
\centering
\includegraphics[width=0.55\textwidth]{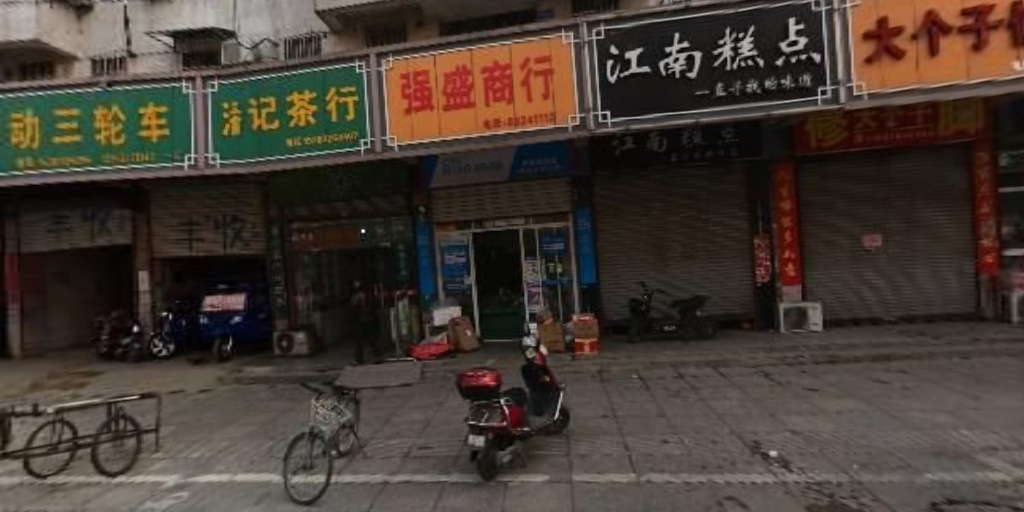}

\medskip
\raggedright
A quiet commercial street in China with closed storefronts, parked scooters and bicycles, visible Chinese signage, overcast daylight, eye-level forward-facing view, no pedestrians or moving traffic.

\medskip
The scene depicts a subdued urban commercial street lined with single-story storefronts beneath weathered multi-story buildings, their facades showing signs of age and exposure. Several shops remain shuttered with metal roll-down doors, while one central store displays an open glass entrance flanked by stacked cardboard boxes and promotional materials. Prominent signage above the shops includes Tricycle, Haiji Tea Shop, Qiang Sheng Trading Company, Jiangnan Pastries, and Big Man, rendered in bold Chinese characters on colored backgrounds---green, orange, black, and red---with some featuring decorative white borders. Parked along the paved sidewalk are two bicycles and a red electric scooter with a rear storage box; further back, a blue three-wheeled utility vehicle is partially visible under an awning. The road surface appears worn, with faint lane markings and scattered debris, suggesting low recent activity. No pedestrians or moving vehicles are present, contributing to a still, almost dormant atmosphere. Lighting is diffuse and even, consistent with overcast daylight, casting soft shadows and muting color saturation. The camera angle is eye-level and forward-facing, capturing the full width of the street frontage without distortion or artistic framing. There are no visible traffic signals, streetlights, or public transit infrastructure. The overall impression is of a typical neighborhood commercial strip during off-hours, possibly early morning or midday lull, where daily commerce has yet to fully activate. This setting aligns with a residential-commercial zone in a smaller Chinese city, where local businesses cater to daily needs and foot traffic remains modest outside peak hours.
\end{tcolorbox}
\caption{A qualitative example of the SV Localization image-to-text and text-to-image tasks. The annotated caption provides a detailed street-level scene description, including storefronts, signage, parked vehicles, lighting conditions, and local commercial context.}
\label{fig:sv_loc_caption_example}
\end{figure*}

\begin{figure*}[p]
\centering
\begin{tcolorbox}[
    colback=blue!4,
    colframe=black,
    coltitle=white,
    colbacktitle=black!75,
    title={Qualitative example of Im2GPS3k caption task (i2t/t2i)},
    fonttitle=\bfseries,
    fontupper=\footnotesize,
    boxrule=0.8pt,
    arc=3mm,
    left=8pt,
    right=8pt,
    top=8pt,
    bottom=8pt
]
\centering
\includegraphics[width=0.55\textwidth]{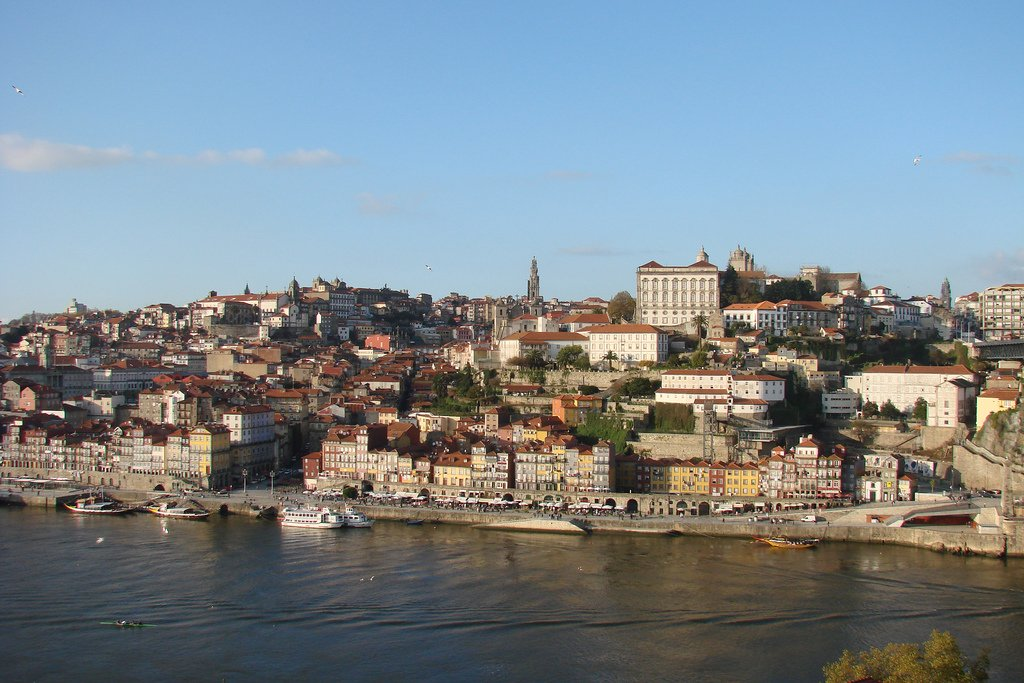}

\medskip
\raggedright
\textbf{Caption ID:} cap\_txt\_315251374

\medskip
The scene presents a panoramic view of a densely built European city situated on a steep, terraced hillside that descends to a wide riverbank. The architecture is characterized by tightly packed, multi-story buildings with predominantly red-tiled roofs and facades painted in a variety of muted pastel and earth tones. The river in the foreground is calm, reflecting the sky and the structures along its edge, with several small boats moored or navigating its surface. A broad, paved riverside promenade runs along the water's edge, lined with arches and outdoor seating areas, suggesting commercial or leisure activity. Above the riverfront, the city climbs the hill in successive tiers, with narrow streets and stairways weaving between buildings. Prominent structures include a large, pale, neoclassical-style building with a symmetrical facade and a central dome, situated near the crest of the hill, and a tall, slender church spire visible further back. The sky is clear and blue with minimal cloud cover, indicating fair weather and bright daylight conditions.

\medskip
The reasoning process for identifying the location is based on several key visual markers:
\begin{enumerate}
    \item The topography---a steep, densely urbanized hillside descending to a major river---is characteristic of Porto, Portugal, which is built along the Douro River.
    \item The architectural style---colorful, tiled-roof buildings clustered tightly on a slope---is emblematic of Porto's Ribeira district.
    \item The large, pale neoclassical building near the top of the hill corresponds to the Palácio de Cristal or the adjacent Museu de Arte Contemporânea, located in the Jardim do Morro area overlooking the river.
    \item The presence of traditional Rabelo boats, historically used to transport Port wine, and modern tour boats along the riverbank further supports the identification, as this is a common sight in Porto's Ribeira.
    \item The overall composition---including the river's curve, the alignment of the buildings, and the vantage point---matches well-known photographic perspectives of Porto from the Dom Luís I Bridge or the opposite bank in Vila Nova de Gaia.
\end{enumerate}

\textbf{True geographic location:} city: Porto; latitude: 41.1496; longitude: -8.6109.
\end{tcolorbox}
\caption{A qualitative example of the Im2GPS3k caption image-to-text and text-to-image tasks. The annotated description provides geolocation-relevant visual and contextual cues, such as topography, riverfront layout, architecture, landmarks, boats, and geographic coordinates, supporting fine-grained location matching.}
\label{fig:im2gps3k_caption_example}
\end{figure*}


\newcommand{\urbanresultheader}{%
\textbf{Dataset} & \textbf{Subtask} & \textbf{Metric} &
\shortstack[c]{CLIP} &
\shortstack[c]{OpenCLIP} &
\shortstack[c]{RemoteCLIP} &
\shortstack[c]{GeoRSCLIP} &
\shortstack[c]{SkyScript-CLIP} &
\shortstack[c]{SigLIP2} &
\shortstack[c]{GeoChat} &
\shortstack[c]{Qwen3-VL-\\Instruct 2B} &
\shortstack[c]{Qwen3-VL-\\Instruct 8B} &
\shortstack[c]{UME-R1} &
\shortstack[c]{VLM2Vec-V2.0} &
\shortstack[c]{OmniEmbed-\\Nemotron} &
\shortstack[c]{Seed1.6-\\Embedding} &
\shortstack[c]{Qwen3-VL-\\Embedding 8B} &
\shortstack[c]{Qwen3-VL-\\Embedding 2B} &
\shortstack[c]{Geo-Embed}%
}

\newpage
\section{Detailed results}
\label{sec:detailed_results}

Appendix Table~\ref{tab:appendix_retrieval_results}$\sim$\ref{tab:appendix_classification_results} report the full task-level results for all 45 GeoMEB evaluation subtasks, grouped by meta-task. These tables complement the aggregate results in Table~\ref{tab:main_results} by showing how each model performs on individual datasets and task formulations, including retrieval directions, VQA subsets, grounding sources, change-detection benchmarks, and classification labels. The detailed breakdown shows that model rankings are not uniform across subtasks. For example, several embedding-specialized models achieve very high scores on Im2GPS3k caption retrieval, but remain weak on GPS-region retrieval from the same dataset. Similarly, change-detection results vary sharply across benchmarks, with some models supporting OSF Recovery but failing to handle mask-based multi-image inputs. This task-level variation supports the need for reporting both aggregate scores and fine-grained results when evaluating geospatial embedding models.

\section{Geo-Embed Training details}
\label{sec:training_eval_details}

This section summarizes the default training configuration used for Geo-Embed. Geo-Embed is initialized from Qwen3-VL-Embedding-2B and fine-tuned with LoRA-based adaptation under the instruction-conditioned contrastive objective described earlier. Appendix Table~\ref{tab:training_configuration} reports the main optimization, adaptation, sequence-length, and hardware settings for our final Geo-Embed training run. The LoRA rank and batch size follow the best-performing setting identified in the hyperparameter analysis.

\section{Baseline model details}
\label{sec:baseline_model_details}

This section summarizes the baseline models evaluated in GeoMEB, grouped by model family. We specify the evaluated checkpoint or model variant for each baseline.

\paragraph{CLIP-family models.}
\begin{itemize}
    \item \textbf{CLIP}~\cite{radfordLearningTransferableVisual2021} is the original contrastive image-text dual encoder trained on large-scale natural image-caption pairs. It represents the standard general-purpose image-text alignment paradigm, where images and text are encoded by separate towers and matched through cosine similarity. We evaluate the CLIP ViT-L/14 checkpoint.

    \item \textbf{OpenCLIP}~\cite{fang2023data} follows the CLIP training framework with open large-scale training data and improved data filtering. It provides a strong open CLIP-family baseline for general image-text retrieval. We evaluate the DFN5B-CLIP-ViT-H-14 checkpoint.

    \item \textbf{RemoteCLIP}~\cite{liu2024remoteclipa} adapts CLIP-style contrastive learning to remote sensing vision-language data. It is expected to better capture overhead imagery, land-use patterns, and remote-sensing visual semantics than generic image-text encoders. We evaluate the RemoteCLIP-ViT-L-14 checkpoint.

    \item \textbf{GeoRSCLIP}~\cite{zhang2024rs5m} is a geospatial CLIP model trained with RS5M, a large-scale remote-sensing image-text dataset. It provides a domain-specialized baseline for aligning overhead images with natural language descriptions. We evaluate the GeoRSCLIP-ViT-H-14 checkpoint.

    \item \textbf{SkyScript-CLIP}~\cite{wang2024skyscript} is trained on SkyScript, a semantically diverse remote-sensing image-text dataset. It tests whether richer remote-sensing captions improve transfer to heterogeneous urban retrieval and recognition tasks. We evaluate the released SkyScript-CLIP checkpoint.

    \item \textbf{SigLIP2}~\cite{tschannen2025siglip} is a recent CLIP-family encoder trained with a sigmoid contrastive objective and designed to improve multilingual semantic understanding, localization, and dense visual representations. We evaluate the SigLIP2 SO400M-384 checkpoint.
\end{itemize}

\paragraph{Instruction-tuned VLMs.}
\begin{itemize}
    \item \textbf{GeoChat}~\cite{kuckreja2024geochat} is a geospatial vision-language model designed for grounded visual instruction following. Although it is not trained as a retrieval embedder, it tests whether geospatial instruction tuning produces transferable embedding representations. We evaluate the GeoChat-7B checkpoint.

    \item \textbf{Qwen3-VL-Instruct}~\cite{bai2025qwen3vl} is a general-purpose instruction-tuned vision-language model from the Qwen3-VL family. It tests whether multimodal instruction-following ability can be directly reused for retrieval in GeoMEB. We evaluate Qwen3-VL-Instruct-2B and Qwen3-VL-Instruct-8B.
\end{itemize}

\paragraph{Embedding-specialized VLMs.}
\begin{itemize}
    \item \textbf{UME-R1}~\cite{lan2025umer1} explores reasoning-driven generative multimodal embeddings. It represents a recent direction that uses reasoning-oriented training to improve embedding quality beyond conventional image-text contrastive alignment. We evaluate the UME-R1-7B checkpoint.

    \item \textbf{VLM2Vec-V2.0}~\cite{mengVLM2VecV2AdvancingMultimodal2025} extends VLM2Vec-style training to broader multimodal embedding settings, including images, videos, and visual documents. It serves as a strong open embedding-specialized VLM baseline. We evaluate the VLM2Vec-V2.0-3B checkpoint.

    \item \textbf{OmniEmbed-Nemotron}~\cite{xu2025omniembednemotrona} is a unified multimodal retrieval model covering text, image, audio, and video. Its training objective emphasizes cross-modal retrieval across multiple input types, making it a relevant baseline for testing whether general omni-modal embedding transfers to geospatial tasks. We evaluate the OmniEmbed-Nemotron-3B checkpoint.

    \item \textbf{Qwen3-VL-Embedding}~\cite{liQwen3VLEmbeddingQwen3VLRerankerUnified2026} is a native multimodal embedding model built on the Qwen3-VL family and one of the strongest open embedding-specialized VLMs in our comparison. It provides strong pretrained cross-modal alignment for general-purpose retrieval, making it both a competitive baseline and the initialization backbone for Geo-Embed. We evaluate Qwen3-VL-Embedding-2B and Qwen3-VL-Embedding-8B.

    \item \textbf{Seed1.6-Embedding}~\cite{bytedanceseed2025seed16embedding} is a closed-source multimodal embedding model released as part of the Seed1.6 model family. It targets general-purpose multimodal retrieval and serves as one of the strongest proprietary baselines in our experiments. We evaluate it through the \texttt{doubao-embedding-vision-251215} API.
\end{itemize}

\begin{table}[t]
\centering
\footnotesize
\setlength{\tabcolsep}{4pt}
\renewcommand{\arraystretch}{0.92}
\begin{tabular}{@{}p{0.38\linewidth}p{0.54\linewidth}@{}}
\toprule
\textbf{Item} & \textbf{Setting} \\
\midrule
Backbone & Qwen3-VL-Embedding 2B \\
Adaptation & LoRA \\
Optimizer & AdamW \\
Learning rate & $2\times10^{-5}$ \\
Weight decay & 0.1 \\
Schedule & Cosine decay \\

Warmup ratio & 0.05 \\
Training epoch & 1 \\
Validation split & 0.01 \\
LoRA rank / alpha & 16 / 32 \\
Temperature & 0.02 \\
Batch size & 1024 \\
Max sequence length & 4096 \\
Precision & bf16 \\
Gradient clipping & 1.0 \\
Random seed & 42 \\
Gradient Checkpoint & True \\
Distributed training & MS-Swift + DeepSpeed ZeRO-2 \\
Hardware & 4$\times$RTX 4090 + 4$\times$L20 (NVIDIA)\\
\bottomrule
\end{tabular}
\caption{Default training configuration for Geo-Embed.}
\label{tab:training_configuration}
\end{table} 

\begin{table*}[t]
\centering
\scriptsize
\setlength{\tabcolsep}{2.0pt}
\renewcommand{\arraystretch}{0.86}
\resizebox{\textwidth}{!}{%
\begin{tabular}{llc*{16}{c}}
\toprule
\urbanresultheader \\
\midrule
RSICD & i2t & Mean R@1/5/10 & 18.88 & 27.08 & 36.47 & 30.83 & 19.82 & 27.69 & 0.46 & 0.37 & 1.04 & 26.96 & 26.72 & 14.91 & 25.37 & 30.59 & 31.93 & 36.29 \\
RSICD & t2i & Mean R@1/5/10 & 18.79 & 28.53 & 35.28 & 29.65 & 21.03 & 27.52 & 0.48 & 0.72 & 0.82 & 27.70 & 28.58 & 15.53 & 26.10 & 34.01 & 30.57 & 33.40 \\
UAV-GeoLoc & country & R@1 & 7.70 & 18.09 & 0.17 & 15.62 & 12.85 & 9.93 & 0.46 & 0.00 & 0.00 & 8.13 & 7.39 & 4.50 & 7.67 & 13.52 & 13.70 & 29.20 \\
UAV-GeoLoc & terrain & R@1 & 11.33 & 18.31 & 0.75 & 22.73 & 19.23 & 5.62 & 0.87 & 0.02 & 0.16 & 8.57 & 10.28 & 6.28 & 11.41 & 17.00 & 17.74 & 34.41 \\
VIGOR & -- & R@1 & 1.49 & 3.42 & 0.26 & 1.75 & 1.75 & 2.41 & 0.07 & 0.00 & 0.22 & 2.38 & 3.16 & 2.60 & 2.86 & 2.27 & 1.78 & 3.12 \\
SV Localization & i2t & Mean R@1/5/10 & 7.40 & 23.27 & 5.12 & 18.21 & 7.47 & 15.95 & 0.01 & 0.03 & 0.03 & 1.30 & 81.61 & 38.72 & 82.47 & 97.76 & 92.93 & 94.97 \\
SV Localization & t2i & Mean R@1/5/10 & 9.11 & 26.16 & 4.39 & 20.88 & 10.69 & 14.92 & 0.01 & 0.03 & 0.03 & 13.54 & 84.40 & 17.36 & 85.30 & 96.43 & 92.48 & 95.82 \\
Im2GPS3k & GPS i2t & R@1 & 0.07 & 0.03 & 0.03 & 0.07 & 0.03 & 0.07 & 0.37 & 0.03 & 0.53 & 5.37 & 0.17 & 0.90 & 9.54 & 8.48 & 0.83 & 2.00 \\
Im2GPS3k & GPS t2i & R@1 & 0.00 & 0.09 & 0.00 & 0.09 & 0.00 & 0.09 & 0.09 & 0.09 & 0.00 & 0.44 & 0.18 & 0.09 & 2.63 & 2.54 & 0.26 & 0.26 \\
Im2GPS3k & Caption i2t & Mean R@1/5/10 & 84.41 & 94.35 & 65.96 & 86.81 & 85.10 & 75.36 & 2.90 & 0.37 & 0.74 & 62.92 & 94.16 & 96.12 & 98.95 & 99.10 & 99.48 & 99.74 \\
Im2GPS3k & Caption t2i & Mean R@1/5/10 & 82.53 & 93.28 & 52.23 & 83.23 & 84.65 & 76.40 & 2.88 & 0.68 & 1.37 & 90.49 & 97.64 & 94.29 & 99.37 & 99.65 & 99.34 & 99.76 \\
VRSBench & caption i2t & Mean R@1/5/10 & 7.02 & 14.01 & 4.94 & 16.14 & 8.68 & 7.89 & 0.58 & 0.11 & 0.59 & 11.08 & 15.34 & 8.36 & 18.12 & 23.32 & 16.63 & 26.52 \\
VRSBench & caption t2i & Mean R@1/5/10 & 6.36 & 13.84 & 3.65 & 14.90 & 8.34 & 9.81 & 0.27 & 0.09 & 0.12 & 14.92 & 17.90 & 5.02 & 17.93 & 25.66 & 18.67 & 25.12 \\
\rowcolor{orange!12}\multicolumn{3}{l}{\textbf{Average}} & \textbf{19.62} & \textbf{27.73} & \textbf{16.10} & \textbf{26.22} & \textbf{21.51} & \textbf{21.05} & \textbf{0.73} & \textbf{0.20} & \textbf{0.43} & \textbf{21.06} & \textbf{35.96} & \textbf{23.44} & \textbf{37.52} & \textbf{42.33} & \textbf{39.72} & \textbf{44.66} \\
\bottomrule
\end{tabular}}
\caption{Detailed results on retrieval tasks.}
\label{tab:appendix_retrieval_results}
\end{table*}

\begin{table*}[t]
\centering
\scriptsize
\setlength{\tabcolsep}{2.0pt}
\renewcommand{\arraystretch}{0.86}
\resizebox{\textwidth}{!}{%
\begin{tabular}{llc*{16}{c}}
\toprule
\urbanresultheader \\
\midrule
VRSBench & vqa & R@1 & 0.99 & 0.93 & 1.48 & 0.41 & 0.89 & 0.09 & 0.02 & 0.04 & 0.09 & 35.45 & 0.52 & 0.06 & 14.33 & 40.05 & 34.42 & 35.66 \\
MME-RealWorld & AutonomousDriving & R@1 & 18.10 & 22.43 & 23.67 & 21.04 & 18.56 & 18.10 & 18.72 & 21.81 & 22.04 & 33.87 & 28.92 & 17.56 & 34.57 & 37.97 & 36.27 & 42.23 \\
MME-RealWorld & MME-HD-CN & R@1 & 2.19 & 9.50 & 9.06 & 14.77 & 2.49 & 3.80 & 15.64 & 28.22 & 27.19 & 55.85 & 36.55 & 28.22 & 61.40 & 68.27 & 64.47 & 59.36 \\
MME-RealWorld & diagram\_and\_table & R@1 & 4.80 & 6.86 & 3.77 & 4.37 & 4.46 & 0.26 & 10.37 & 24.68 & 32.56 & 64.87 & 32.73 & 9.68 & 55.01 & 76.18 & 66.67 & 60.41 \\
MME-RealWorld & monitoring\_images & R@1 & 11.23 & 11.68 & 11.68 & 10.18 & 10.63 & 9.43 & 18.41 & 14.97 & 16.62 & 39.37 & 15.42 & 10.18 & 26.50 & 41.77 & 38.77 & 39.82 \\
MME-RealWorld & ocr\_cc & R@1 & 6.72 & 20.15 & 7.36 & 7.12 & 6.72 & 2.75 & 9.55 & 29.61 & 32.20 & 61.08 & 40.94 & 21.44 & 60.92 & 72.25 & 67.15 & 62.94 \\
MME-RealWorld & remote\_sensing & R@1 & 7.58 & 10.26 & 8.74 & 8.39 & 7.69 & 9.32 & 10.61 & 26.11 & 20.05 & 47.32 & 20.75 & 11.89 & 43.12 & 56.88 & 47.79 & 49.30 \\
UrBench & counting & R@1 & 20.45 & 27.27 & 7.95 & 18.18 & 19.32 & 20.45 & 27.27 & 1.14 & 15.91 & 22.73 & 2.27 & 11.36 & 43.18 & 44.32 & 23.86 & 39.77 \\
UrBench & object-attribute-recognition oe & R@1 & 3.53 & 0.00 & 1.18 & 3.53 & 3.53 & 25.88 & 2.35 & 11.76 & 10.59 & 28.24 & 14.12 & 2.35 & 15.29 & 4.71 & 4.71 & 27.06 \\
UrBench & road-understanding oe & R@1 & 50.98 & 49.02 & 49.02 & 50.98 & 50.98 & 50.98 & 50.98 & 49.02 & 49.02 & 50.98 & 47.06 & 50.98 & 50.98 & 56.86 & 50.98 & 84.31 \\
UrBench & role-based-reasoning & R@1 & 1.78 & 4.00 & 0.89 & 1.78 & 2.22 & 1.78 & 0.44 & 0.00 & 0.89 & 22.67 & 6.22 & 0.44 & 19.56 & 41.78 & 33.33 & 28.44 \\
UrBench & traffic-sign-reasoning & R@1 & 1.00 & 1.00 & 2.00 & 2.00 & 2.00 & 2.00 & 1.00 & 2.00 & 2.00 & 11.00 & 2.00 & 1.00 & 12.00 & 17.00 & 14.00 & 11.00 \\
UrBench & visual-prompt-reasoning & R@1 & 2.00 & 2.00 & 0.00 & 0.00 & 2.00 & 0.00 & 4.00 & 2.00 & 4.00 & 20.00 & 8.00 & 4.00 & 24.00 & 30.00 & 28.00 & 26.00 \\
\rowcolor{cyan!12}\multicolumn{3}{l}{\textbf{Average}} & \textbf{10.10} & \textbf{12.70} & \textbf{9.75} & \textbf{10.98} & \textbf{10.11} & \textbf{11.14} & \textbf{13.03} & \textbf{16.26} & \textbf{17.94} & \textbf{37.96} & \textbf{19.65} & \textbf{13.01} & \textbf{35.45} & \textbf{45.23} & \textbf{39.26} & \textbf{43.56} \\
\bottomrule
\end{tabular}}
\caption{Detailed results on visual question answering tasks.}
\label{tab:appendix_vqa_results}
\end{table*}

\begin{table*}[t]
\centering
\scriptsize
\setlength{\tabcolsep}{2.0pt}
\renewcommand{\arraystretch}{0.86}
\resizebox{\textwidth}{!}{%
\begin{tabular}{llc*{16}{c}}
\toprule
\urbanresultheader \\
\midrule
Cityscapes & -- & R@1 & 1.72 & 1.84 & 1.20 & 2.20 & 1.56 & 2.56 & 0.04 & 0.04 & 0.08 & 6.08 & 4.28 & 2.00 & 8.56 & 8.16 & 8.20 & 10.20 \\
Mapillary & -- & R@1 & 0.78 & 1.47 & 0.28 & 0.81 & 0.76 & 2.32 & 0.01 & 0.03 & 0.02 & 3.03 & 2.22 & 1.92 & 4.19 & 5.87 & 1.79 & 8.06 \\
VRSBench & referring & R@1 & 4.16 & 5.41 & 0.95 & 5.36 & 5.58 & 5.08 & 0.53 & 0.01 & 0.59 & 4.17 & 5.85 & 2.81 & 7.68 & 6.27 & 3.56 & 13.00 \\
\rowcolor{red!12}\multicolumn{3}{l}{\textbf{Average}} & \textbf{2.22} & \textbf{2.91} & \textbf{0.81} & \textbf{2.79} & \textbf{2.63} & \textbf{3.32} & \textbf{0.19} & \textbf{0.03} & \textbf{0.23} & \textbf{4.43} & \textbf{4.12} & \textbf{2.24} & \textbf{6.81} & \textbf{6.77} & \textbf{4.52} & \textbf{10.42} \\
\bottomrule
\end{tabular}}
\caption{Detailed results on grounding tasks.}
\label{tab:appendix_grounding_results}
\end{table*}

\begin{table*}[t]
\centering
\scriptsize
\setlength{\tabcolsep}{2.0pt}
\renewcommand{\arraystretch}{0.86}
\resizebox{\textwidth}{!}{%
\begin{tabular}{llc*{16}{c}}
\toprule
\urbanresultheader \\
\midrule
LEVIR-CD & -- & R@1 & 2.34 & 0.78 & 1.56 & 1.56 & 3.12 & 0.78 & 0.78 & 0.00 & 0.78 & 0.00 & 0.78 & 0.78 & 3.12 & 2.34 & 1.56 & 5.47 \\
SYSU-CD & -- & R@1 & - & - & - & - & - & - & - & 0.00 & 0.03 & 0.10 & 0.07 & 0.00 & 0.20 & 0.07 & 0.10 & 0.83 \\
VL-CMU-CD & -- & R@1 & - & - & - & - & - & - & - & 0.23 & 0.23 & 0.47 & 0.70 & 1.17 & 2.10 & 0.93 & 0.23 & 2.10 \\
OSF Recovery & -- & R@1 & - & - & - & - & - & - & - & 41.61 & 33.58 & 37.96 & 29.93 & 34.31 & 42.34 & 11.68 & 35.77 & 37.96 \\
\rowcolor{green!12}\multicolumn{3}{l}{\textbf{Average}} & \textbf{0.58} & \textbf{0.20} & \textbf{0.39} & \textbf{0.39} & \textbf{0.78} & \textbf{0.20} & \textbf{0.20} & \textbf{10.46} & \textbf{8.65} & \textbf{9.63} & \textbf{7.87} & \textbf{9.07} & \textbf{11.94} & \textbf{3.75} & \textbf{9.42} & \textbf{11.59} \\
\bottomrule
\end{tabular}}
\caption{Detailed results on change detection tasks. For models that do not support multi-image inputs, ``--'' denotes unavailable results and is counted as 0 when computing averages.}
\label{tab:appendix_change_detection_results}
\end{table*}

\begin{table*}[t]
\centering
\scriptsize
\setlength{\tabcolsep}{2.0pt}
\renewcommand{\arraystretch}{0.86}
\resizebox{\textwidth}{!}{%
\begin{tabular}{llc*{16}{c}}
\toprule
\urbanresultheader \\
\midrule
AID & scene classification & Acc. & 68.80 & 74.65 & 81.75 & 74.95 & 72.40 & 67.60 & 27.75 & 2.75 & 2.75 & 62.80 & 63.50 & 46.30 & 71.30 & 70.75 & 66.20 & 76.00 \\
PlacePulse & safety & R@1 & 1.60 & 3.85 & 0.29 & 10.53 & 1.27 & 2.07 & 12.38 & 3.70 & 7.66 & 11.62 & 4.28 & 8.53 & 12.45 & 4.94 & 2.80 & 9.01 \\
PlacePulse & lively & R@1 & 0.31 & 6.35 & 0.81 & 11.13 & 0.17 & 6.65 & 7.36 & 0.64 & 0.10 & 0.68 & 6.79 & 5.02 & 6.41 & 0.14 & 4.51 & 3.95 \\
PlacePulse & beautiful & R@1 & 5.13 & 5.33 & 1.05 & 7.70 & 4.67 & 10.59 & 13.42 & 1.58 & 0.33 & 6.78 & 8.03 & 10.13 & 11.64 & 1.25 & 5.46 & 4.92 \\
PlacePulse & wealthy & R@1 & 1.90 & 0.15 & 0.29 & 8.45 & 1.17 & 1.17 & 10.28 & 1.31 & 0.07 & 5.32 & 8.09 & 9.33 & 11.81 & 0.58 & 13.70 & 7.58 \\
PlacePulse & depressing & R@1 & 4.20 & 1.78 & 0.10 & 5.98 & 3.46 & 7.87 & 11.96 & 5.77 & 0.00 & 6.30 & 2.94 & 6.72 & 0.52 & 0.00 & 6.61 & 6.40 \\
PlacePulse & boring & R@1 & 3.52 & 1.99 & 0.15 & 7.03 & 5.66 & 3.67 & 11.16 & 3.06 & 0.00 & 6.12 & 6.57 & 5.05 & 3.98 & 0.00 & 3.52 & 5.05 \\
CityLens & gdp & R@1 & 5.10 & 4.30 & 4.90 & 4.20 & 6.40 & 4.90 & 6.30 & 6.90 & 6.20 & 5.80 & 5.90 & 5.90 & 4.40 & 9.30 & 6.20 & 8.50 \\
CityLens & pop & R@1 & 5.00 & 5.80 & 5.20 & 6.00 & 5.60 & 3.90 & 4.80 & 9.00 & 5.80 & 5.30 & 9.20 & 7.60 & 6.00 & 9.50 & 6.80 & 10.30 \\
CityLens & height & R@1 & 4.40 & 4.40 & 4.00 & 6.20 & 4.50 & 4.60 & 6.30 & 5.40 & 5.20 & 5.40 & 5.40 & 5.40 & 7.60 & 8.80 & 5.40 & 8.50 \\
CityLens & health & R@1 & 12.40 & 14.90 & 0.40 & 2.90 & 15.00 & 15.50 & 5.20 & 0.00 & 24.20 & 3.80 & 3.80 & 0.60 & 7.70 & 5.40 & 4.60 & 8.30 \\
UrBench & scene-recognition & R@1 & 26.67 & 9.17 & 33.33 & 12.50 & 21.67 & 5.00 & 3.33 & 19.17 & 10.00 & 39.17 & 37.50 & 21.67 & 41.67 & 37.50 & 55.00 & 57.50 \\
\rowcolor{blue!12}\multicolumn{3}{l}{\textbf{Average}} & \textbf{11.58} & \textbf{11.06} & \textbf{11.02} & \textbf{13.13} & \textbf{11.83} & \textbf{11.13} & \textbf{10.02} & \textbf{4.94} & \textbf{5.19} & \textbf{13.25} & \textbf{13.50} & \textbf{11.02} & \textbf{15.46} & \textbf{12.35} & \textbf{15.07} & \textbf{17.17} \\
\bottomrule
\end{tabular}}
\caption{Detailed results on classification tasks.}
\label{tab:appendix_classification_results}
\end{table*}

\end{document}